\pdfoutput=1
\documentclass[10pt,twocolumn,letterpaper]{article}
\usepackage[table,dvipsnames]{xcolor}

\usepackage{cvpr}
\usepackage{times}
\usepackage{epsfig}
\usepackage{graphicx}
\usepackage{amsmath}
\usepackage{amssymb}
\usepackage{enumitem}
\usepackage{multirow}
\usepackage{multicol}
\usepackage{pifont}
\usepackage{stfloats}
\usepackage{authblk}

\newcommand{\xmark}{\ding{55}}%

\usepackage{hyperref}
\hypersetup{pagebackref=true,breaklinks=true,colorlinks,bookmarks=false}

\hypersetup{
linkcolor=BrickRed
,citecolor=Green
,filecolor=Mulberry
,urlcolor=NavyBlue
,menucolor=BrickRed
,runcolor=Mulberry
,linkbordercolor=BrickRed
,citebordercolor=Green
,filebordercolor=Mulberry
,urlbordercolor=NavyBlue
,menubordercolor=BrickRed
,runbordercolor=Mulberry
}

\cvprfinalcopy 

\def\cvprPaperID{10594} 

\ifcvprfinal\fi
\begin{document}

\title{{SPARE3D}: A Dataset for {SPA}tial {RE}asoning on Three-View Line Drawings}

\renewcommand*{\Authsep}{\qquad}
\renewcommand*{\Authand}{\qquad}
\renewcommand*{\Authands}{\qquad}
\setlength{\affilsep}{0.5em}

\author[]{\vspace{-6mm} Wenyu Han\thanks{The first two authors contributed equally.}}
\author[]{Siyuan Xiang\protect\footnotemark[1]{}}
\author[]{Chenhui Liu}
\author[]{Ruoyu Wang}
\author[]{Chen Feng\thanks{Chen Feng is the corresponding author. \texttt{\href{mailto:cfeng@nyu.edu}{cfeng@nyu.edu}}}}
\affil[]{New York University Tandon School of Engineering}
\affil[]{\vspace{-1mm} \textbf{\url{https://ai4ce.github.io/SPARE3D}} \vspace{-5mm}}

\maketitle
\thispagestyle{empty}

\begin{abstract}
Spatial reasoning is an important component of human intelligence. We can imagine the shapes of 3D objects and reason about their spatial relations by merely looking at their three-view line drawings in 2D, with different levels of competence. Can deep networks be trained to perform spatial reasoning tasks? How can we measure their ``spatial intelligence''? To answer these questions, we present the SPARE3D dataset. Based on cognitive science and psychometrics, SPARE3D contains three types of 2D-3D reasoning tasks on view consistency, camera pose, and shape generation, with increasing difficulty. We then design a method to automatically generate a large number of challenging questions with ground truth answers for each task. They are used to provide supervision for training our baseline models using state-of-the-art architectures like ResNet. Our experiments show that although convolutional networks have achieved superhuman performance in many visual learning tasks, their spatial reasoning performance on SPARE3D tasks \textcolor{blue}{is either lower than average human performance or even close to random guesses.}\footnote{\textcolor{blue}{In our follow-up work after CVPR'20, we found outliers in our previous dataset. We remove the ourliers and modify this paper accordingly (highlighted in blue), although the main conclusions are not changed.}} We hope SPARE3D can stimulate new problem formulations and network designs for spatial reasoning to empower intelligent robots to operate effectively in the 3D world via 2D sensors.
\end{abstract}

\section{Introduction}

\begin{figure}[t!]
\centering
\includegraphics[width=\columnwidth]{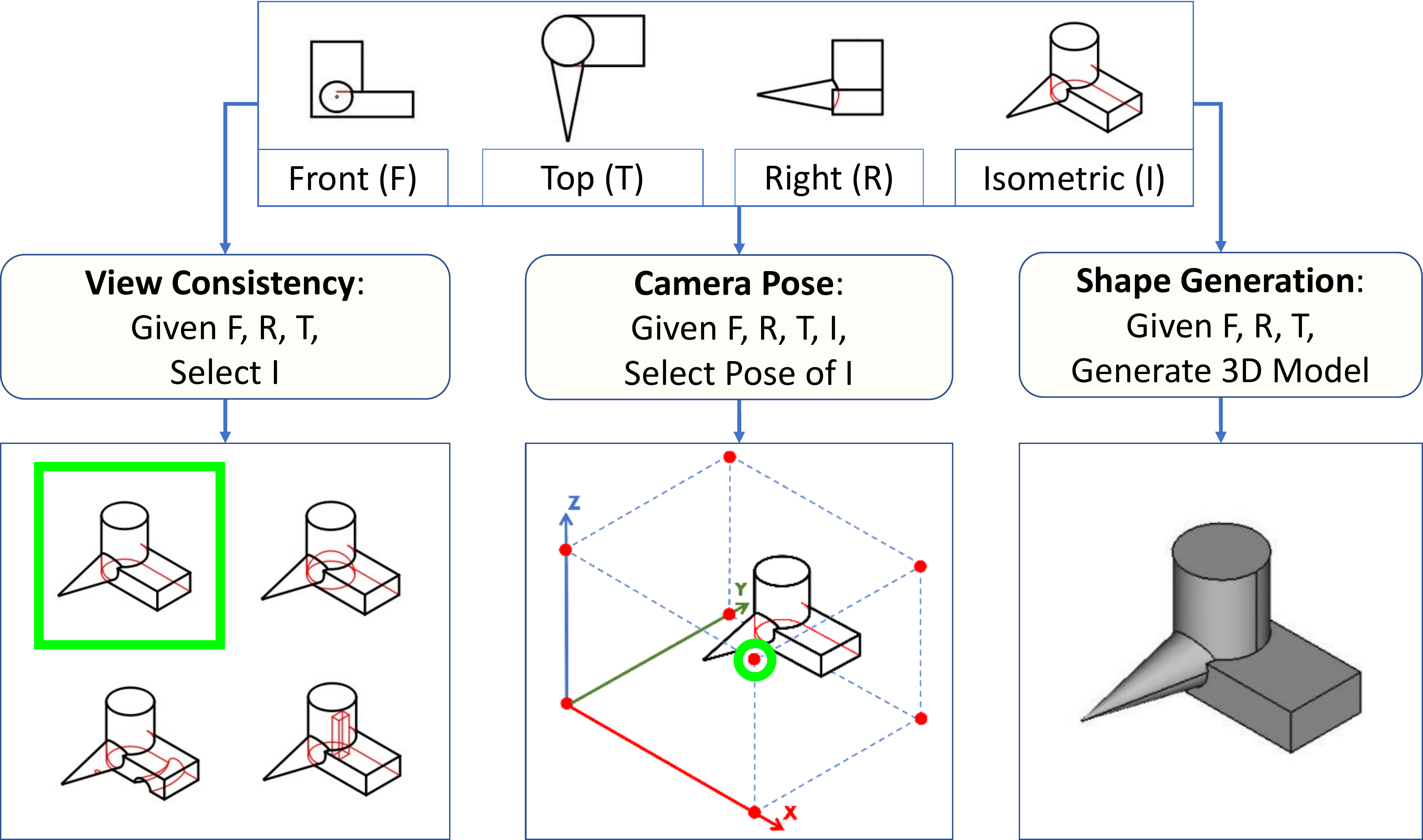}
\caption{\textbf{SPARE3D task overview}. The input to each task is either the whole or a subset of four different orthographic views of a 3D object as line drawings, i.e., front (F), top (T), right (R), and isometric (I) views. Based on the input, an intelligent agent needs to answer three types of questions: 1) select a consistent view describing the same object, 2) reason about the camera pose of a view, and 3) generate the object shape as an isometric view or a 3D model. The green box (left) and circle (middle) indicate the correct answers in this example. Best viewed in color.}
\label{fig_1}
\vspace{-5mm}
\end{figure}

Spatial reasoning is ``the ability to generate, retain, retrieve, and transform well-structured visual images''~\cite{lohman1996spatial}. It allows an intelligent agent to understand and reason about the relations among objects in three or two dimensions. As a part of general intelligence, spatial reasoning allows people to interpret their surrounding 3D world~\cite{lowrie2002influence} and affect their spatial task performances in large-scale environments~\cite{hsi1997role}. Moreover, statistics from many psychological and educational studies~\cite{kell2013creativity, lubinski2010spatial, wai2009spatial} have empirically proved that good spatial reasoning ability can benefit performance in STEM (science, technology, engineering, and math) areas.

Therefore, when we are actively developing intelligent systems such as self-driving cars and smart service robots, it is natural to ask how good their spatial reasoning abilities are, especially if they are not equipped with expensive 3D sensors. Because deep convolutional networks empower most state-of-the-art visual learning achievements (such as object detection and scene segmentation) in those systems, and they are typically trained and evaluated on a large amount of data, it is then important to design a set of non-trivial tasks and develop a large-scale dataset to facilitate the study of spatial reasoning for intelligent agents. 

As an important topic in psychometrics, there exist several spatial reasoning test datasets, including the Mental Rotation Tests~\cite{vandenberg1978mental}, Purdue Spatial Visualization Test (PSVT)~\cite{bodner1997purdue}, and Revised Purdue Spatial Visualization Test~\cite{yoon2011revised}. However, those human-oriented tests are not directly suitable for our purpose of developing and testing the spatial reasoning capability of an intelligent system, or a deep network. First, the amount of data in these datasets, typically less than a hundred of questions, is not enough for most deep learning methods. Second, the manual way to design and generate questions in these tests are not easily scalable. Third, many of them focus mostly on various forms of rotation reasoning tests, ignoring other spatial reasoning aspects that may be deemed either too easy to answer (e.g., to reason about the consistency between different views) or too difficult to evaluate (e.g., imagine and visualize the 3D shape or views mentally from different pose) for human, which are non-trivial for a machine.
In addition, some tests use line drawings without hidden lines (not directly visible due to occlusion), which might cause ambiguity and make it unnecessarily difficult for our purpose.

In the vision community, some Visual Question Answering (VQA) datasets that are reviewed in the next section are the closest efforts involving spatial reasoning. However, these datasets are heavily coupled with natural language processing and understanding, instead of purely focusing on spatial reasoning itself. Also, these datasets are mainly designed for visual relationship reasoning instead of spatial reasoning about geometric shapes and poses.

Therefore, we propose the SPARE3D dataset to promote the development and facilitate the evaluation of intelligent systems' spatial reasoning abilities. We use orthographic line drawings as the primary input modality for our tasks. Line drawings are widely used in engineering, representing 3D mechanical parts in computer-aided design or structures in building information models from several 2D views, with surface outlines and creases orthogonally projected onto the image plane as straight or curved lines. Compared with realistic images, line drawings are not affected by illumination and texture in rendering, providing pure, compact, and most prominent object geometry information. It is even possible to encode depth cues in a single drawing with hidden lines.

Moreover, line drawing interpretation has been extensively studied in computer vision and graphics for a few decades, leading to theories such as line labeling and region identification~\cite{macworth1973interpreting,sugihara1986machine,malik1987interpreting,varley2005frontal}, and for single-view reconstruction~\cite{ramalingam2013lifting}.
Many of these methods were trying to convert 2D line drawings to 3D models based on projective geometry theories and rule-based correspondence discovery, which is arguably different from human's seemingly instinctive and natural understanding of those drawings. We hope SPARE3D can stimulate new studies in this direction using data-driven approaches.

SPARE3D contains five spatial reasoning tasks in three categories of increasing difficulty, including view consistency reasoning, camera pose reasoning, and shape generation reasoning, as illustrated in Figure~\ref{fig_1}. The first two categories are discriminative. View consistency reasoning requires an intelligent agent to select a compatible line drawing of the same object observed from a different pose than the given drawings. The more difficult camera pose reasoning requires the agent to establish connections between drawings and their observed poses, which is similar to the aforementioned Mental Rotation Tests and PSVT.
The shape generation is the most difficult, where we test for higher-level abilities to directly generate 2D (line drawings from other views) or 3D (point clouds or meshes) representations of an object, based on the given line drawings. If an agent can solve this type of tasks accurately, then the previous two categories can be solved directly. Note that although there are other types of spatial reasoning tasks in the psychometrics literature, we focus on these three because they are some of the most fundamental ones.

In summary, our contributions are the following:
\begin{itemize}[nosep,nolistsep]
    \item To the best of our knowledge, SPARE3D is the first dataset with a series of challenging tasks to evaluate purely the spatial reasoning capability of an intelligent system, which could stimulate new data-driven research in this direction.
    \item We design a scalable method to automatically generate a large number of non-trivial testing questions and ground truth answers for training and evaluation.
    \item We design baseline deep learning methods for each task and provide a benchmark of their performance on SPARE3D, in comparison with human beings.
    \item We find that state-of-the-art convolutional networks perform almost the same as random guesses on SPARE3D, which calls for more investigations.
    \item We release the dataset and source code for data generation, baseline methods, and benchmarking.
\end{itemize}
\section{Related Works}

\begin{table}[t!]
    \label{t_mlp}
    \centering
    \resizebox{\columnwidth}{!}{%
    \begin{tabular}{l|llllll}
    \hline
    
    Dataset   & 2D & 3D & pure geometry &line drawing & reasoning \\ \hline
    
    Visual Reasoning~\cite{johnson2017clevr,bisk2018learning,yang2018dataset,chen2019touchdown,hudson2019gqa,krishna2017visual,santoro2018measuring}& \checkmark & \xmark & \xmark & \xmark & \checkmark \\ \hline
    
    Phyre~\cite{bakhtin2019phyre} & \checkmark & \xmark  &\xmark &\xmark &\checkmark \\ \hline
    
    ShapeNet~\cite{chang2015shapenet} & \checkmark & \checkmark &\xmark &\xmark &\xmark \\ \hline
    
    ScanNet~\cite{dai2017scannet} & \checkmark &\checkmark &\xmark &\xmark & \xmark  \\ \hline
    
    Line Drawing~\cite{cole2008people, cole2009well, gryaditskaya2019opensketch, QuickDrawWebsite}
    & \checkmark & \xmark & \xmark & \checkmark & \xmark \\ \hline
    
    ABC~\cite{koch2019abc} & \checkmark &\checkmark &\checkmark &\xmark &\xmark \\ \hline
    
    \textbf{SPARE3D (ours)} &\checkmark &\checkmark &\checkmark &\checkmark &\checkmark \\ \hline

    \end{tabular}%
    }
    \vspace{3pt}
    \caption{\textbf{Summary of related public datasets}. \textit{2D, 3D and line drawing} indicate the types of data in a dataset. \textit{Pure geometry} means the dataset is only focusing on geometry, without other modalities (language/semantics/physics). \textit{Reasoning} means whether a dataset is designed directly for reasoning.}
    \label{tab_1}
\end{table}

Spatial reasoning has been studied for decades in cognitive science and psychology. With the advancements of artificial intelligence (AI), researchers begin to design AI systems with visual/spatial understanding and reasoning abilities. As mentioned, classical human-centered spatial reasoning tests are not designed for AI and not readily transferable for developing spatial reasoning AI. Thus we only focus on reviewing datasets and methods related to spatial reasoning in the broad context of AI, where the main differences with SPARE3D are summarized in Table~\ref{tab_1}.

\textbf{Visual Reasoning Dataset.}
Recently there has been substantial growth in the number of visual reasoning datasets. They facilitate the development and evaluation of AI's visual and verbal reasoning ability by asking common sense questions about an image in the form of natural language~\cite{johnson2017clevr, bisk2018learning, yang2018dataset, kafle2018dvqa, chen2019touchdown, hudson2019gqa, zellers2019vcr, krishna2017visual, santoro2018measuring} (except for~\cite{bakhtin2019phyre} that focuses on physics). SPARE3D has two major differences. First, it only involves visual/spatial information of an object; therefore, natural language processing is not required. The tasks in SPARE3D is already very challenging, so decoupling them from other input modalities allows researchers to focus on spatial reasoning. Second, SPARE3D focuses on reasoning about two fundamental geometric properties: the shape of a 3D object, and the pose it was observed from, rather than the relative position, size, or other semantic information comparisons between objects.

\textbf{3D Object/Scene Dataset.}
Recent years have also seen the booming large-scale 3D datasets designed for representation learning tasks such as classification and segmentation as a way to 3D scene understanding. For example, ShapeNet~\cite{chang2015shapenet} as a 3D object dataset with rich semantic and part annotations, and ScanNet~\cite{dai2017scannet} as an RGB-D video dataset for 3D reconstruction of indoor scenes. Some of those datasets are then utilized in the visual navigation studies~\cite{zhu2017target,xiazamirhe2018gibsonenv}. Although visual navigation can be seen as involving spatial reasoning, it focuses more on a scene level goal-achieving than the object level shape and pose reasoning in SPARE3D.
In SPARE3D, we take advantage of 3D solid models from the ABC dataset~\cite{koch2019abc}, which is proposed for digital geometry processing tasks. We then generate line drawings from these CAD models as our 2D drawing sources. Note that none of these datasets are specifically designed for spatial reasoning, as in our context.

\textbf{Line Drawing Dataset.}
Interpreting line drawings has been a long-term research topic, as discussed. With the development of deep learning, the recent efforts in this direction are to understand line drawings by analyzing a large number of them. Cole~\etal~\cite{cole2008people,cole2009well} studied on how the drawings created by artists correlate with the mathematical properties of the shapes, and how people interpret hand-drawn or computer-generated drawings. OpenSketch~\cite{gryaditskaya2019opensketch} is designed to provide a wealth of information for many computer-aided design tasks.
These works, however, mainly focus on 2D line drawing interpretation and lack 3D information paired with 2D drawings. Unlike them, SPARE3D contains paired 2D-3D data, thus can facilitate an AI system to reason about 3D object information from 2D drawings, or vice versa.

\textbf{Other Related Methods.}
We also briefly discuss some machine learning methods that we believe might help tackle spatial reasoning tasks in SPARE3D in the future. 
Research about single view depth estimation, e.g., ~\cite{garg2016unsupervised, wu2019phasecam3d}, may be used to reason about the 3D object from a 2D isometric drawing (if trained on a large number of such drawings) by predicting 3D structures to rule out some less likely candidates in the questions.
Similarly, spatial reasoning ability for an intelligent agent could also be connected with neural scene representation and rendering~\cite{eslami2018neural,kato2018renderer}. For example, Eslami~\etal~\cite{eslami2018neural} introduced the Generative Query Network (GQN) that learns a scene representation as a neural network from a collection of 2D views and their poses. Indeed, when trying to solve the SPARE3D tasks, people seem to first ``render'' the shape of a 3D object in our minds and then match that with the correct answer. If such analysis-by-synthesis approaches are how we acquired the spatial reasoning ability, then those methods could lead to better performance on SPARE3D.
\section{Spatial Reasoning Tasks}

SPARE3D contains five tasks in three categories, including view consistency reasoning, camera pose reasoning, and shape generation reasoning. The first two categories contain three discriminative tasks, where all questions are similar to single-answer questions in standardized tests with only one correct and three similar but incorrect answers. The last category contains two generative tasks, where no candidate answers are given, but instead, the answer has to be generated. Next, we discuss first how we design these tasks, and then how to generate non-trivial question instances.

\subsection{Task Design}
\begin{figure}[t!]
\centering
\includegraphics[width=0.95\columnwidth]{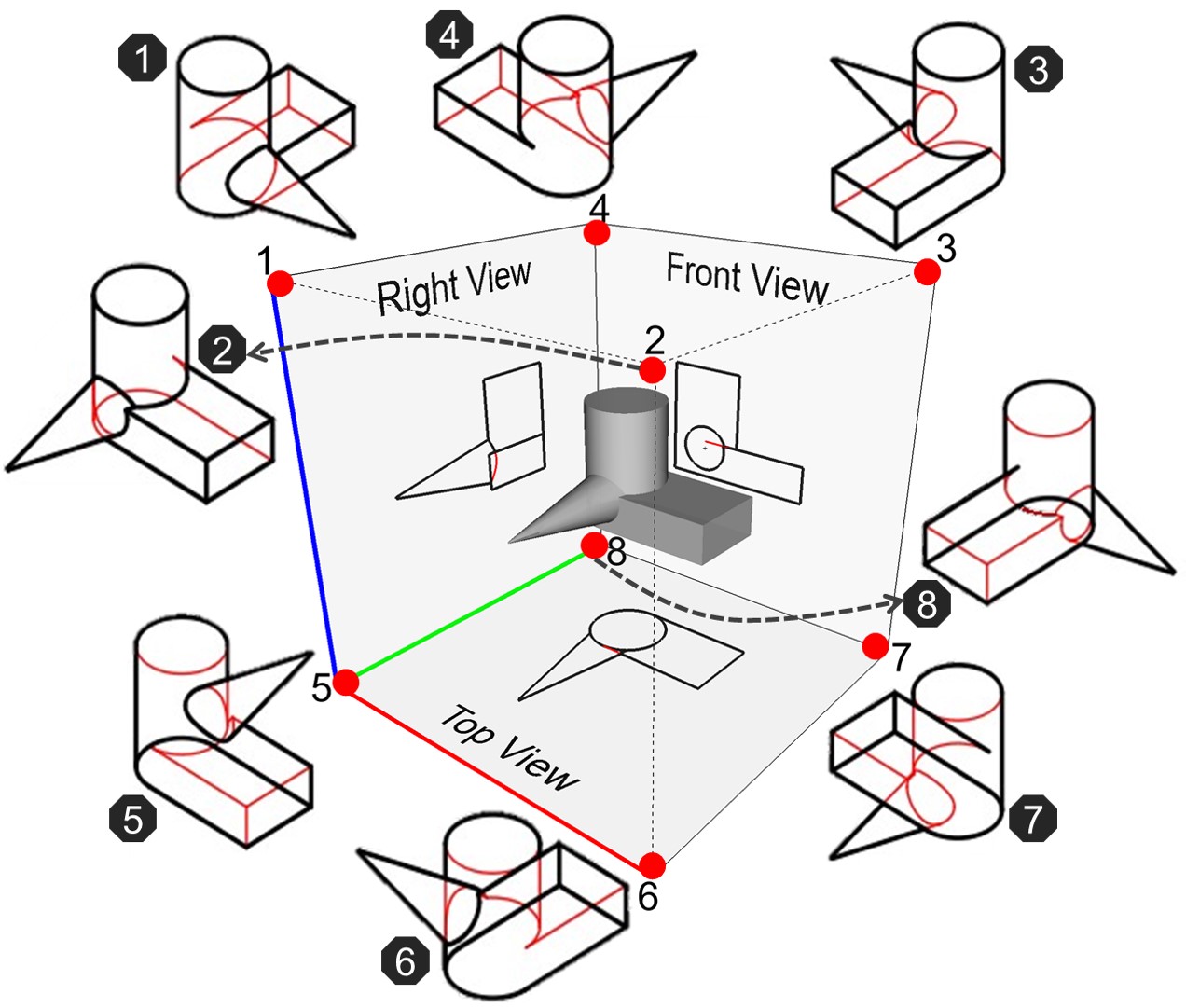}
\caption{\textbf{Illustration of the eight isometric views in SPARE3D}. Imagine a 3D object is placed in the center of a cube (grey). Each vertex of the cube represents the viewpoint of an isometric drawing, correspondingly labeled from 1 to 8.
The front/top/right (F/T/R) view's viewpoint is located on the centers of rectangles 1-5-6-2/1-2-3-4/2-6-7-3 respectively. Note that hidden lines are drawn in red. Best viewed in color.}
\label{fig_2}
\vspace{-3mm}
\end{figure}

In a SPARE3D task, an intelligent agent is given several views of orthographic line drawings of an object as the basic input for its reasoning. Without loss of generality and following conventions in engineering and psychometrics, in SPARE3D, we only focus on 11 fixed viewing poses surrounding an object: front (F), top (T), right (R), and eight isometric (I) viewing poses, as illustrated in Figure~\ref{fig_2}. Note that drawings from F, T, and R views are usually termed as three-view drawings. And an isometric view means the pairwise angles between all three projected principal axes are equal. Note that there are more than one possible isometric drawings from the same view point~\cite{yue2006spatial}, and without loss of generality, we choose the eight common ways as in Figure~\ref{fig_2}. Although geometrically equal, the F/T/R views and I views have a significant statistical difference in appearance. Because our 3D objects are mostly hand-designed by humans, many lines are axis-aligned and overlap with each other more frequently when projected to F/T/R views than I views. Therefore, I views can usually keep more information about the 3D object.

Nonetheless, it is well-known that in general, a 3D shape cannot be uniquely determined using only two corresponding views of line drawings unless three different views of line drawings are given with mild assumptions~\cite{huang1989motion}. Moreover, finding the unique solution requires methods to establish correspondences of lines and junctions across different views, which itself is non-trivial. Thus, even at least three views of line drawings are given as input in all SPARE3D tasks, it is still not straightforward to solve them.

\vspace{-3mm}
\paragraph{View Consistency Reasoning.}
A basic spatial reasoning ability should be grouping different views of the same 3D object together. In other words, an intelligent agent with spatial reasoning ability should be able to tell whether different line drawings could be depicting the same object from different viewing poses. This is the origin of the view consistency reasoning task. It is partly linked to some mental rotation tests in psychometrics, where one is asked to determine whether two views after rotation can be identical. We factor out the rotation portion to the second task category, leave only the consistency checking part, resulting in the first task below.

\begin{figure}[t!]
\centering
\vspace{15pt}%
\includegraphics[width=0.98\columnwidth]{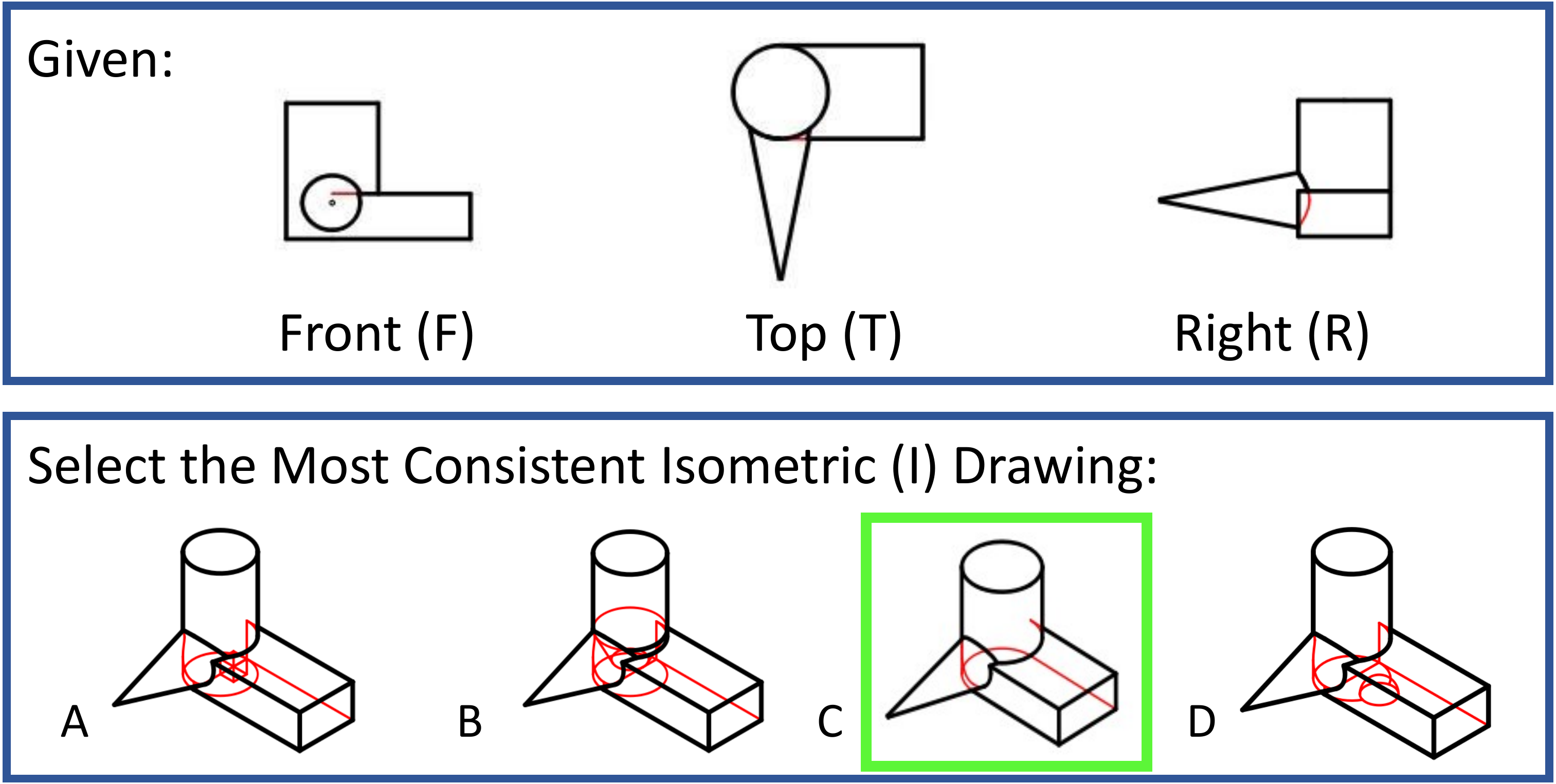}
\caption{\textbf{An example \textit{3-View to Isometric} task}. The candidate isometric views in the second row are all from pose 2. The correct answer is highlighted in green, and hidden lines are drawn in red in this and the following two figures. Best viewed in color.}
\label{fig_3}
\vspace{-5mm}
\end{figure}

\textbf{\textit{3-View to Isometric}.}
Given front, right, and top view line drawings of a 3D object, an intelligent agent is asked to select the correct isometric view drawing of the same object captured from pose 2 defined in Figure~\ref{fig_2}. We use pose 2 since it is the most common pose in conventional isometric drawings (see an example in Figure~\ref{fig_3}).

\begin{figure}[!htb]
\vspace{10pt}
\centering
\includegraphics[width=0.98\columnwidth]{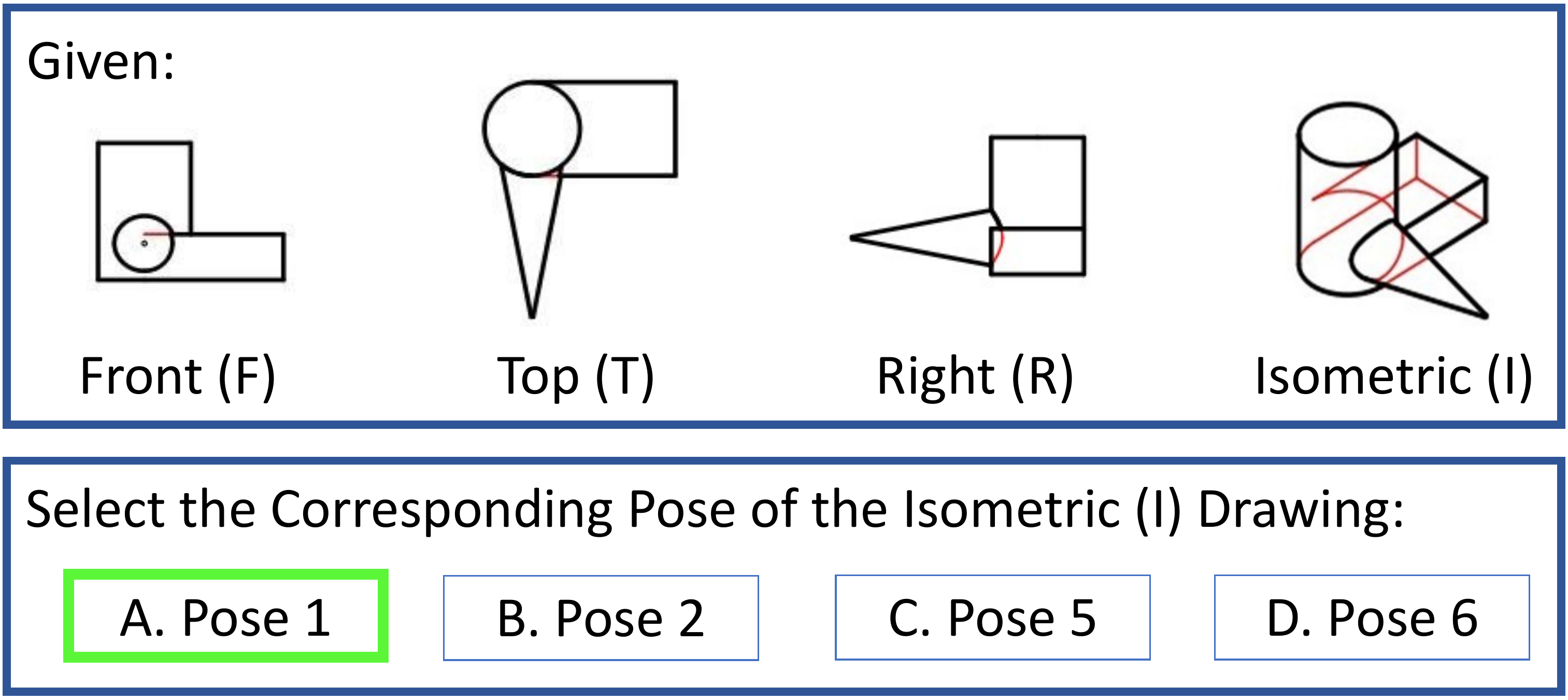}
\caption{\textbf{An example \textit{Isometric to Pose} task}.}
\label{fig_4}
\vspace{-5mm}
\end{figure}

\begin{figure}[!htb]
\centering
\vspace{10pt}%
\includegraphics[width=0.98\columnwidth]{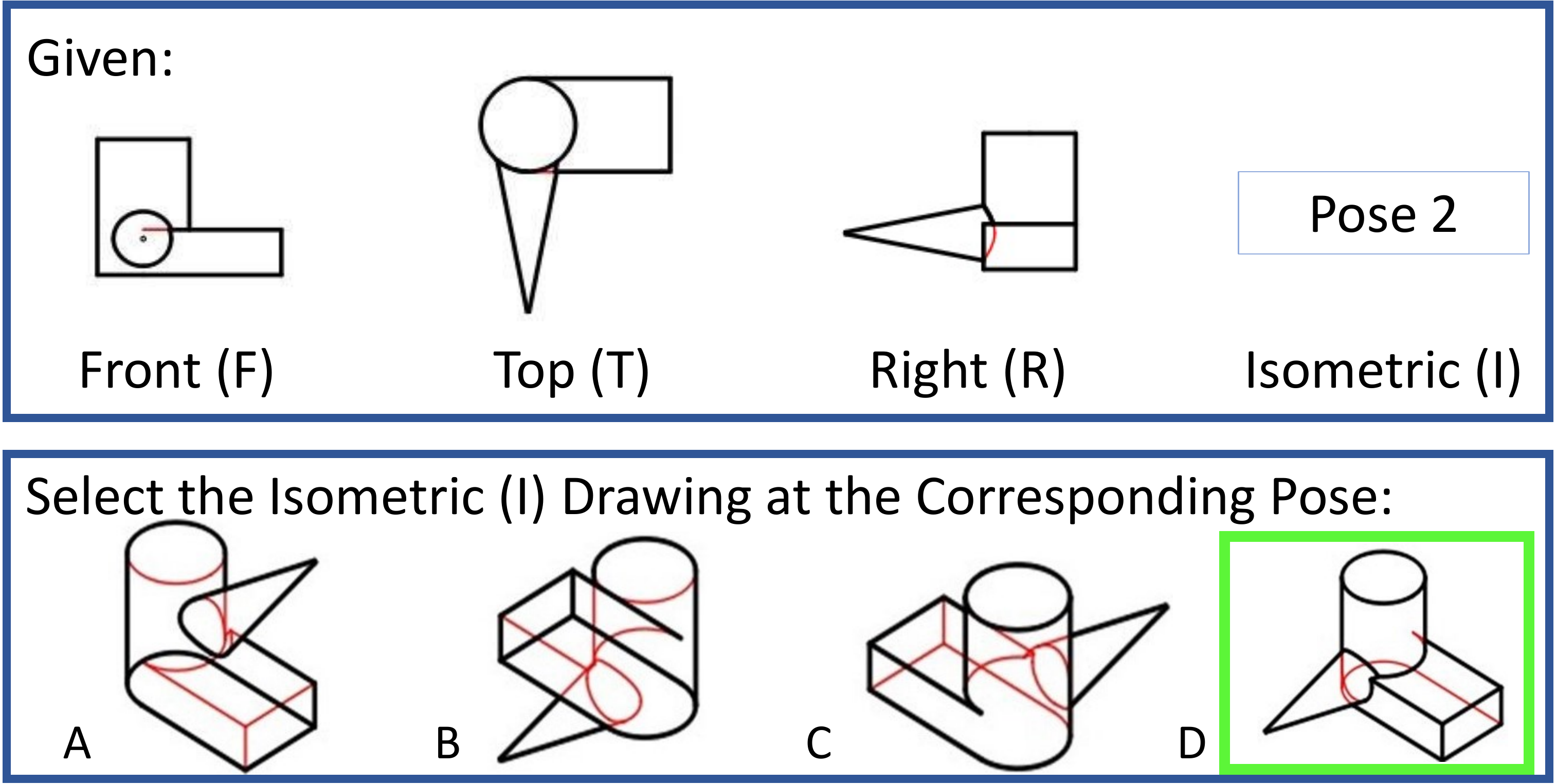}
\caption{\textbf{An example \textit{Pose to Isometric} task}.}
\label{fig_5}
\vspace{-3mm}
\end{figure}

\vspace{-3mm}
\paragraph{Camera Pose Reasoning.}
Mental rotation ability is an important spatial reasoning ability that an intelligent agent should have. By thoroughly understanding the shape of a 3D object from several 2D drawings, the agent should be able to establish correspondences between a 2D drawing of the object and its viewing pose. This leads to the following two tasks (see examples in Figure~\ref{fig_4} and ~\ref{fig_5}).

\textbf{\textit{Isometric to Pose}.}
Given the front, right, top view and a specific isometric view line drawings, an intelligent agent is asked to determine the camera pose of that isometric drawing. We consider only four poses, 1/2/5/6, for isometric drawings in this task. 

\textbf{\textit{Pose to Isometric}.}
As the ``inverse'' process of the previous task, this task asks an intelligent agent to select the correct isometric drawing from a given viewing pose in addition to the given three-view drawings. To further increase the difficulty, we consider all the eight isometric poses. 

\vspace{-3mm}
\paragraph{Shape Generation Reasoning.}
Generating the 2D or 3D shape of an object from several 2D drawings is a fundamental aspect of spatial reasoning as suggested by its definition. We believe it is such a top level capability that if possessed can solve most of the spatial reasoning tasks: by extracting spatial information contained in 2D drawings and reconstruct 3D shapes, it could enable the agent to answer view consistency, or camera pose reasoning questions by searching for possible solutions and eliminate less possible ones. Therefore we design this category of tasks. Different from the previous discriminative tasks, where the solution space is discrete and finite, the following two tasks in this category do not provide any candidate solutions, thus being the most challenging among all.

\textbf{\textit{Isometric View Generation}}.
An intelligent agent is provided with front, right, and top view drawings and asked to generate the corresponding isometric view drawing from pose 2 (without loss of generality). 

\textbf{\textit{Point Cloud Generation}}.
Given the same input as in the previous task, the agent is asked to generate a complete 3D model represented as a point cloud.

\subsection{Task Generation}

\noindent\textbf{3D Object Repositories.}
To automatically generate different instances of the above designed tasks, we create two 3D object repositories: \textit{SPARE3D-ABC}, where 10,369 3D CAD objects are sampled from the ABC dataset~\cite{koch2019abc}, and \textit{SPARE3D-CSG}, where 11,149 3D constructive solid geometry (CSG) objects of simple 3D primitives are randomly generated. Given a 3D model repository, we use PythonOCC~\cite{pythonocc}, a Python wrapper for the CAD-Kernel OpenCASCADE, to generate front/top/right/isometric view drawings from the 11 fixed poses. This directly provides us datasets for the shape generation reasoning tasks. We generate all tasks independently on each repository. The baseline results of all tasks run on both the SPARE3D-ABC and SPARE3D-CSG models are shown and discussed in the benchmark result section. 

To use 3D objects from the ABC dataset, we remove all duplicates by choose objects with unique hash values of their front view image files. We also skip some objects whose STEP-format file size exceed a certain limit to reduce the computing load. Note that there are many objects in the ABC dataset whose corresponding front, top, or right view drawings contains only a small point. We exclude all these objects to ensure 2D drawings in our dataset cover a reasonably large image area so as to be legible for an intelligent agent even after downsampling. \textcolor{blue}{
We also remove all the questions whose line drawings contain extremely thick or thin lines, due to the inconsistent scaling.
Furthermore, we remove questions with ambiguous answers, such as the same line drawing in different candidate answers for the \textit{3-View to Isometric} task, or questions generated from symmetric CAD models for the \textit{Isometric to Pose} task.
}

\noindent\textbf{Avoiding Data Bias.}
Given a large number of line drawings and corresponding 3D objects, cares must be taken when generating instances of the above spatial reasoning tasks.
An important consideration is to avoid data bias, which could be undesirably exploited by deep networks to ``solve'' a task from irrelevant statistical patterns rather than really possessing the corresponding spatial reasoning ability, leading to trivial solutions. Therefore, we make sure that all images in the dataset have the same size, resolution, and scale. We also ensure that our correct and incorrect answers are uniformly distributed in the solution space, respectively. Besides, we ensure each drawing only appears once across all tasks, either in questions or in answers, to avoid memorization possibilities.

The biggest challenge of avoiding data bias is to automatically generate non-trivial \textit{incorrect} candidate answers for the view consistency reasoning task. If incorrect answers are just randomly picked from a different object's line drawings, according to our experiments, a deep network can easily exploit some local appearance similarities between views to achieve high testing performance in this task.
Therefore, we further process 3D objects for this task. We first cut a 3D object by some basic primitive shapes like sphere, cube, cone and torus for four times to get four cut objects. Then we randomly choose one of the four objects to generate F, T, R, and I drawings as question and correct answer drawings. And the three I drawings from the remaining three cut objects are used as the wrong candidate answers. We record the index of the correct isometric drawing as the ground truth label for supervised learning.
We prepare 5,000 question instances in total for the \textit{3-View to Isometric} task. We perform an $8:1:1$ train/validation/test dataset split. 
We use almost the same settings to generate camera pose reasoning tasks except that no 3D object cutting is needed.

\section{Baseline Methods}

We try to establish a reasonable benchmark for SPARE3D tasks using the most suitable baseline methods that we could find in the literature. \textit{3-View to Isometric} and \textit{Pose to Isometric} are formulated as either binary classification or metric learning, \textit{Isometric to Pose} as multi-class classification, \textit{Isometric View Generation} as conditional image generation, \textit{Point Cloud Generation} as multi-view image to point cloud translation.
For each task, images are encoded by a convolutional neural network (CNN) as fixed dimensional feature vectors, and a camera pose is represented by a one-hot encoding because of the small number of fixed poses in each task. Note that our dataset offers both vector (SVG) and raster (PNG) representations of line drawings. Raster files can be readily consumed by CNN, while vector files offer more possibilities such as point cloud or graph neural networks. Currently, we focus only on raster files because of the relative maturity of CNN. We will benchmark more networks suitable for vector data in the future.

For the backbone network architectures, we select ResNet-50~\cite{he2016deep} and VGGNet-16~\cite{simonyan2014very} to model the image feature extraction function, due to their proved performance in various visual learning tasks. We also select BagNet~\cite{brendel2019approximating}, which shows surprisingly high performance on ImageNet~\cite{deng2009imagenet} even with a limited receptive field. Detailed baseline formulation and network architectures are explained in the supplementary material.

\textbf{Human Performance.}
We design a crowd-sourcing website to collect human performance for \textit{3-View to Isometric}, \textit{Pose to Isometric}, and \textit{Pose to Isometric} reasoning tasks.
Two types of human performance are recorded: untrained vs. trained. In the untrained type, we distributed the website on certain NYU engineering classes and social media platforms and had no control of the participants. We collected testing results from more than 100 untrained people, with each of them answering four randomly selected questions in each task. We report their average performance as the first human baseline. The second type comes from five randomly selected engineering master students. Each of them is trained by us for about 30 minutes using questions from the training set, and then answers 100 questions for each task with limited time. We report their max performance as the second human baseline.

\begin{figure*}[b!]
\vspace{-2mm}
\centering

\includegraphics[width=0.15\textwidth,height=0.035\textheight ]{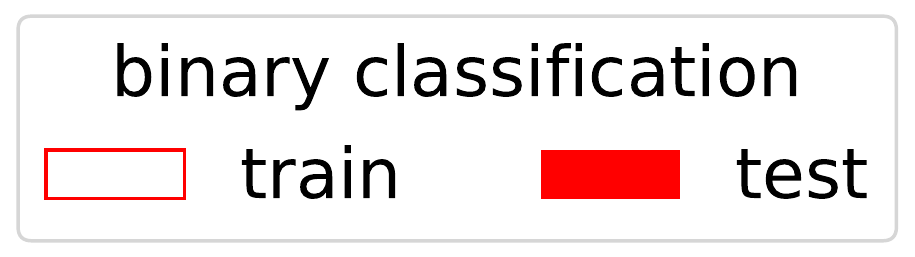}
\includegraphics[width=0.15\textwidth,height=0.035\textheight ]{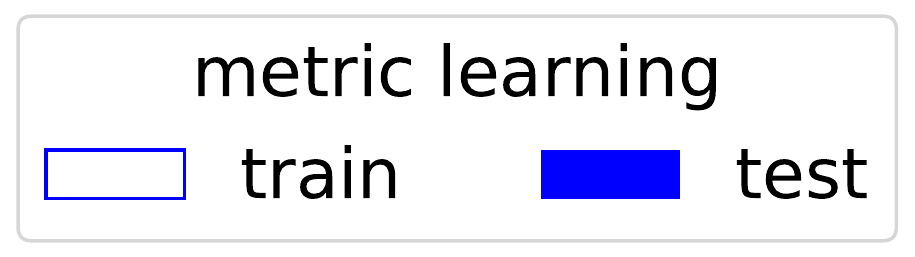}
\includegraphics[width=0.15\textwidth,height=0.035\textheight ]{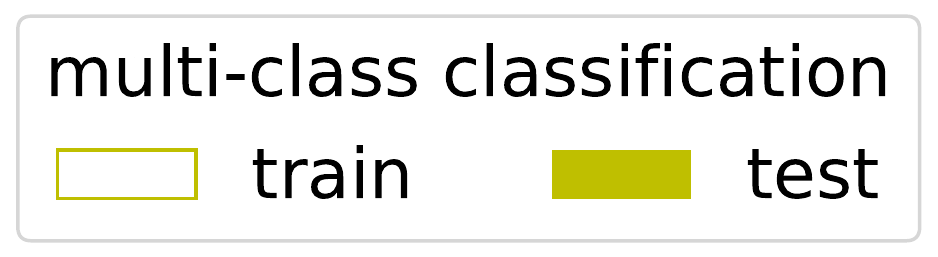}
\includegraphics[width=0.15\textwidth,height=0.035\textheight ]{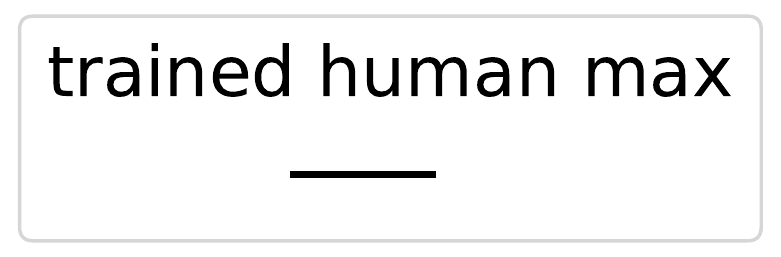}
\includegraphics[width=0.18\textwidth,height=0.035\textheight ]{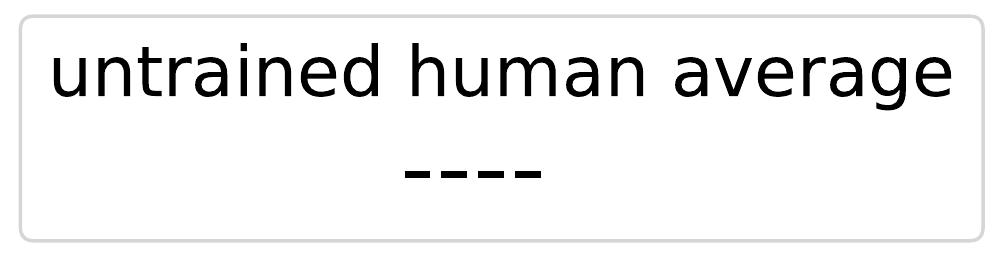}\\\vspace{+3mm}

\includegraphics[width=0.31\textwidth,trim=.6cm .3cm .8cm .0cm,clip, ]{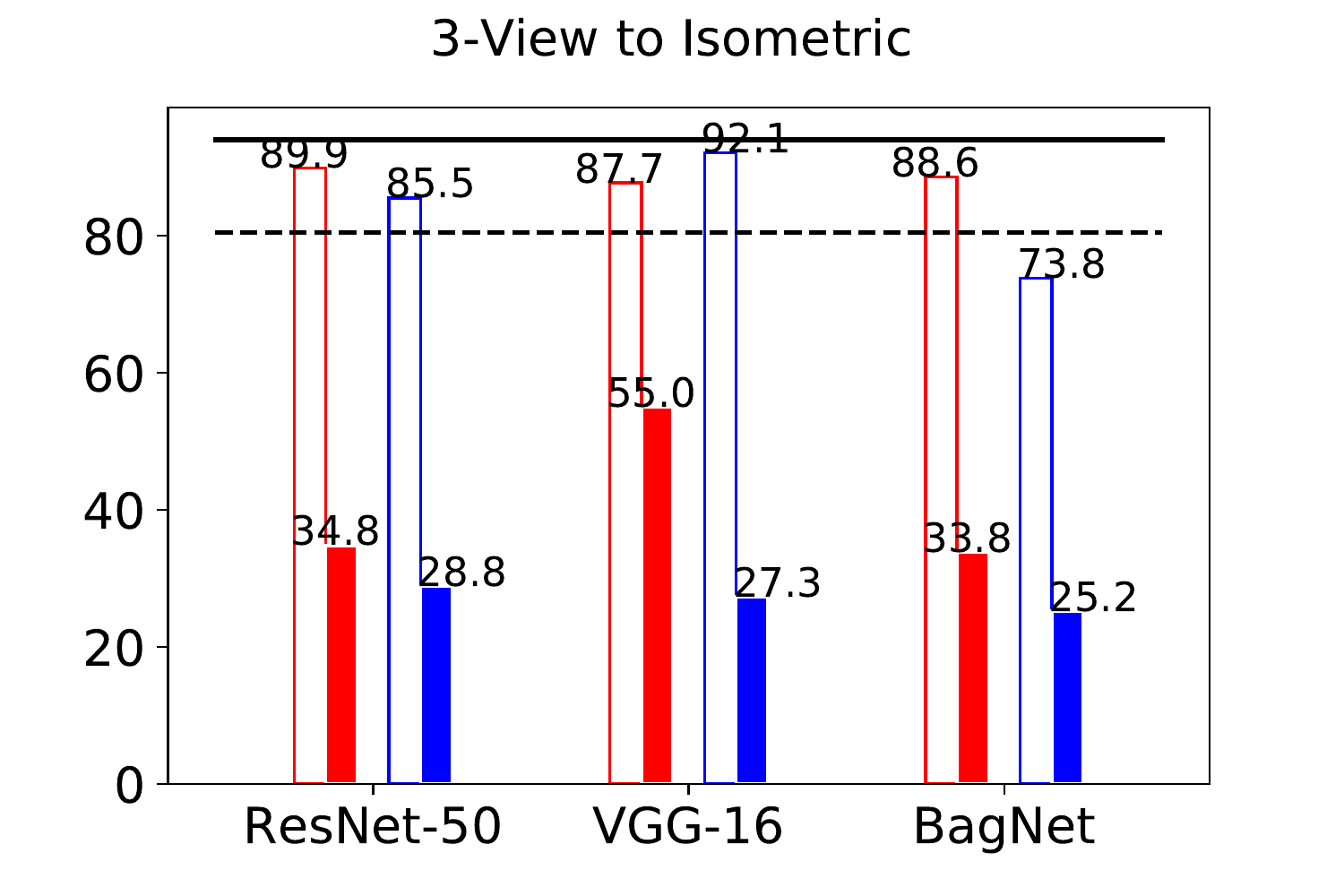}
\includegraphics[width=0.31\textwidth,trim=.6cm .3cm .8cm .0cm,clip, ]{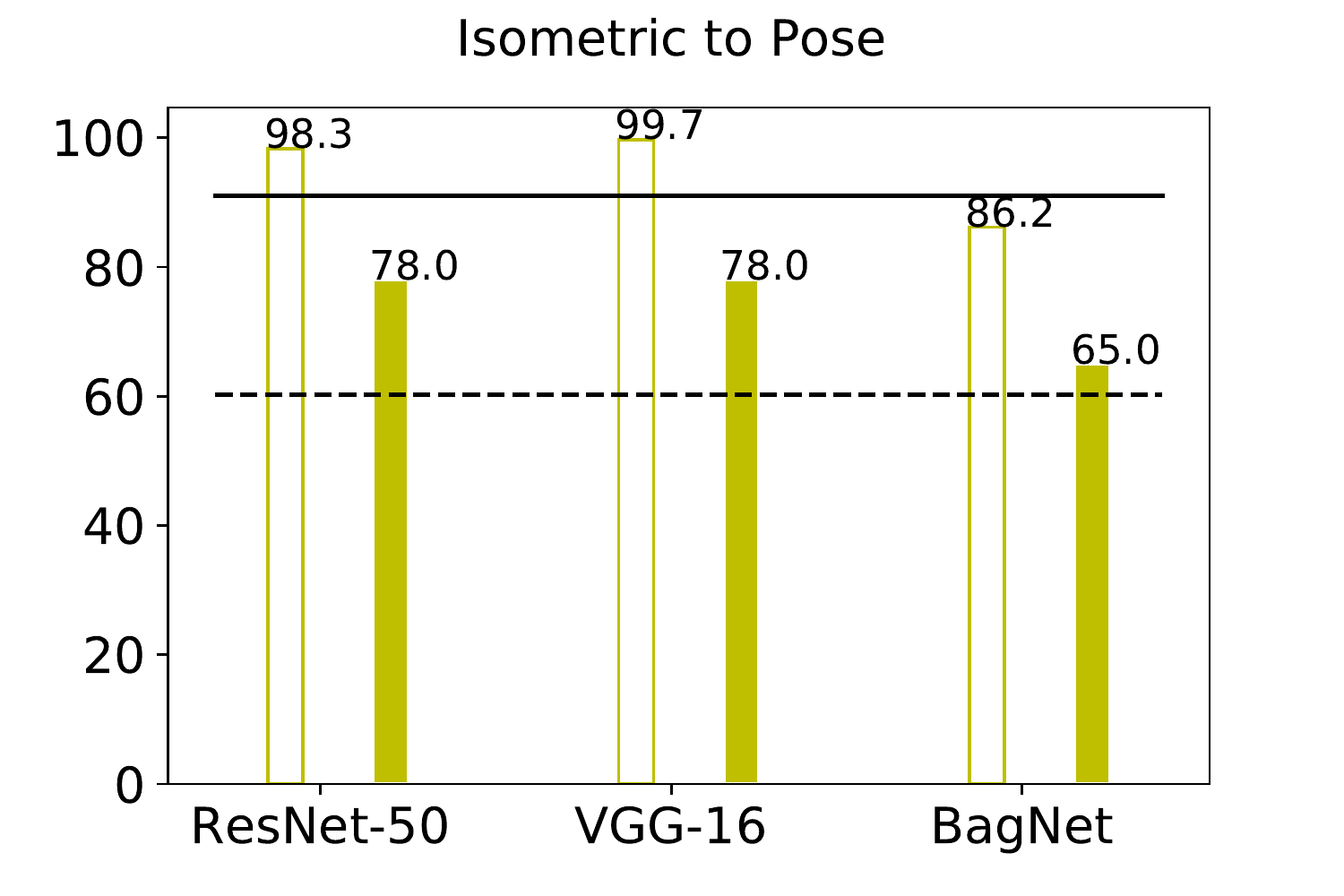}
\includegraphics[width=0.31\textwidth,trim=.6cm .3cm .8cm .0cm,clip, ]{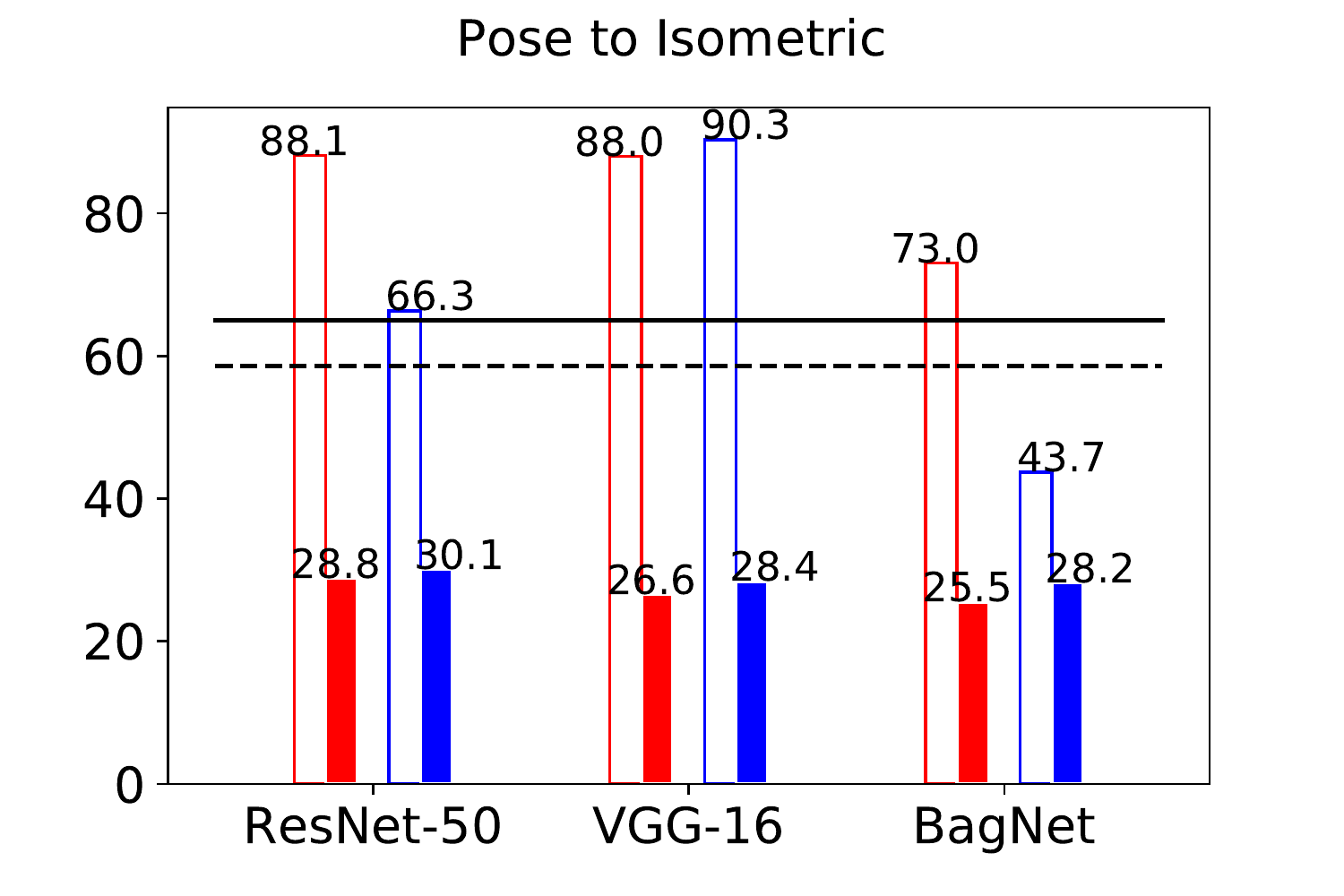}

\includegraphics[width=0.31\textwidth,trim=.6cm .3cm .8cm .0cm,clip, ]{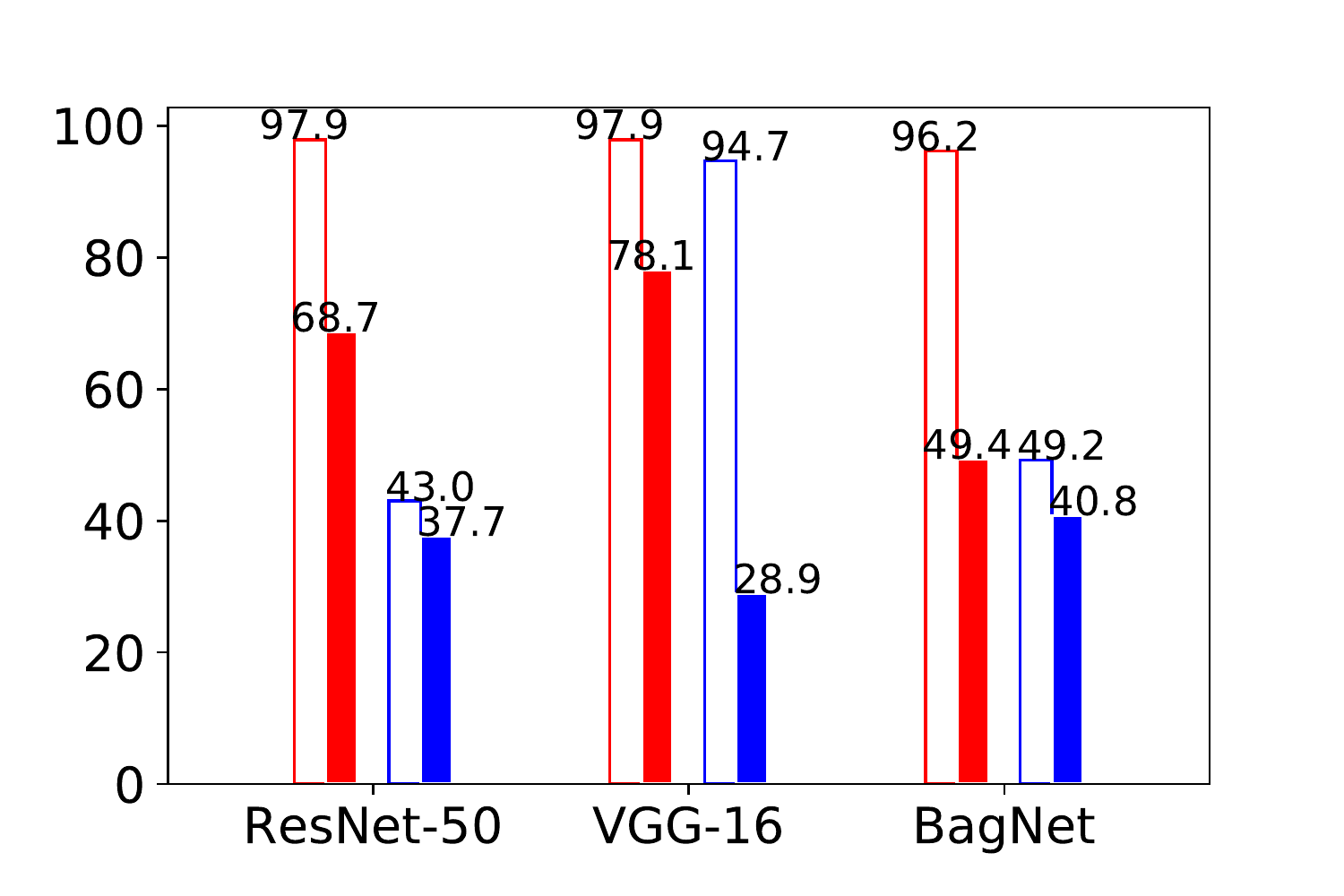}
\includegraphics[width=0.31\textwidth,trim=.6cm .3cm .8cm .0cm,clip, ]{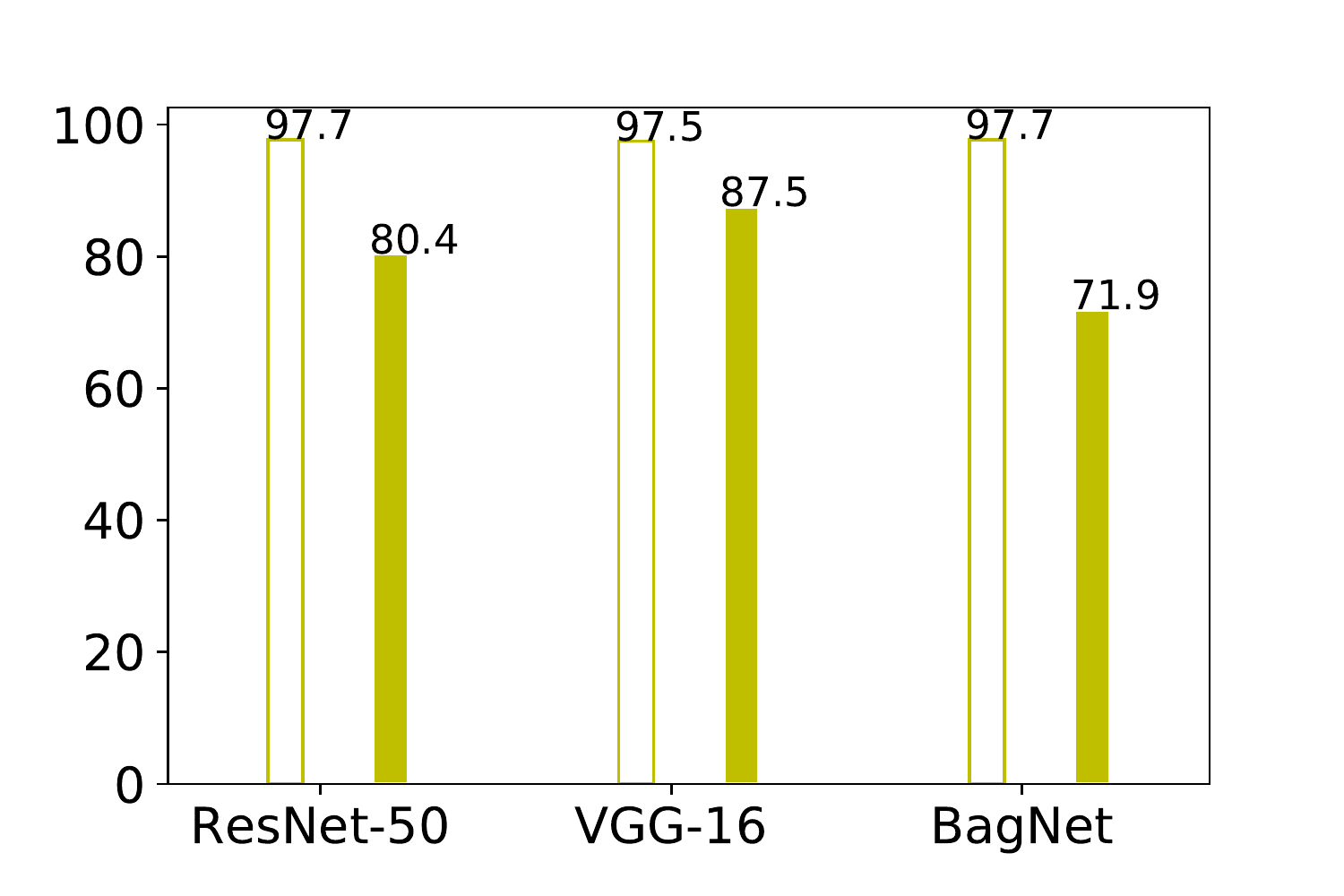}
\includegraphics[width=0.31\textwidth,trim=.6cm .3cm .8cm .0cm,clip, ]{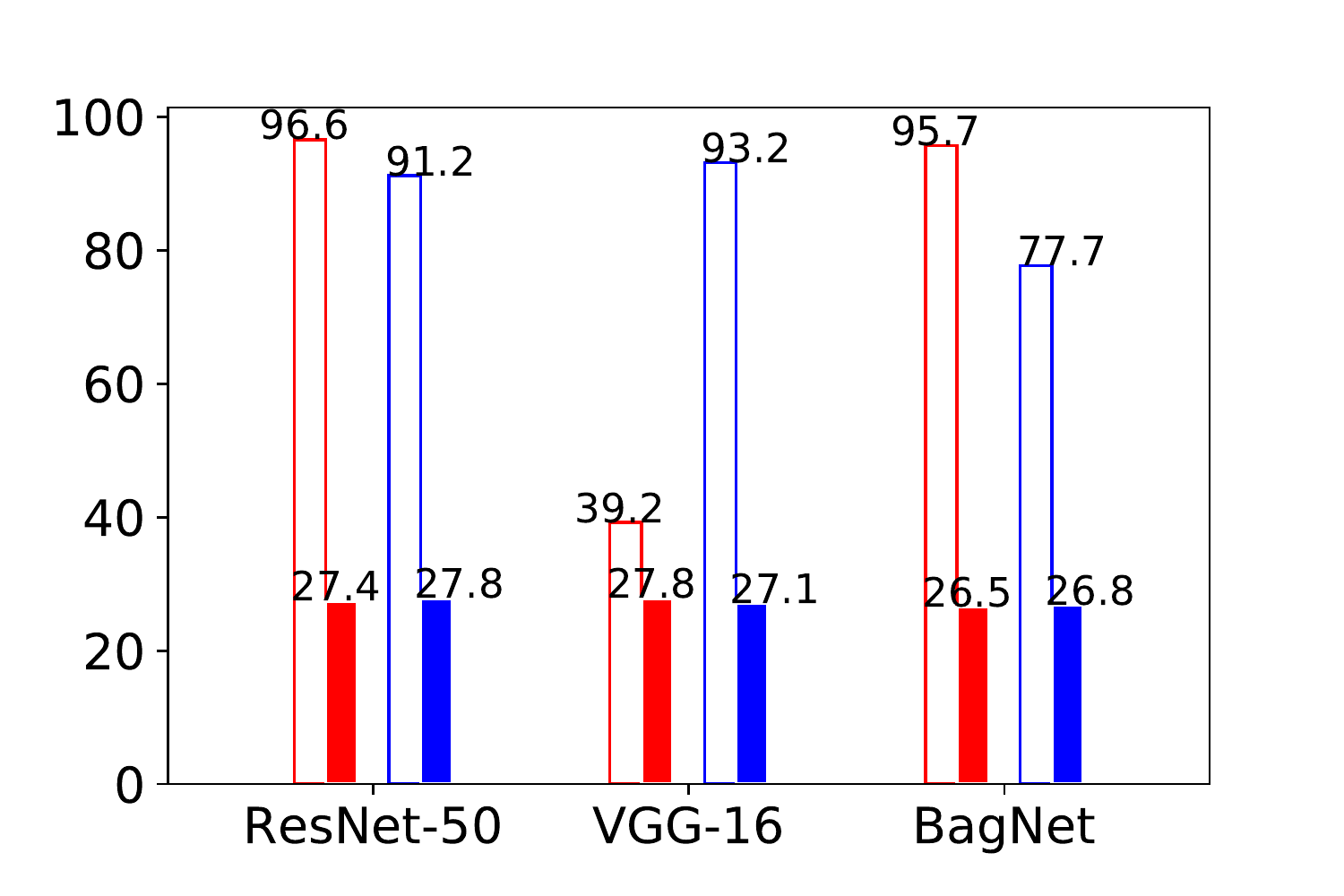}
\caption{\textbf{SPARE3D benchmark results} of baseline methods and human performance on the first three tasks on SPARE3D-ABC (top) and SPARE3D-CSG (bottom). The average untrained human performance results for \textit{3-View to Isometric}, \textit{Isometric to Pose}, and \textit{Pose to Isometric} are $80.5\%$, $60.2\%$, and $58.6\%$ respectively. The max trained human performance results for these three tasks are $94.0\%$, $91.0\%$, and $65.0\%$ respectively.}
\label{fig_6}
\end{figure*}

\section{Benchmark Results}

All our baselines were implemented using PyTorch~\cite{PyTorch:NIPS19}, and run on NVIDIA GeForce GTX 1080 Ti GPU. The results for the first three tasks are summarized in Figure~\ref{fig_6}.

\textbf{\textit{3-View to Isometric}.}
In Figure~\ref{fig_6} top left, except for the VGG-16 binary classification, all other results on SPARE3D-ABC reveal that these networks failed to obtain enough spatial reasoning ability from the supervision to solve the problems, with their performance on testing dataset close to random selection. An interesting observation is that many baseline methods achieved high training accuracy, indicating severe over-fittings.
An unexpected result is that VGG-16 binary classification achieves higher accuracy than ResNet-50 on the testing dataset (although still low), while ResNet has been repeatedly shown to surpass VGG networks in many visual learning tasks.
Compare the two images in the first column of Figure~\ref{fig_6}, the baseline performance on SPARE3D-CSG data is better than SPARE3D-ABC. We believe this is because objects in the SPARE3D-CSG repository are geometrically simpler in terms of the basic primitives of objects.

\textbf{\textit{Isometric to Pose}.}
The multi-class classification results on SPARE3D-ABC are shown in Figure~\ref{fig_6} top middle. For ResNet-50 and VGG-16, the testing accuracy is \textcolor{blue}{$78\%$, whereas BagNet's testing accuracy is $65\%$}. As for SPARE3D-CSG, we obtain \textcolor{blue}{higher accuracy than SPARE3D-ABC for all cases}, as shown in Figure~\ref{fig_6} bottom middle\textcolor{blue}{, due to the same reason as in the previous task}. 

This is surprising.
\textcolor{blue}{All the three networks achieved higher accuracy than average untrained human performance}, while before experiments, we hypothesized that none of the baselines could achieve human-level performance. \textcolor{blue}{In fact, the camera pose related tasks are often viewed more difficult to solve by our volunteers.}
This result gives us more confidence in learning-based methods for addressing these spatial reasoning tasks. 

\textbf{\textit{Pose to Isometric}.}
In Figure~\ref{fig_6} top row right, all baseline methods perform poorly, with the highest testing accuracy $30.1\%$ on ResNet-50 for metric learning and average accuracy around $27.5\%$ for other baseline methods. Similar results are obtained on SPARE3D-CSG.

Moreover, we notice that the accuracy of BagNet is almost always lower than that of ResNet-50 and VGG-16 in all tasks. It could be due to the smaller receptive field in BagNet than the other two, which constrains BagNet to exploit only local rather than global information. This indicates the SPARE3D tasks are more challenging and require higher-level information processing than ImageNet tasks, which can be solved surprisingly well by BagNet.

\textbf{\textit{Human performance}.}
In Figure~\ref{fig_6}, the untrained or trained human performance is better than most baseline methods for the same task. It reveals that most state-of-the-art networks are far from achieving the same spatial reasoning ability as humans have on SPARE3D. 

\begin{figure*}[t!]
\centering
\includegraphics[width=\columnwidth]{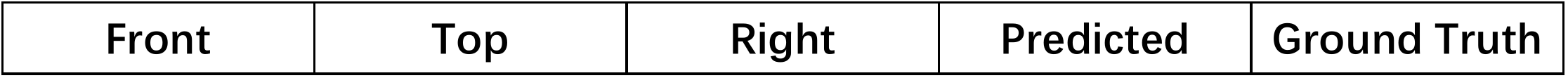}
\includegraphics[width=\columnwidth]{figs/fig_6_title/title.pdf}

\includegraphics[width=\columnwidth]{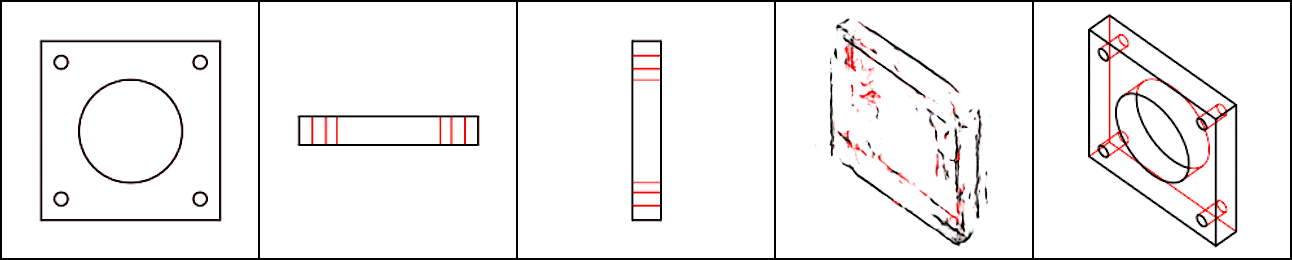}
\includegraphics[width=\columnwidth]{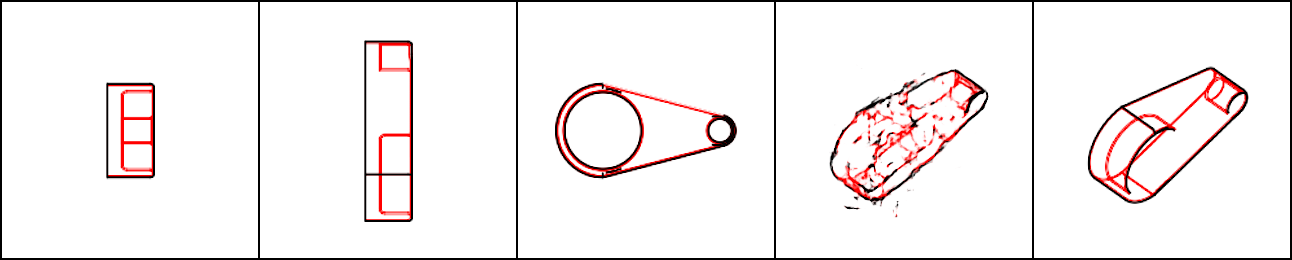}

\includegraphics[width=\columnwidth]{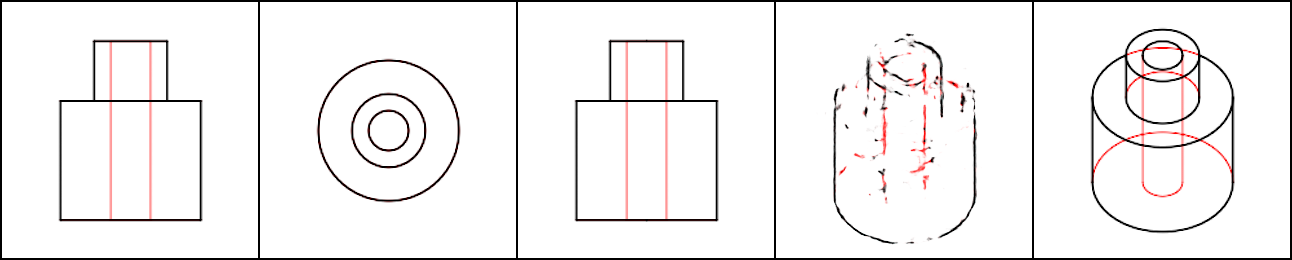}
\includegraphics[width=\columnwidth]{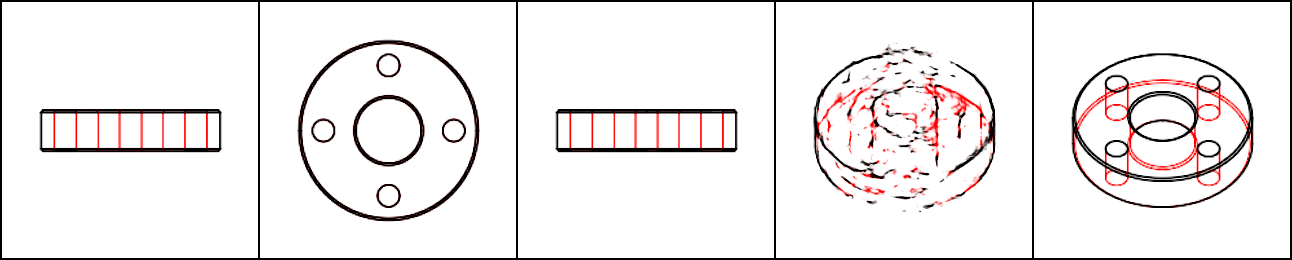}

\includegraphics[width=\columnwidth]{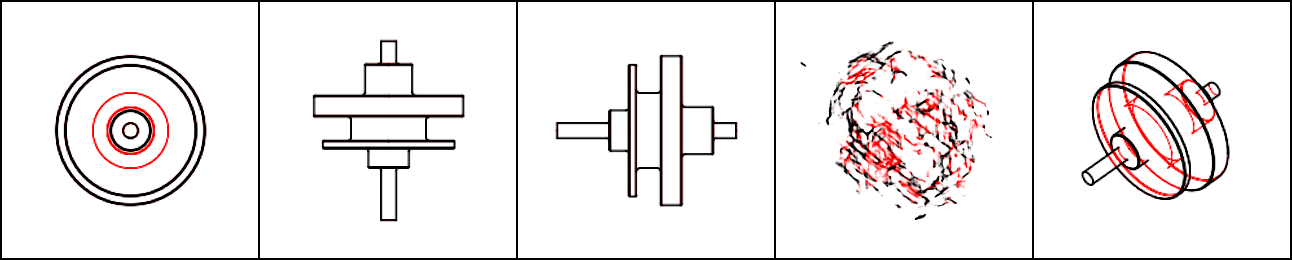}
\includegraphics[width=\columnwidth]{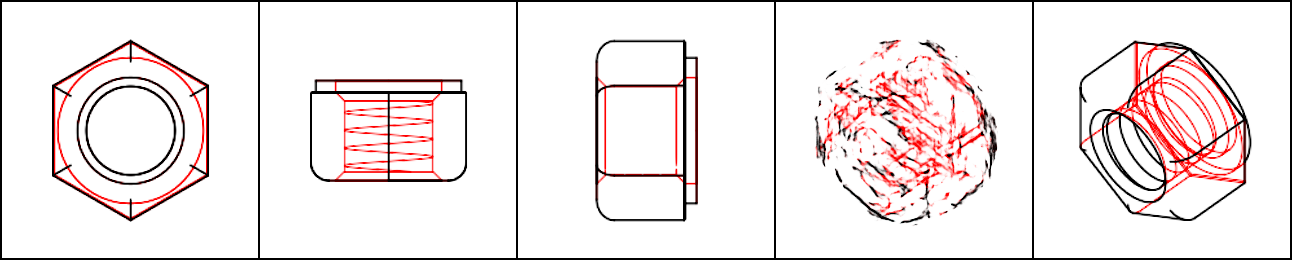}

\includegraphics[width=\columnwidth]{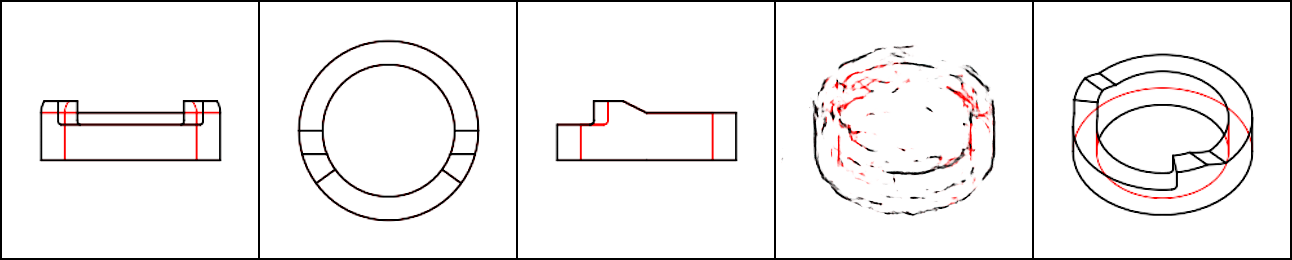}
\includegraphics[width=\columnwidth]{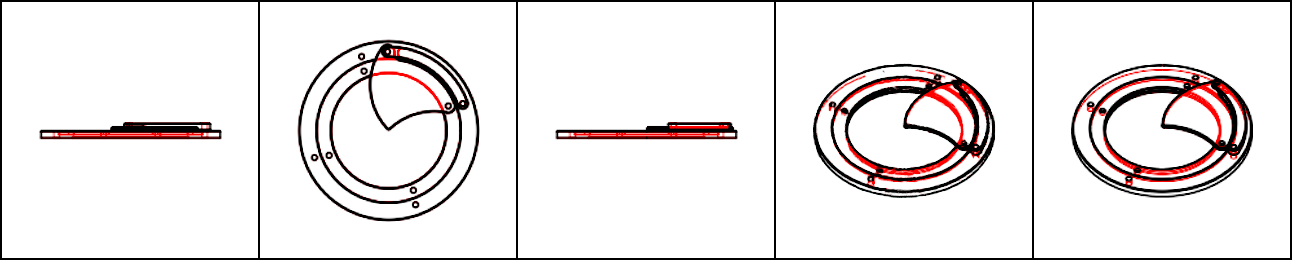}

\caption{\textbf{\textit{Isometric View Generation} testing examples}. The fourth column is the generated I drawing from the first three columns as input, and the last column is the ground truth. The baseline methods show reasonable results, yet not precise enough for solving the previous discriminative reasoning tasks. A surprisingly good result is the last one, possibly due to its near-planar structure.}
\label{fig_7}
\vspace{-3mm}
\end{figure*}

\textbf{\textit{Isometric View Generation}.}
In Figure~\ref{fig_7}, the generated results are still very coarse, although reasonable and better than our expectation given the poor performance of CNN in previous tasks. Using the generated isometric drawing to select the most similar answers (in terms of the pixel-level L2 distance) in the \textit{3-View to Isometric} task leads to a $19.8\%$ testing accuracy.  This reveals that using Pix2Pix~\cite{isola2017image} in a naive way can have reasonable generation performance, but cannot yet generate detailed and correct isometric drawings for solving the reasoning task. Therefore, new architectures for this task are still needed in the future.

\begin{figure*}[t!]
\centering

\includegraphics[width=\columnwidth]{figs/fig_6_title/title.pdf}
\includegraphics[width=\columnwidth]{figs/fig_6_title/title.pdf}

\includegraphics[width=\columnwidth]{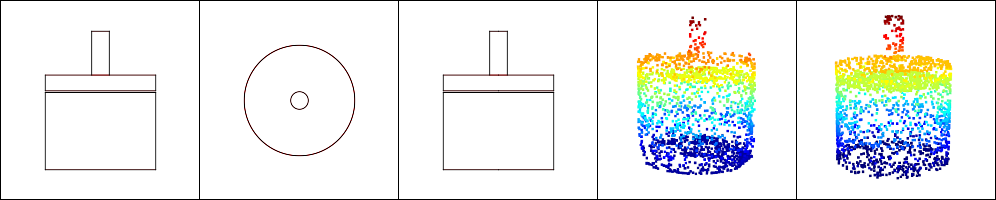}
\includegraphics[width=\columnwidth]{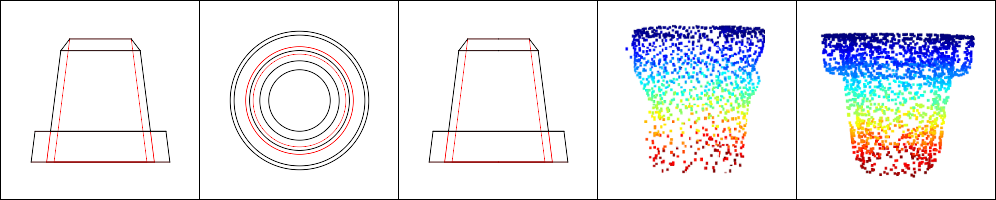}
\includegraphics[width=\columnwidth]{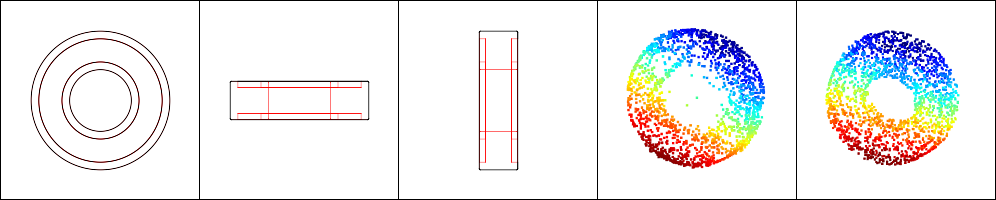}
\includegraphics[width=\columnwidth]{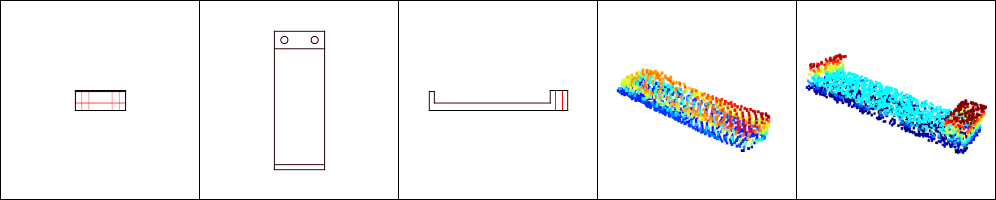}
\includegraphics[width=\columnwidth]{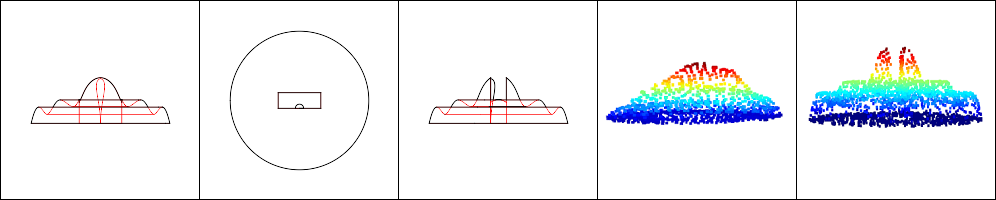}
\includegraphics[width=\columnwidth]{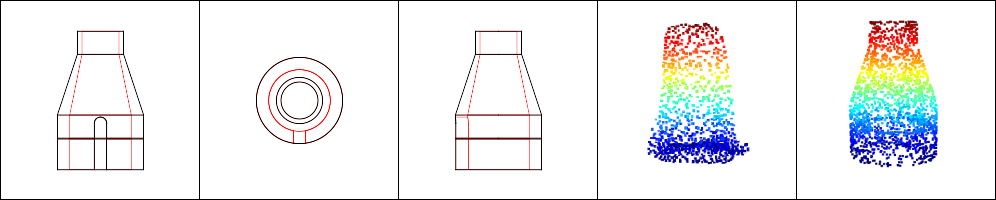}
\includegraphics[width=\columnwidth]{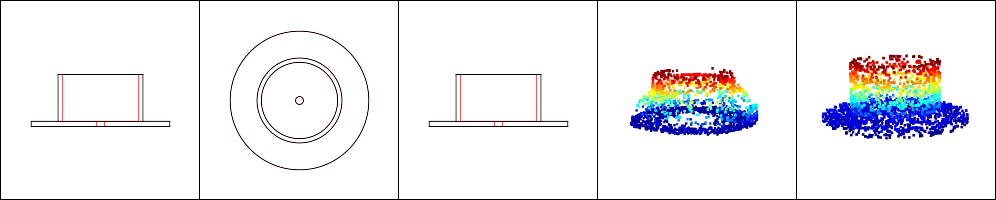}
\includegraphics[width=\columnwidth]{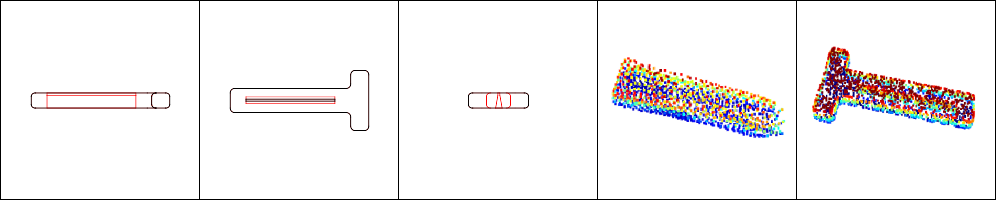}

\caption{\textbf{\textit{Point Cloud Generation} testing examples}. The left columns displays AtlasNet~\cite{groueix2018atlasnet} results and the right shows FoldingNet~\cite{yang2018foldingnet} results. The AtlasNet performs slightly better than FoldingNet in terms of details, but none of them are good enough for analysis-by-synthesis reasoning in previous discriminative tasks.
}
\label{fig_8}
\vspace{-3mm}
\end{figure*}

\textbf{\textit{Point Cloud Generation}.}
In Figure~\ref{fig_8}, the point cloud generation results are also reasonable yet unsatisfactory: the overall shape is generated correctly, while detailed features are often omitted. One possible reason is that the point cloud decoding network is not powerful enough, or the encoding CNN lacks the ability to extract the spatial information from three-view drawings. Therefore the current network baseline cannot be used to reason about complicated structures by generating them. Also, just concatenating F, R, T drawings as the input of the network is a simple yet naive way, and more effective methods are needed to synthesis these 3D objects more reasonably.

\noindent\textbf{Why baseline performance on SPARE3D is low?}
In Figure~\ref{fig_6}, except for the binary classification in \textit{3-View to Isometric} task of VGG-16 achieve \textcolor{blue}{$55\%$}, and multi-class classification in \textit{Isometric to Pose} task of ResNet-50, VGG-16, BagNet achieve \textcolor{blue}{$78\%$, $78\%$, and $65\%$} testing accuracy, all other results are close to random selection. Three following challenges in SPARE3D may cause the low performance. 

\textit{Non-categorical dataset}. SPARE3D is different from many existing datasets that contain objects from a limited number of semantic classes. Without strong shape similarities among objects in SPARE3D, it becomes significantly more difficult for networks to ``trivially'' exploit local visual patterns for ``memorization'', and forces future solutions to tackle the reasoning challenge instead of resorting to statistical correlation. We believe this unique feature is ignored by the community but necessary for moving forward towards human-level performance: people can solve our tasks without category information.  

\textit{Line drawing images}. Unlike many other datasets based on textured images or rough sketches~\cite{su2015render,delanoy20183d}, SPARE3D uses line drawings that are sparse and geometrically concise, making them closer to symbolic representations for reasoning that is difficult for existing CNN.

\textit{Reasoning is not retrieval}.
In some baselines, we use metric learning, which is common for image retrieval~\cite{funkhouser2003search} that searches for an image in a \textit{fixed} large database. But it does not fit for SPARE3D, where each question gives only four candidate answers that also \textit{vary across questions}.

\noindent\textbf{Would self-supervised 2D/3D information help?}
In the three discriminative tasks, we only use 2D information to train our baselines. One might be wondering whether using more 3D information would significantly improve the performance, as shown in~\cite{olszewski2019transformable,su2015render,sitzmann2019deepvoxels}. Although results in the two generation tasks, which is not worse than voxel reconstruction in~\cite{delanoy20183d}, have indicated that naively generated coarse shape does not help, it is still valid to ask whether we can use 2D/3D shape information implicitly via self-supervision. This leads to the following two experiments.

\textit{Pretrained Pix2Pix}.
As mentioned in \textit{Isometric View Generation}, we use the trained Pix2Pix model to generate the I drawing from the given F/T/R drawings in \textit{3-View to Isometric} questions. Instead of naively using this generated I drawing with L2 distance, which leads to a 19.8\% testing accuracy, now we train an additional CNN to select answers in a learned feature space (instead of pixel space). This CNN is similar to the binary classification network for \textit{3-View to Isometric} but takes as input the concatenation of the answer and the generated images. The new accuracy raises to 37.6\% yet is still very low. 

\textit{Pretrained FoldingNet}.
In \textit{Point Cloud Generation}, we trained a CNN encoder via 2D-3D self-supervision. Now we use this encoder as a warm start to initialize the ResNet-50 models of metric learning for \textit{3-View to Isometric} and \textit{Pose to Isometric} tasks. As shown in Table~\ref{tab_2}, accuracy has no significant increase and is still close to random selection. So our naive way of using 3D information does not work well, and further design is needed.

\begin{table}[t!]
    \centering
    \resizebox{\columnwidth}{!}{%
    \begin{tabular}{l|llllll}
    \hline
    Task   & 3-View to Isometric & Pose to Isometric \\ 
     \hline
     
    Without FoldingNet & $85.5\%$/$28.8\%$ & $66.3\%$/$30.1\%$  \\
     \hline
     
    With FoldingNet & $81.0\%$/$30.4\%$ & $87.5\%$/$27.2\%$ \\
     \hline

    \end{tabular}%
    }
    \vspace{3pt}
    \caption{\textbf{Effect of self-supervised 3D information in training}. The first row is the training/testing accuracy by random CNN initialization. The last row is using CNN initialized from a pretrained 2D-to-3D FoldingNet.}
    \vspace{-3mm}
    \label{tab_2}
\end{table}

\section{Conclusion}
SPARE3D is designed for the development and evaluation of AI's spatial reasoning abilities. Our baseline results show that some state-of-the-art deep learning methods cannot achieve good performance on SPARE3D. We believe this reveals important research gaps and motivates new problem formulations, architectures, or learning paradigms to improve an intelligent agent's spatial reasoning ability.

\section*{Acknowledgment}
The research is supported by NSF CPS program under CMMI-1932187. Siyuan Xiang gratefully thanks the IDC Foundation for its scholarship. The authors gratefully thank our human test participants and the helpful comments from Zhaorong Wang, Zhiding Yu, Srikumar Ramalingam, and the anonymous reviewers.

{\small
\bibliographystyle{ieee_fullname}
\bibliography{egbib}

\begin{thebibliography}{10}\itemsep=-1pt

\bibitem{QuickDrawWebsite}
Quick! {Draw}.
\newblock https://quickdraw.withgoogle.com/.

\bibitem{bakhtin2019phyre}
Anton Bakhtin, Laurens van~der Maaten, Justin Johnson, Laura Gustafson, and
  Ross Girshick.
\newblock Phyre: A new benchmark for physical reasoning.
\newblock {\em arXiv preprint arXiv:1908.05656}, 2019.

\bibitem{bisk2018learning}
Yonatan Bisk, Kevin~J Shih, Yejin Choi, and Daniel Marcu.
\newblock Learning interpretable spatial operations in a rich 3d blocks world.
\newblock In {\em Thirty-Second AAAI Conference on Artificial Intelligence},
  2018.

\bibitem{bodner1997purdue}
George~M Bodner and Roland~B Guay.
\newblock The purdue visualization of rotations test.
\newblock {\em The Chemical Educator}, 2(4):1--17, 1997.

\bibitem{brendel2019approximating}
Wieland Brendel and Matthias Bethge.
\newblock Approximating cnns with bag-of-local-features models works
  surprisingly well on imagenet.
\newblock {\em arXiv preprint arXiv:1904.00760}, 2019.

\bibitem{chang2015shapenet}
Angel~X Chang, Thomas Funkhouser, Leonidas Guibas, Pat Hanrahan, Qixing Huang,
  Zimo Li, Silvio Savarese, Manolis Savva, Shuran Song, Hao Su, et~al.
\newblock Shapenet: An information-rich 3d model repository.
\newblock {\em arXiv preprint arXiv:1512.03012}, 2015.

\bibitem{chen2019touchdown}
Howard Chen, Alane Suhr, Dipendra Misra, Noah Snavely, and Yoav Artzi.
\newblock Touchdown: Natural language navigation and spatial reasoning in
  visual street environments.
\newblock In {\em Proceedings of the IEEE Conference on Computer Vision and
  Pattern Recognition}, pages 12538--12547. Ieee, 2019.

\bibitem{cole2008people}
Forrester Cole, Aleksey Golovinskiy, Alex Limpaecher, Heather~Stoddart Barros,
  Adam Finkelstein, Thomas Funkhouser, and Szymon Rusinkiewicz.
\newblock Where do people draw lines?
\newblock {\em ACM Transactions on Graphics (ToG)}, 27(3):88, 2008.

\bibitem{cole2009well}
Forrester Cole, Kevin Sanik, Doug Decarlo, Adam Finkelstein, Thomas~Allen
  Funkhouser, Szymon~M Rusinkiewicz, and Manish Singh.
\newblock How well do line drawings depict shape?
\newblock {\em ACM Transactions on Graphics}, 28(3):28, 2009.

\bibitem{dai2017scannet}
Angela Dai, Angel~X Chang, Manolis Savva, Maciej Halber, Thomas Funkhouser, and
  Matthias Nie{\ss}ner.
\newblock Scannet: Richly-annotated 3d reconstructions of indoor scenes.
\newblock In {\em Proceedings of the IEEE Conference on Computer Vision and
  Pattern Recognition}, pages 5828--5839. Ieee, 2017.

\bibitem{delanoy20183d}
Johanna Delanoy, Mathieu Aubry, Phillip Isola, Alexei~A Efros, and Adrien
  Bousseau.
\newblock 3d sketching using multi-view deep volumetric prediction.
\newblock {\em Proceedings of the ACM on Computer Graphics and Interactive
  Techniques}, 1(1):1--22, 2018.

\bibitem{deng2009imagenet}
Jia Deng, Wei Dong, Richard Socher, Li-Jia Li, Kai Li, and Li Fei-Fei.
\newblock Imagenet: A large-scale hierarchical image database.
\newblock In {\em Proceedings of the IEEE Conference on Computer Vision and
  Pattern Recognition}, pages 248--255. Ieee, 2009.

\bibitem{eslami2018neural}
SM~Ali Eslami, Danilo~Jimenez Rezende, Frederic Besse, Fabio Viola, Ari~S
  Morcos, Marta Garnelo, Avraham Ruderman, Andrei~A Rusu, Ivo Danihelka, Karol
  Gregor, et~al.
\newblock Neural scene representation and rendering.
\newblock {\em Science}, 360(6394):1204--1210, 2018.

\bibitem{funkhouser2003search}
Thomas Funkhouser, Patrick Min, Michael Kazhdan, Joyce Chen, Alex Halderman,
  David Dobkin, and David Jacobs.
\newblock A search engine for 3d models.
\newblock {\em ACM Transactions on Graphics (TOG)}, 22(1):83--105, 2003.

\bibitem{garg2016unsupervised}
Ravi Garg, Vijay~Kumar BG, Gustavo Carneiro, and Ian Reid.
\newblock Unsupervised cnn for single view depth estimation: Geometry to the
  rescue.
\newblock In {\em European Conference on Computer Vision}, pages 740--756.
  Springer, 2016.

\bibitem{groueix2018atlasnet}
Thibault Groueix, Matthew Fisher, Vladimir~G Kim, Bryan~C Russell, and Mathieu
  Aubry.
\newblock A papier-m{\^a}ch{\'e} approach to learning 3d surface generation.
\newblock In {\em Proceedings of the IEEE Conference on Computer Vision and
  Pattern Recognition}, pages 216--224. Ieee, 2018.

\bibitem{gryaditskaya2019opensketch}
Yulia Gryaditskaya, Mark Sypesteyn, Jan~Willem Hoftijzer, Sylvia Pont, Fredo
  Durand, and Adrien Bousseau.
\newblock Opensketch: a richly-annotated dataset of product design sketches.
\newblock {\em ACM Transactions on Graphics (TOG)}, 38(6):232, 2019.

\bibitem{he2016deep}
Kaiming He, Xiangyu Zhang, Shaoqing Ren, and Jian Sun.
\newblock Deep residual learning for image recognition.
\newblock In {\em Proceedings of the IEEE Conference on Computer Vision and
  Pattern Recognition}, pages 770--778. Ieee, 2016.

\bibitem{hsi1997role}
Sherry Hsi, Marcia~C Linn, and John~E Bell.
\newblock The role of spatial reasoning in engineering and the design of
  spatial instruction.
\newblock {\em Journal of engineering education}, 86(2):151--158, 1997.

\bibitem{huang1989motion}
Thomas~S. Huang and Chia-Hoang Lee.
\newblock Motion and structure from orthographic projections.
\newblock {\em IEEE Transactions on Pattern Analysis and Machine Intelligence},
  11(5):536--540, 1989.

\bibitem{hudson2019gqa}
Drew~A Hudson and Christopher~D Manning.
\newblock Gqa: A new dataset for real-world visual reasoning and compositional
  question answering.
\newblock In {\em Proceedings of the IEEE Conference on Computer Vision and
  Pattern Recognition}, pages 6700--6709. Ieee, 2019.

\bibitem{isola2017image}
Phillip Isola, Jun-Yan Zhu, Tinghui Zhou, and Alexei~A Efros.
\newblock Image-to-image translation with conditional adversarial networks.
\newblock In {\em Proceedings of the IEEE Conference on Computer Vision and
  Pattern Recognition}, pages 1125--1134. Ieee, 2017.

\bibitem{johnson2017clevr}
Justin Johnson, Bharath Hariharan, Laurens van~der Maaten, Li Fei-Fei, C
  Lawrence~Zitnick, and Ross Girshick.
\newblock Clevr: A diagnostic dataset for compositional language and elementary
  visual reasoning.
\newblock In {\em Proceedings of the IEEE Conference on Computer Vision and
  Pattern Recognition}, pages 2901--2910. Ieee, 2017.

\bibitem{kafle2018dvqa}
Kushal Kafle, Brian Price, Scott Cohen, and Christopher Kanan.
\newblock Dvqa: Understanding data visualizations via question answering.
\newblock In {\em Proceedings of the IEEE Conference on Computer Vision and
  Pattern Recognition}, pages 5648--5656. Ieee, 2018.

\bibitem{kato2018renderer}
Hiroharu Kato, Yoshitaka Ushiku, and Tatsuya Harada.
\newblock Neural 3d mesh renderer.
\newblock In {\em Proceedings of the IEEE Conference on Computer Vision and
  Pattern Recognition}, pages 3907--3916, 2018.

\bibitem{kell2013creativity}
Harrison~J Kell, David Lubinski, Camilla~P Benbow, and James~H Steiger.
\newblock Creativity and technical innovation: Spatial ability’s unique role.
\newblock {\em Psychological science}, 24(9):1831--1836, 2013.

\bibitem{koch2019abc}
Sebastian Koch, Albert Matveev, Zhongshi Jiang, Francis Williams, Alexey
  Artemov, Evgeny Burnaev, Marc Alexa, Denis Zorin, and Daniele Panozzo.
\newblock Abc: A big cad model dataset for geometric deep learning.
\newblock In {\em Proceedings of the IEEE Conference on Computer Vision and
  Pattern Recognition}, pages 9601--9611. Ieee, 2019.

\bibitem{krishna2017visual}
Ranjay Krishna, Yuke Zhu, Oliver Groth, Justin Johnson, Kenji Hata, Joshua
  Kravitz, Stephanie Chen, Yannis Kalantidis, Li-Jia Li, David~A Shamma, et~al.
\newblock Visual genome: Connecting language and vision using crowdsourced
  dense image annotations.
\newblock {\em International Journal of Computer Vision}, 123(1):32--73, 2017.

\bibitem{lohman1996spatial}
David~F Lohman.
\newblock Spatial ability and {G}.
\newblock {\em Human abilities: Their nature and measurement}, 97:116, 1996.

\bibitem{lowrie2002influence}
Tom Lowrie.
\newblock The influence of visual and spatial reasoning in interpreting
  simulated 3d worlds.
\newblock {\em International Journal of Computers for Mathematical Learning},
  7(3):301, 2002.

\bibitem{lubinski2010spatial}
David Lubinski.
\newblock Spatial ability and stem: A sleeping giant for talent identification
  and development.
\newblock {\em Personality and Individual Differences}, 49(4):344--351, 2010.

\bibitem{macworth1973interpreting}
AK Macworth.
\newblock Interpreting pictures of polyhedral scenes.
\newblock {\em Artificial intelligence}, 4(2):121--137, 1973.

\bibitem{malik1987interpreting}
Jitendra Malik.
\newblock Interpreting line drawings of curved objects.
\newblock {\em International Journal of Computer Vision}, 1(1):73--103, 1987.

\bibitem{olszewski2019transformable}
Kyle Olszewski, Sergey Tulyakov, Oliver Woodford, Hao Li, and Linjie Luo.
\newblock Transformable bottleneck networks.
\newblock In {\em Proceedings of the IEEE International Conference on Computer
  Vision}, pages 7648--7657, 2019.

\bibitem{PyTorch:NIPS19}
Adam Paszke, Sam Gross, Francisco Massa, Adam Lerer, James Bradbury, Gregory
  Chanan, Trevor Killeen, Zeming Lin, Natalia Gimelshein, Luca Antiga, Alban
  Desmaison, Andreas Kopf, Edward Yang, Zachary DeVito, Martin Raison, Alykhan
  Tejani, Sasank Chilamkurthy, Benoit Steiner, Lu Fang, Junjie Bai, and Soumith
  Chintala.
\newblock {PyTorch}: An imperative style, high-performance deep learning
  library.
\newblock In {\em Advances in Neural Information Processing Systems}, pages
  8024--8035, 2019.

\bibitem{pythonocc}
Python{OCC}.
\newblock {3D} {CAD/CAE/PLM} development framework for the {P}ython programming
  language, 2018.
\newblock http://www.pythonocc.org.

\bibitem{ramalingam2013lifting}
Srikumar Ramalingam and Matthew Brand.
\newblock Lifting 3d manhattan lines from a single image.
\newblock In {\em Proceedings of the IEEE International Conference on Computer
  Vision}, pages 497--504. Ieee, 2013.

\bibitem{santoro2018measuring}
Adam Santoro, Felix Hill, David Barrett, Ari Morcos, and Timothy Lillicrap.
\newblock Measuring abstract reasoning in neural networks.
\newblock In {\em International Conference on Machine Learning}, pages
  4477--4486, 2018.

\bibitem{simonyan2014very}
Karen Simonyan and Andrew Zisserman.
\newblock Very deep convolutional networks for large-scale image recognition.
\newblock {\em arXiv preprint arXiv:1409.1556}, 2014.

\bibitem{sitzmann2019deepvoxels}
Vincent Sitzmann, Justus Thies, Felix Heide, Matthias Nie{\ss}ner, Gordon
  Wetzstein, and Michael Zollh{\"o}fer.
\newblock Deepvoxels: Learning persistent 3d feature embeddings.
\newblock In {\em Proceedings of the IEEE Conference on Computer Vision and
  Pattern Recognition}, pages 2437--2446. Ieee, 2019.

\bibitem{su2015render}
Hao Su, Charles~R Qi, Yangyan Li, and Leonidas~J Guibas.
\newblock Render for cnn: Viewpoint estimation in images using cnns trained
  with rendered 3d model views.
\newblock In {\em Proceedings of the IEEE International Conference on Computer
  Vision}, pages 2686--2694, 2015.

\bibitem{sugihara1986machine}
Kokichi Sugihara.
\newblock {\em Machine interpretation of line drawings}, volume~1.
\newblock MIT press Cambridge, 1986.

\bibitem{vandenberg1978mental}
Steven~G Vandenberg and Allan~R Kuse.
\newblock Mental rotations, a group test of three-dimensional spatial
  visualization.
\newblock {\em Perceptual and motor skills}, 47(2):599--604, 1978.

\bibitem{varley2005frontal}
PAC Varley, Ralph~R Martin, and Hiromasa Suzuki.
\newblock Frontal geometry from sketches of engineering objects: is line
  labelling necessary?
\newblock {\em Computer-Aided Design}, 37(12):1285--1307, 2005.

\bibitem{wai2009spatial}
Jonathan Wai, David Lubinski, and Camilla~P Benbow.
\newblock Spatial ability for stem domains: Aligning over 50 years of
  cumulative psychological knowledge solidifies its importance.
\newblock {\em Journal of Educational Psychology}, 101(4):817, 2009.

\bibitem{wu2019phasecam3d}
Yicheng Wu, Vivek Boominathan, Huaijin Chen, Aswin Sankaranarayanan, and Ashok
  Veeraraghavan.
\newblock Phasecam3d—learning phase masks for passive single view depth
  estimation.
\newblock In {\em 2019 IEEE International Conference on Computational
  Photography (ICCP)}, pages 1--12. Ieee, 2019.

\bibitem{xiazamirhe2018gibsonenv}
Fei Xia, Amir R.~Zamir, Zhi-Yang He, Alexander Sax, Jitendra Malik, and Silvio
  Savarese.
\newblock Gibson env: real-world perception for embodied agents.
\newblock In {\em Proceedings of the IEEE Conference on Computer Vision and
  Pattern Recognition}, pages 9068--9079. Ieee, 2018.

\bibitem{yang2018dataset}
Guangyu~Robert Yang, Igor Ganichev, Xiao-Jing Wang, Jonathon Shlens, and David
  Sussillo.
\newblock A dataset and architecture for visual reasoning with a working
  memory.
\newblock In {\em European Conference on Computer Vision}, pages 729--745,
  2018.

\bibitem{yang2018foldingnet}
Yaoqing Yang, Chen Feng, Yiru Shen, and Dong Tian.
\newblock Foldingnet: Point cloud auto-encoder via deep grid deformation.
\newblock In {\em Proceedings of the IEEE Conference on Computer Vision and
  Pattern Recognition}, pages 206--215. Ieee, 2018.

\bibitem{yoon2011revised}
SY Yoon.
\newblock Revised purdue spatial visualization test: Visualization of rotations
  (revised psvt: R).
\newblock {\em Texas A\&M University, College Station, TX}, 2011.

\bibitem{yue2006spatial}
Jianping Yue.
\newblock Spatial visualization by isometric drawing.
\newblock In {\em Proceedings of the2006 IJMEINTERTECH Conference, Union, New
  Jersey}, 2006.

\bibitem{zellers2019vcr}
Rowan Zellers, Yonatan Bisk, Ali Farhadi, and Yejin Choi.
\newblock From recognition to cognition: Visual commonsense reasoning.
\newblock In {\em Proceedings of the IEEE Conference on Computer Vision and
  Pattern Recognition}, pages 6720--6731. Ieee, June 2019.

\bibitem{zhu2017target}
Yuke Zhu, Roozbeh Mottaghi, Eric Kolve, Joseph~J Lim, Abhinav Gupta, Li
  Fei-Fei, and Ali Farhadi.
\newblock Target-driven visual navigation in indoor scenes using deep
  reinforcement learning.
\newblock In {\em 2017 IEEE international conference on robotics and automation
  (ICRA)}, pages 3357--3364. Ieee, 2017.

\end{thebibliography}
}

\appendix
\section*{Supplementary}
\section{SPARE3D-CSG}
\paragraph{Why CSG models?}
CSG models are randomly generated from simple primitives, like sphere, cube, cone, and torus, with boolean operations including union, intersection, and difference. Therefore, it allows us to control the complexity of 3D models. In the \textit{SPARE3D-CSG} dataset, we generate three sets of $4000$ 3D models, i.e., a total of $12000$, from two, three, and four simple random primitives respectively. With more primitives in a model, the complexity of the model increases, and so does the difficulty level of \textit{SPARE3D-CSG} tasks generated from those models.

When generating tasks for view consistency reasoning and camera pose reasoning, for training and testing dataset, we select the same number of 2D drawings from two, three, and four simple primitive model sets. In this way, we ensure that our baseline methods are trained and tested on tasks with the same difficulty levels.

\begin{figure*}[h!]
\centering
\includegraphics[width=\textwidth,trim=.3cm 3.3cm .2cm .0cm,clip, ]{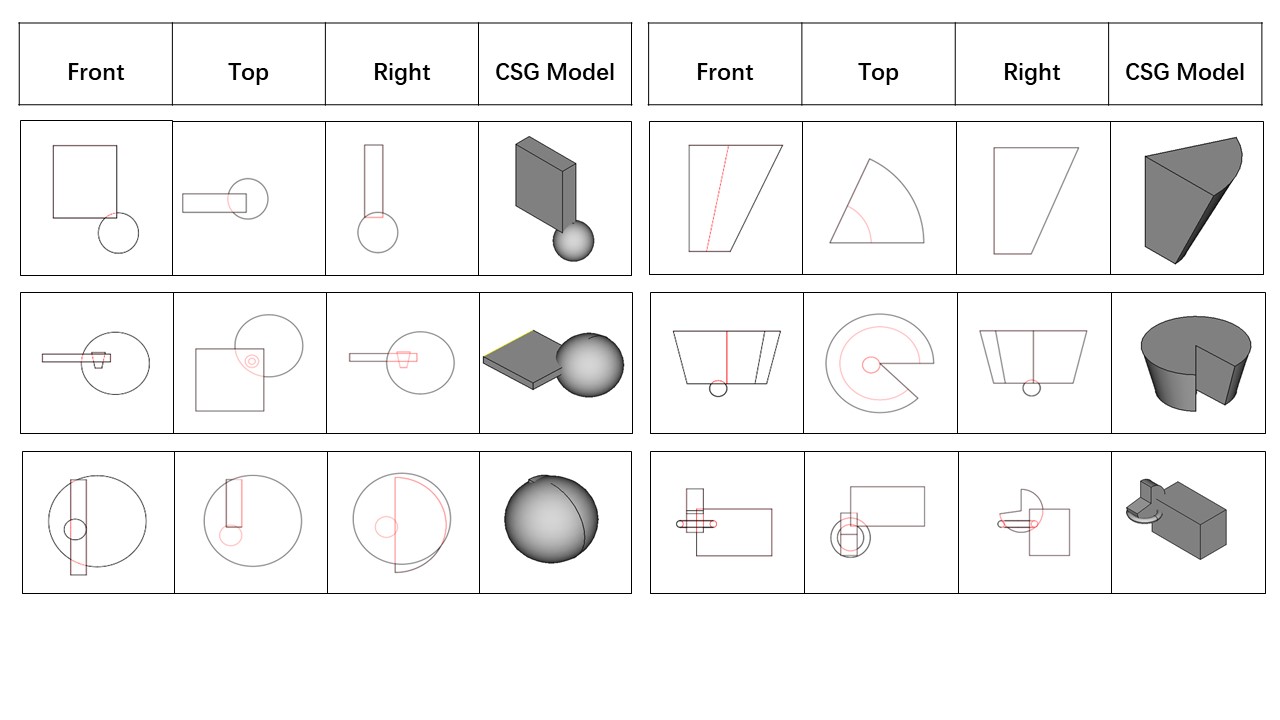}
\caption{\textbf{CSG model examples}. In each example chunk, the first three columns are F, T, R drawings, respectively; the fourth column is the rendered CSG model (seeing from pose 2 as explained in the main paper). The models in the second row, third row, and fourth row are generated from two, three, and four simple primitives, respectively.
}
\label{csg_example}
\end{figure*}

\paragraph{CSG model generation.}
Most of the objects in the real world look reasonably regular in shape because they are usually designed and organized in certain rules manually. The \textit{SPARE3D-CSG} dataset is generated using the following two rules. First, to create a CSG model from simple primitives, rotation angles for these primitives are randomly selected from $0^{\circ}$, $90^{\circ}$, $180^{\circ}$, and $270^{\circ}$. Second, these primitives are only rotated about $X$, $Y$, or $Z$ axes. Example models can be seen from Figure~\ref{csg_example}.

\section{Baseline Methods Formulation}
We formulate the \textit{3-View to Isometric} and \textit{Pose to Isometric} tasks as either binary classification or metric learning. The \textit{Pose to Isometric} task is formulated as the multi-class classification. \textit{Isometric View Generation} is treated as conditional image generation, \textit{Point Cloud Generation} is expressed as 3D point cloud generation from multi-view image. In this section, we use $I_F$, $I_T$, and $I_R$ to represent line drawings from the front, top, and right view, respectively, each of which is a 3-channel RGB image. The backbone neural networks are represented as feature extraction function $f$ for each task. The detailed formula of each task is shown in the following subsections.  
\subsection{Three-View to Isometric}
\paragraph{Binary Classification.}
$I_F$, $I_T$, $I_R$, and a query image $I_q$ from the choices are concatenated along the feature dimension, to form a 12-channel composite image $I_c$.
Then a CNN-based binary classifier $f_\theta:\mathbb{R}^{12\times H\times W}\to[0, 1]$ is trained to map $I_c$ to $\hat{p}(\theta)$, which is the probability that $I_q$ is the isometric image. $\theta$ represents the parameters of the neural network. 
Binary cross-entropy (BCE) loss is applied to train the neural nework:
 \begin{equation}\label{eq:bce}
     L(\theta)= -p \log{\hat{p}(\theta)}-(1-p)\log{(1-\hat{p}(\theta))},
 \end{equation} where $p\in\mathbb{Z}_2$ is the ground truth label of whether $I_q$ is the isometric drawing consistent with the input. 

We take four images (three images from the question and one image from answer) as a group. Therefore, each time, we have four groups of data to process. We use VGG and ResNet to encode a group of images to a feature vector in $R^{1}$ space. Then we concatenate four feature vectors and use softmax to get a $4\times1$ vector of distribution probability. 

\paragraph{Metric Learning.}
$I_{F}$, $I_{T}$, and $I_{R}$  are concatenated to form a 9-channel composite image $I_c$. Then $I_c$ is fed into a CNN-based encoder $f_\theta:\mathbb{R}^{9\times H \times  W}\to \mathbb{R}^M$. A query image 
$I_q$ from the choices is fed into another CNN-based encoder $g_\phi:\mathbb{R}^{3\times H \times  W}\to \mathbb{R}^M$. $\theta$ and $\phi$ represent the parameters of the two neural networks respectively. We use $l_2$ distance $d(\theta, \phi)=\|f_\theta(I_c)-g_\theta(I_q)\|$ to measure the correctness of $I_q$. Smaller $d(\theta, \phi)$ indicates higher correctness that $I_q$ is the isometric image among the four choices. We apply margin ranking loss to train the networks:
\begin{equation}\label{eq:rankingloss}
     L(\theta, \phi)=\sum_{k=1}^3\max{(0, d_c(\theta, \phi)-d_w^{k}(\theta, \phi)+m)},
\end{equation} 
where $d_c(\theta, \phi)$ is the correctness measurement of the correct $I_q$, and $d_w^{k}(\theta, \phi)$ is the correctness measurement of the $k$th wrong $I_q$. $m=2$ is the margin we use during training. We set $M=128$ in this task.

\subsection{Isometric to Pose}
\paragraph{Multi-class Classification.}
$I_F$, $I_T$, $I_R$ and the isometric image $I_i$ are concatenated to form a 12-channel composite image $I_c$. Then a CNN-based classifier $f_\theta:\mathbb{R}^{12\times H\times W}\to[0, 1]^4$ is trained to map $I_c$ to a four-vector $\hat{\mathbf{p}}(\theta)=[\hat{p}_1(\theta),\,\hat{p}_2(\theta),\,\hat{p}_3(\theta),\,\hat{p}_4(\theta)]^T$ that represents the probability of $I_i$ is taken at pose 1, 5, 2 and 6 respectively. Cross-entropy loss is applied to train the neural network.
\begin{equation}
    L(\theta)= -\sum_{k=1}^{4}p_k\log{\hat{p}_k(\theta)},
\end{equation} where $p_k=1$ if $I_i$ is taken at the $k$th view point.
For this task, we encode the concatenated four images in the question into a $R^{4}$ feature vector using function F. Then we use softmax to get the probability distribution and compute the cross-entropy loss between the feature vector and the encoding of the answer.

\subsection{Pose to Isometric}
\paragraph{Binary Classification.}
$I_F$, $I_T$, $I_R$ and a query image $I_q$ from the choices are concatenated to form a 12-channel composite image $I_c$. This composite image is fed into a CNN $f:\mathbb{R}^{12\times H\times W}\to\mathbb{R}^{K}$. Then, we concatenate the output with a 8-dimensional one-hot vector $z\in \mathbb{Z}_2^8$, representing the given camera pose to create a codeword $c\in\mathbb{R}^{K}\times\mathbb{Z}_2^8$. $c$ is then fed into a fully-connected network $g_\phi:\mathbb{R}^{K}\times\mathbb{Z}_2^8\to[0, 1]$. We apply BCE loss as equation~\eqref{eq:bce} to train the neural network. Here we set $K=128$.


\paragraph{Metric Learning.}
Similar to the binary 
classification formulation, we again obtain  $c\in\mathbb{R}^{K}\times\mathbb{Z}_2^8$ from $I_F$, $I_T$, $I_R$ and $z$. $c$ is then fed into a fully-connected network $g_\theta:\mathbb{R}^{K}\times\mathbb{Z}_2^8\to R^M$. For each answer image, we obtain a feature vector in $\mathbb{R}^{M}$ space using another CNN-based encoder $h_\omega:\mathbb{R}^{3\times H\times W}\to \mathbb{R}^{M}$. Then we can caculate the margin ranking loss similar to equation~\eqref{eq:rankingloss}. In our experiment, $K = 128$ and $M = 50$.

\subsection{Isometric View Generation}
For this task, we use Pix2Pix~\cite{isola2017image}, a conditional generative adversarial network, to generate the isometric drawing for each question. The generator network $G(x)$ needs to learn a mapping from the three-view drawings to the isometric drawing. The input $x$ is a $\mathbb{R}^{9 \times H\times W}$ tensor generated by concatenating F, R, T images. When training the pix2pix model on our dataset, we use label flipping and label smoothing to improve the stability of the model.

\subsection{Point Cloud Generation}

We use a FoldingNet~\cite{yang2018foldingnet}-like and AtlasNet~\cite{groueix2018atlasnet}-like decoding architectures to generate a 3D object's point cloud with 2025 points from a latent code $c \in \mathbb{R}^{512}$, which is encoded by a ResNet-18 CNN from a 9-channel concatenated F, T, R image tensor.

\section{Implementation Details of Baseline Methods}

In SPARE3D-ABC, we use ResNet-50, VGG-16, and BagNet as our deep network architectures for \textit{3-View to Isometric}, \textit{Pose to Isometric}, and \textit{Isometric to Pose} tasks, to extract features from given drawings. The network architecture details are explained below for each baseline method.

All the hyper-parameters in each baseline method for each task, whose drawings are generated from models in ABC dataset, are tuned using a validation set of 500 questions, although we have not searched for the optimal hyper-parameters extensively using methods like grid search. 

\subsection{3-View to Isometric}
\paragraph{Binary classification.}

We slightly modify the ResNet-50 base network to adapt to our tasks. The first convolutional layer has $12$ input channels, $64$ output channels, with kernel size $(3,3)$, instead of the original $(7,7)$, stride and padding $(1,1)$, instead of the original stride$(2,2)$ and padding $(3,3)$. The last fully-connected layer maps the feature vector from $\mathbb{R}^{2048}\to \mathbb{R}^{1}$. Other layers are exactly the same as the original ResNet-50 network. And the above modifications are applied to all the remaining baseline methods involving ResNet-50. The learning rate is \textcolor{blue}{$0.0003$}, the batch size is $9$, and the network is trained for $50$ epochs.

Similarly, for the VGG-16 network, the first convolutional layer is modified in the same way as ResNet-50. The last fully-connected layer maps the feature vector from $\mathbb{R}^{4096}\to \mathbb{R}^{1}$. The learning rate is \textcolor{blue}{$0.00001$}, the batch size is $20$, and the network is trained for $50$ epochs.

For the BagNet-33 base network, the first convolutional layer has $12$ input channels, $64$ output channels, with kernel size $1$, stride $1$, padding $0$. The last fully-connected layer maps the feature vector from $\mathbb{R}^{2048}\to \mathbb{R}^{1}$. The learning rate is \textcolor{blue}{$0.00005$}, the batch size is $8$, and the network is trained for $49$ epochs.

\paragraph{Metric learning.}

In this formulation, two functions, $f$ and $g$, are implemented using two similar base networks for extracting features from drawings in questions and in answers, respectively. 

\textit{ResNet-50 as the base network:} For $f$, the first convolutional layer has $9$ input channels, $64$ output channels, with kernel size $(3,3)$, stride and padding $(1,1)$. The last fully connected layer maps the feature vector from $\mathbb{R}^{2048}\to \mathbb{R}^{128}$. For $g$, the first convolutional layer has $3$ input channels, $64$ output channels, with kernel size $(3,3)$, stride and padding $(1,1)$. The last fully connected layer is the same as $f$. The learning rate is $0.0001$, the batch size is $4$, and the network is trained for $50$ epochs.

\textit{VGG-16 as the base network:} For $f$, the first convolutional layer is the same as ResNet-50 $f$ for metric learning in \textit{3-View to Isometric}. The last fully-connected layer maps the feature vector from $\mathbb{R}^{4096}\to \mathbb{R}^{128}$. For $g$, the first convolutional layer is the same as ResNet-50 $g$ for metric learning in \textit{3-View to Isometric}. The last fully connected layer is the same as $f$ in this method. The learning rate is $0.00002$, the batch size is $8$, and the network is trained for $50$ epochs.

\textit{BagNet-33 as the base network:} For $f$, the first convolutional layer has $9$ input channels, $64$ output channels, with kernel size $1$, stride $1$ and padding $0$. The last fully-connected layer maps the feature vector from $\mathbb{R}^{2048}\to \mathbb{R}^{128}$. For $g$, the first convolutional layer has $3$ input channels, $64$ output channels, with kernel size $1$, stride $1$ and padding$0$. The last fully connected layer is the same as $f$ in this method. The learning rate is $0.0001$, the batch size is $4$, and the network is trained for $50$ epochs.

\subsection{Isometric to Pose}

\paragraph{Multi-class classification.}
For ResNet-50, the first convolutional layer is the same as the ResNet-50 network in binary classification for \textit{3-View to Isometric}. The last fully-connected layer maps the feature vector from $\mathbb{R}^{2048}\to \mathbb{R}^{4}$. The learning rate is \textcolor{blue}{$0.0003$}, the batch size is $70$, and the network is trained for $50$ epochs.

For VGG-16, the first convolutional layer is the same as the VGG-16 network in binary classification for \textit{3-View to Isometric}. The last fully-connected layer maps the feature vector from $\mathbb{R}^{4096}\to \mathbb{R}^{4}$. The learning rate is \textcolor{blue}{$0.0001$}, the batch size is $80$, and the network is trained for $50$ epochs.

For BagNet, the first convolutional layer is the same as the BagNet network in binary classification for \textit{3-View to Isometric}. The last fully-connected layer maps the feature vector from $\mathbb{R}^{2048}\to \mathbb{R}^{4}$. The learning rate is \textcolor{blue}{$0.001$}, the batch size is $30$, and the network is trained for $50$ epochs.

\subsection{Pose to Isometric}
\paragraph{Binary classification.}
For ResNet-50, the first convolutional layer is the same as the ResNet-50 network in binary classification for \textit{3-View to Isometric}. A fully connected layer maps the feature vector from $\mathbb{R}^{2048}\to \mathbb{R}^{128}$. After concatenating with a one-hot encoder $\mathbb{Z}_2^8$, a fully connected layer maps the concatenated feature vector from $\mathbb{R}^{136}\to \mathbb{R}^{1}$. The learning rate is $0.00002$, the batch size is $9$, and the network is trained for $50$ epochs.

For VGG-16, the first convolutional layer is the same as the VGG-16 network in binary classification for \textit{3-View to Isometric}. The last fully-connected layer maps the feature vector from $\mathbb{R}^{4096}\to \mathbb{R}^{128}$. The last layer is the same as the ResNet-50 network in binary classification for \textit{Pose to Isometric}. The learning rate is $0.0001$, the batch size is $30$, and the network is trained for $50$ epochs.

For BagNet, the first convolutional layer is the same as the BagNet network in binary classification for \textit{3-View to Isometric}. A fully connected layer maps the feature vector from $\mathbb{R}^{2048}\to \mathbb{R}^{128}$. The last layer is the same as the ResNet-50 network in binary classification for \textit{Pose to Isometric}. The learning rate is $0.00005$, the batch size is $8$, and the network is trained for $50$ epochs.

\paragraph{Metric learning.}
Similar to the metric learning formulation for the task ``\textit{3-View to Isometric}'', there are two functions $f$ and $g$ used to extract features from drawings in the question and the answers respectively.

For ResNet-50, the first convolutional layer of $f$ is the same as ResNet-50 $f$ for metric learning in \textit{3-View to Isometric}. The fully connected layer maps the feature vector from $\mathbb{R}^{2048}\to \mathbb{R}^{128}$. After concatenating with a one-hot encoder $\mathbb{Z}_2^8$, a linear layer maps the concatenated feature vector from $\mathbb{R}^{136}\to \mathbb{R}^{50}$. For $g$, the first convolutional layer is the same as ResNet-50 $g$ for metric learning in \textit{3-View to Isometric}. The last fully-connected layer maps the feature vector from $\mathbb{R}^{2048}\to \mathbb{R}^{50}$. The learning rate is $0.00001$, the batch size is $4$, and the network is trained for $47$ epochs.

For VGG-16, the first convolutional layer of $f$ is the same as VGG-16 $f$ for metric learning in \textit{3-View to Isometric}. The fully connected layer maps the feature vector from $\mathbb{R}^{4096}\to \mathbb{R}^{128}$. For $g$, the first convolutional layer is the same as VGG-16 $g$ for metric learning in \textit{3-View to Isometric}. The last fully-connected layer maps the feature vector from $\mathbb{R}^{4096}\to \mathbb{R}^{50}$. Other architectures are the same as VGG-16 for metric learning in \textit{Pose to Isometric}. The learning rate is $0.000005$, the batch size is $10$, and the network is trained for $42$ epochs.

For BagNet-33, the first convolutional layer of $f$ is the same as BagNet $f$ for metric learning in \textit{3-View to Isometric}. For $g$, the first convolutional layer is the same as BagNet $g$ for metric learning in \textit{3-View to Isometric}. Other architectures are the same as BagNet for metric learning in \textit{Pose to Isometric}. The learning rate is $0.0001$, the batch size is $4$, and the network is trained for $41$ epochs.

\subsection{Isometric View Generation}
As mentioned in the baseline method part, we use the Pix2Pix network to generate isometric drawings for each question. The first layer has $9$ input channels.

\subsection{Point Cloud Generation}
For FoldingNet-like and AtlasNet-like architectures, the number of output points for a 3D object is $2025$. The latent code is $c \in \mathbb{R}^{512}$. Other architectures are the same as in FoldingNet paper and AtlasNet paper, respectively, except that the original point cloud encoder is replaced with a ResNet-18 with 9 input channels. The network is trained for $1000$ epochs.

\subsection{Crowd-sourcing Website}
\begin{figure*}
    \centering
    \includegraphics[width=\columnwidth]{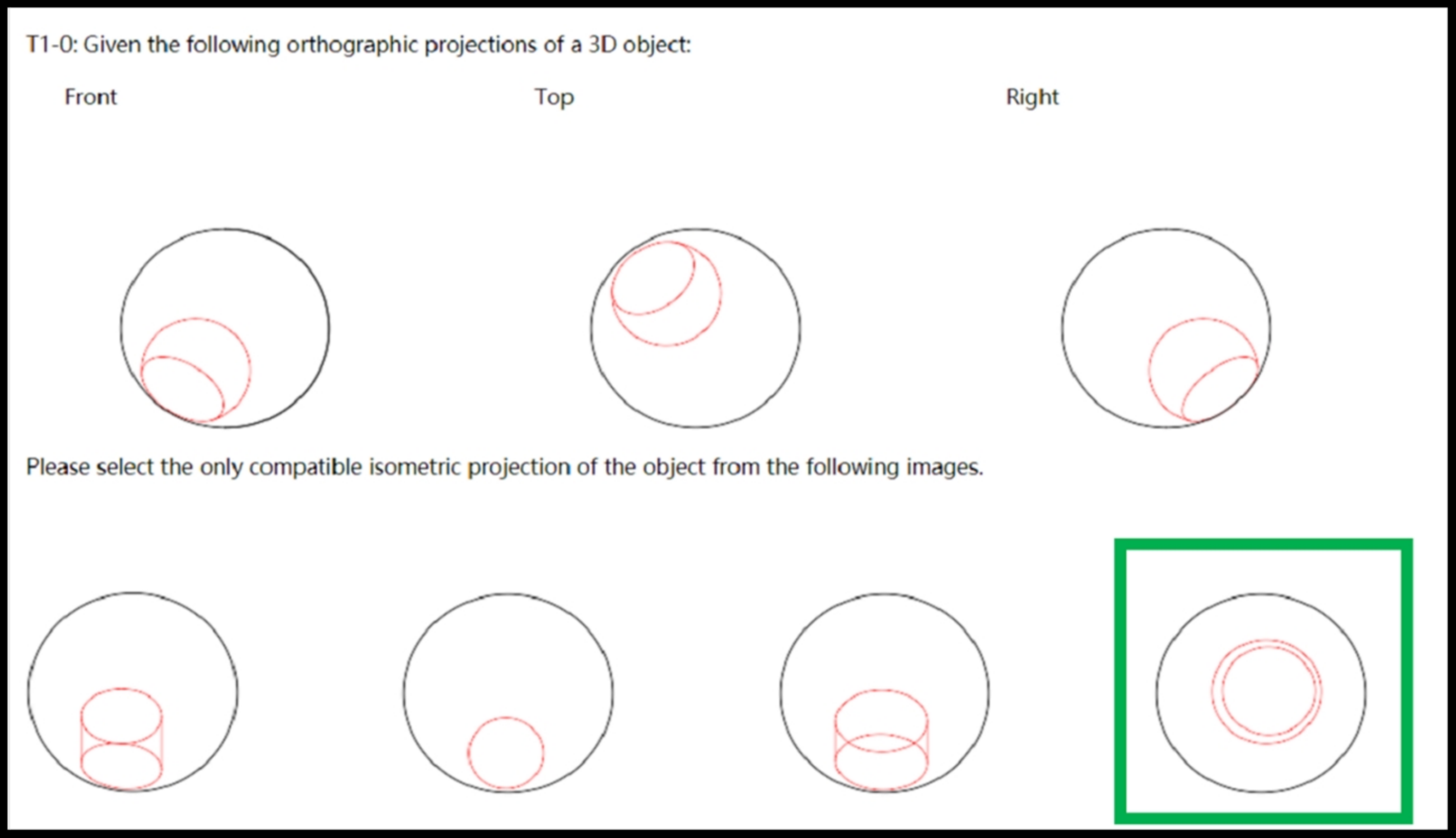}
    \includegraphics[width=\columnwidth]{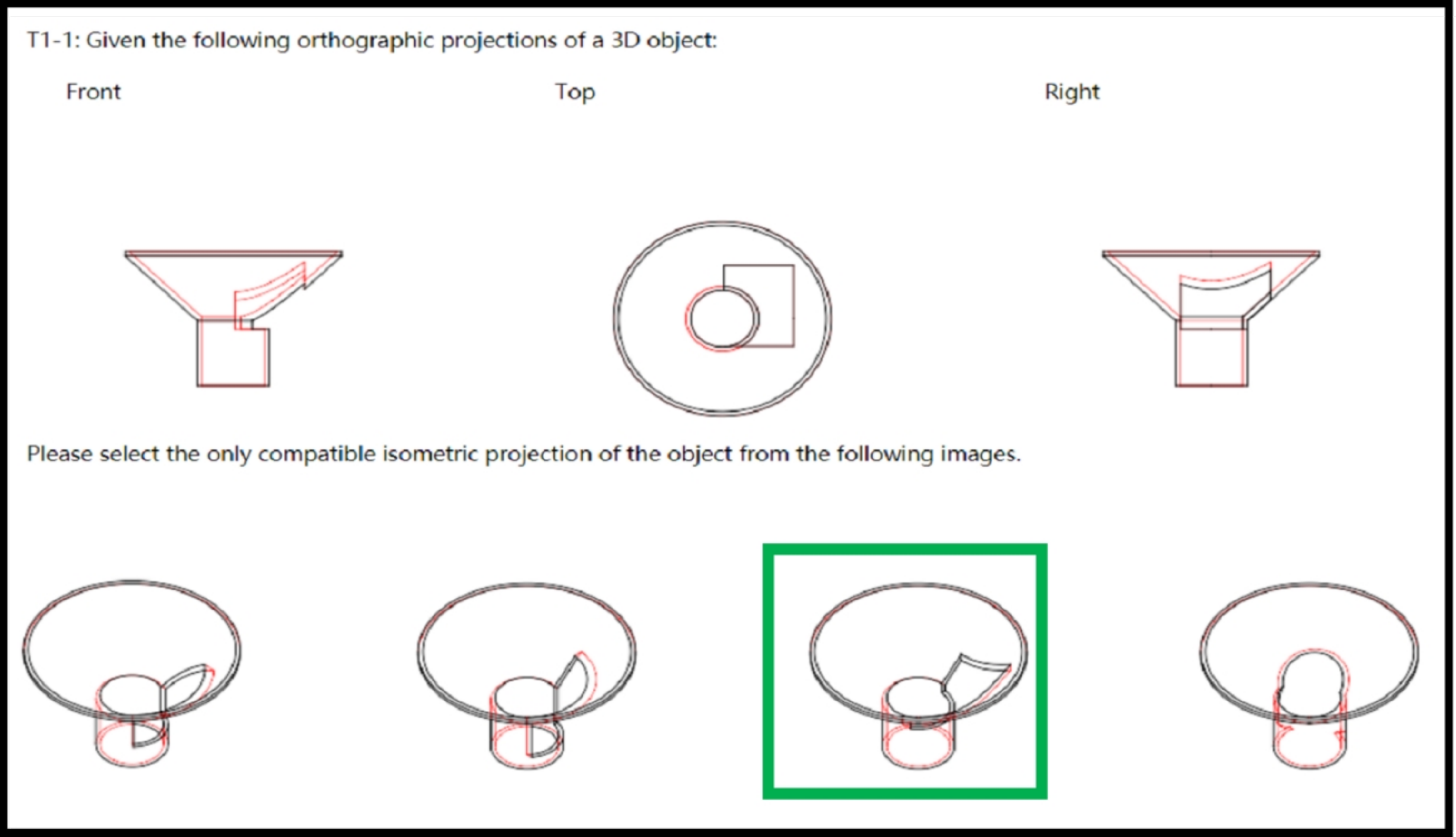}
    \includegraphics[width=\columnwidth]{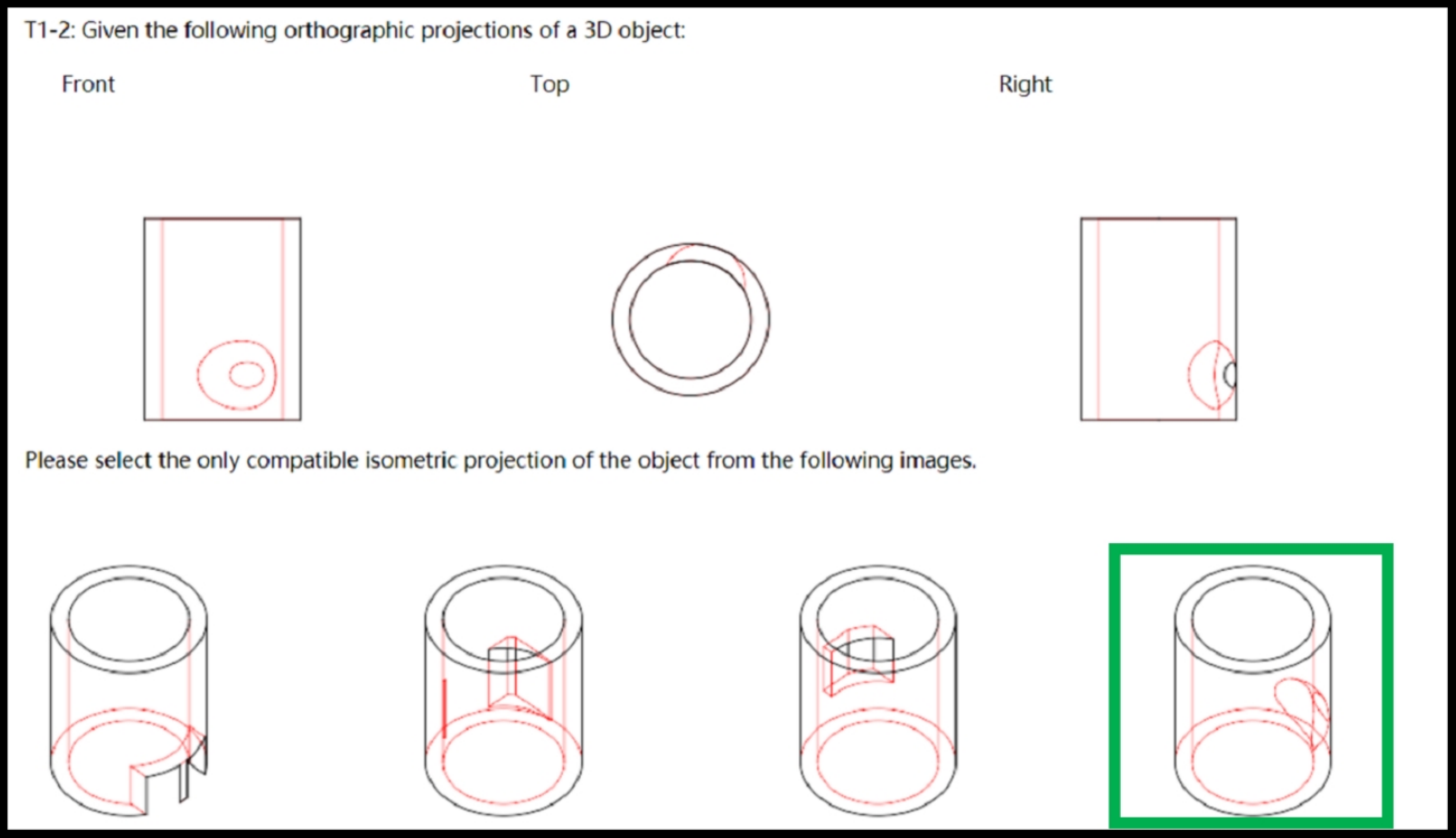}
    \includegraphics[width=\columnwidth]{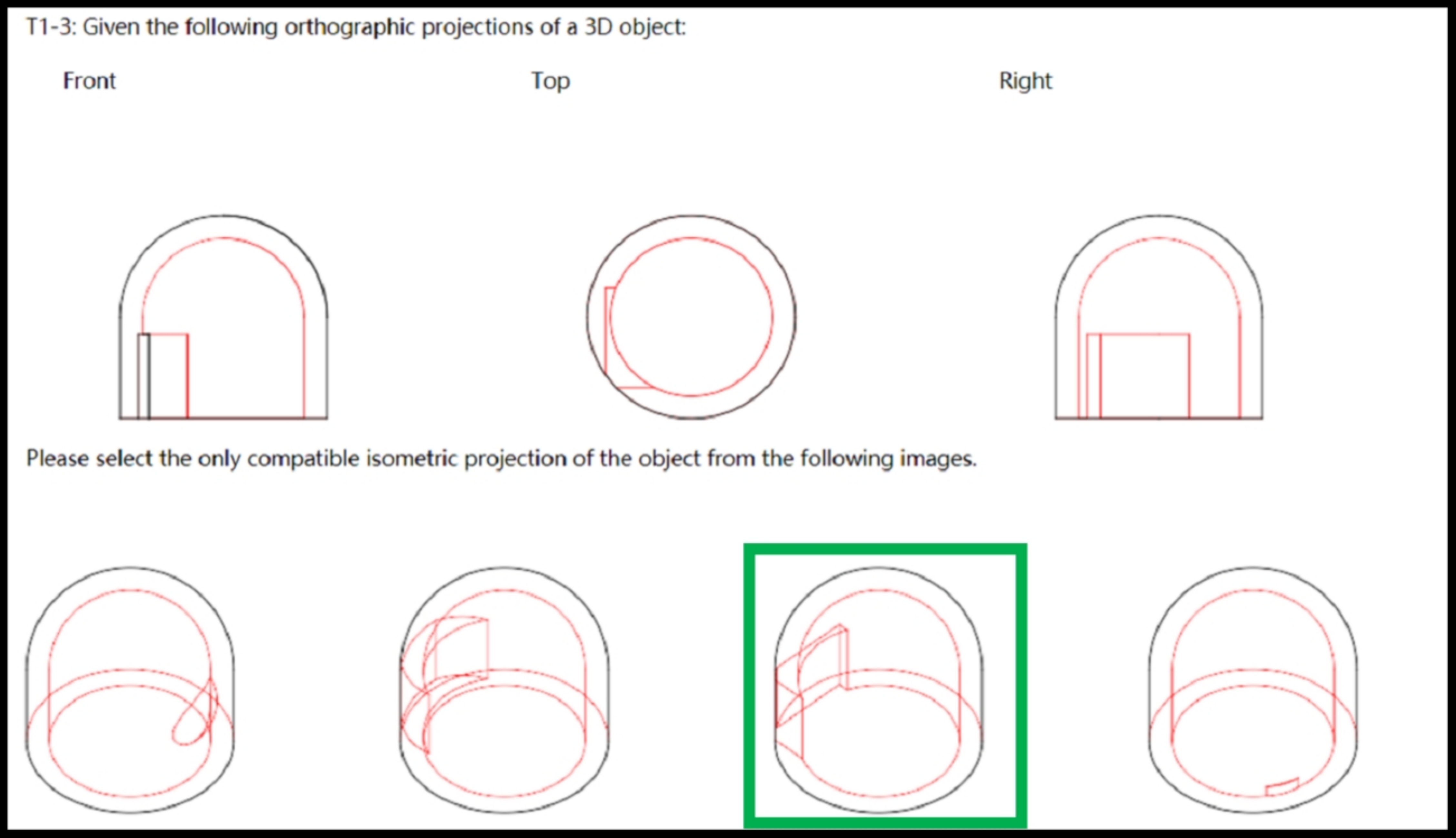}
    \includegraphics[width=\columnwidth]{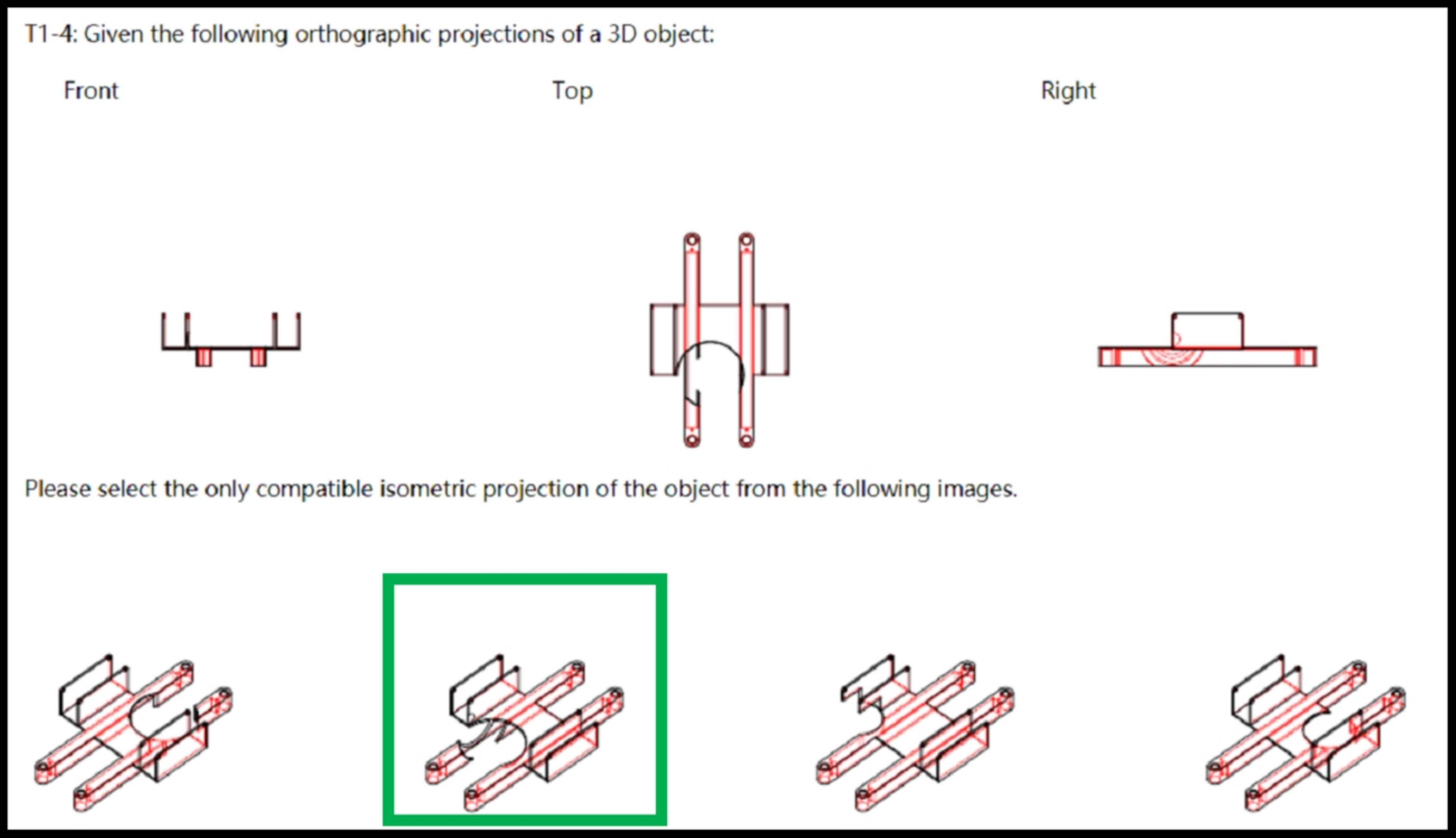}
    \includegraphics[width=\columnwidth]{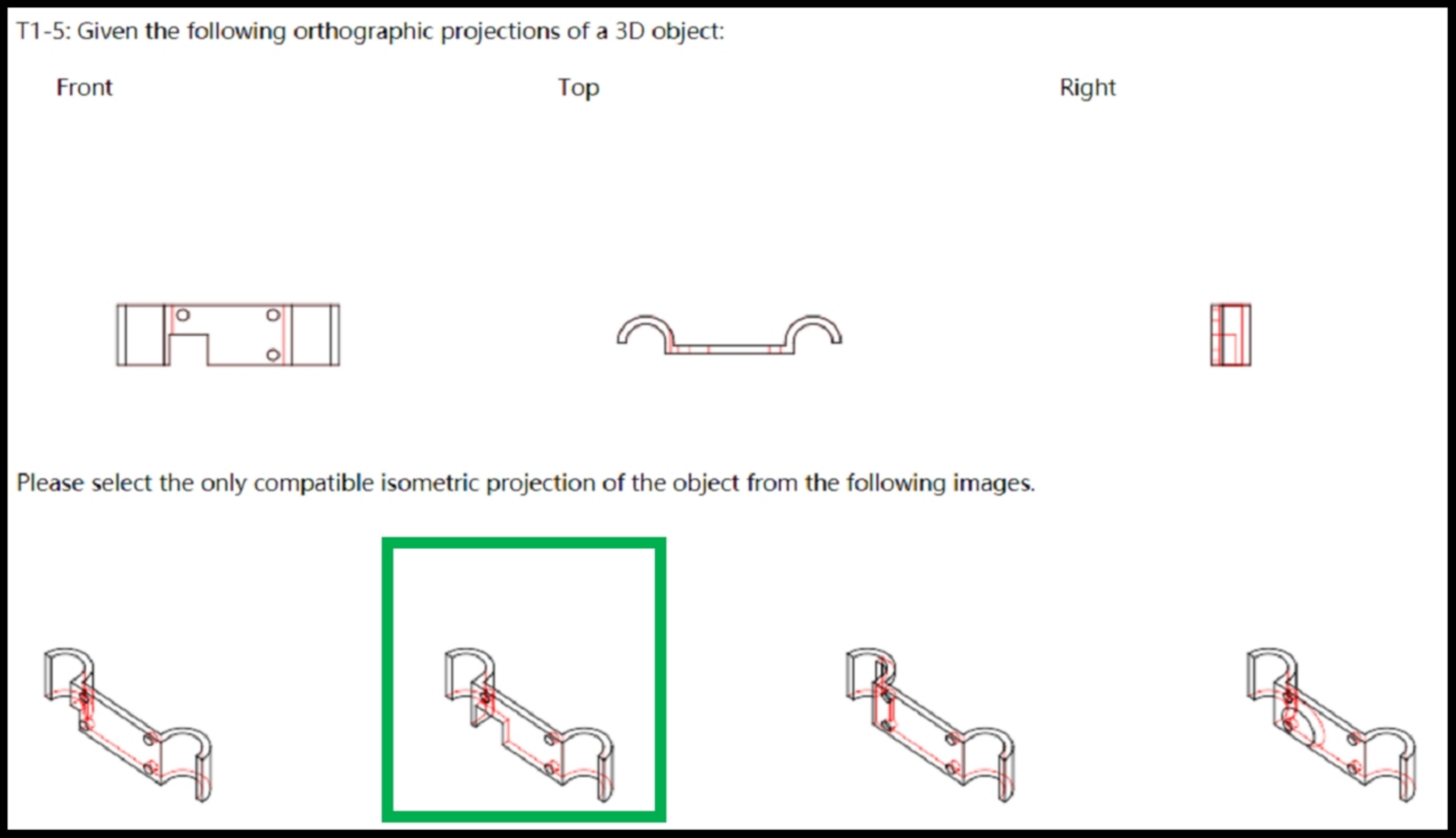}
    \includegraphics[width=\columnwidth]{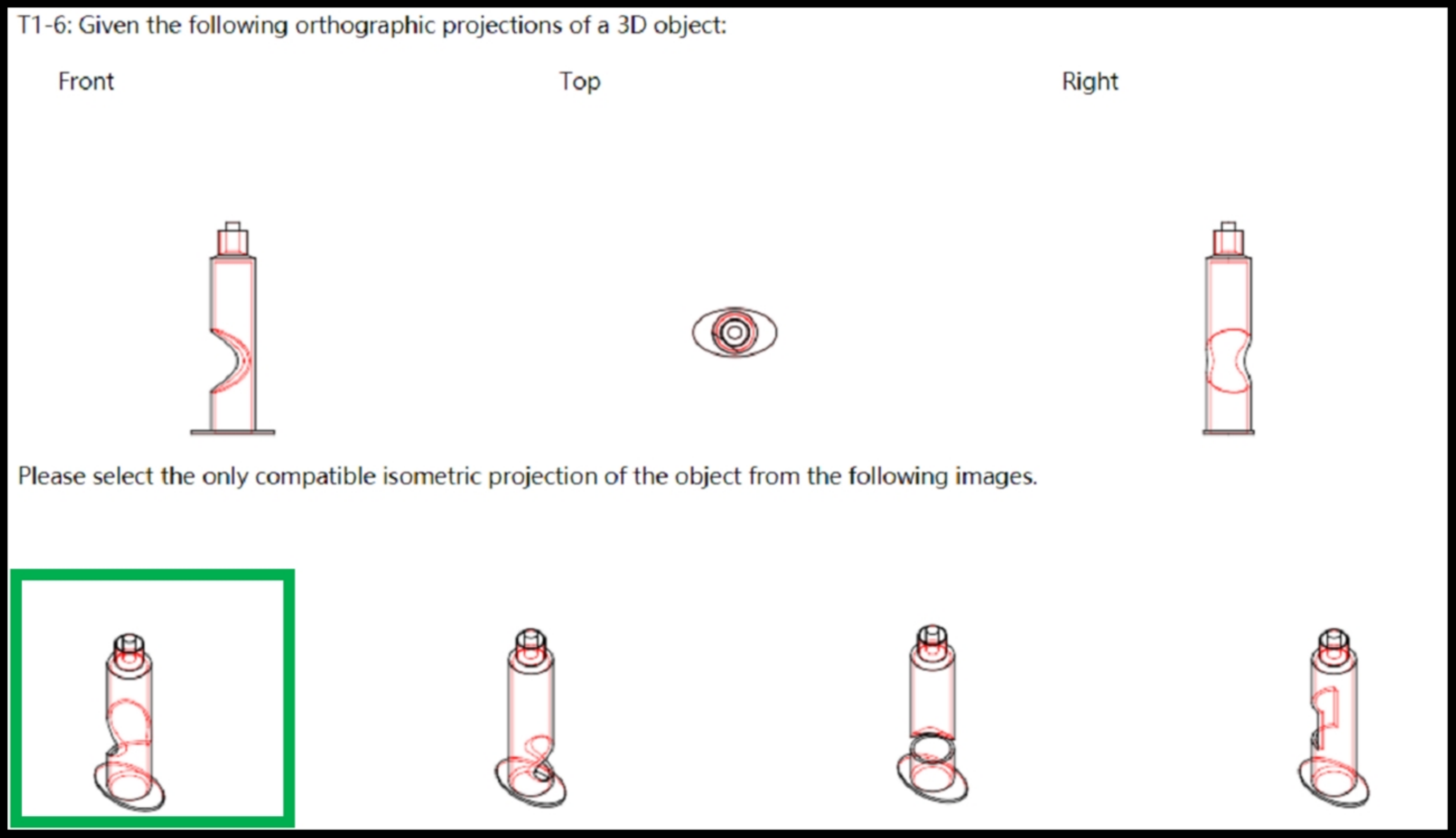}
    \includegraphics[width=\columnwidth]{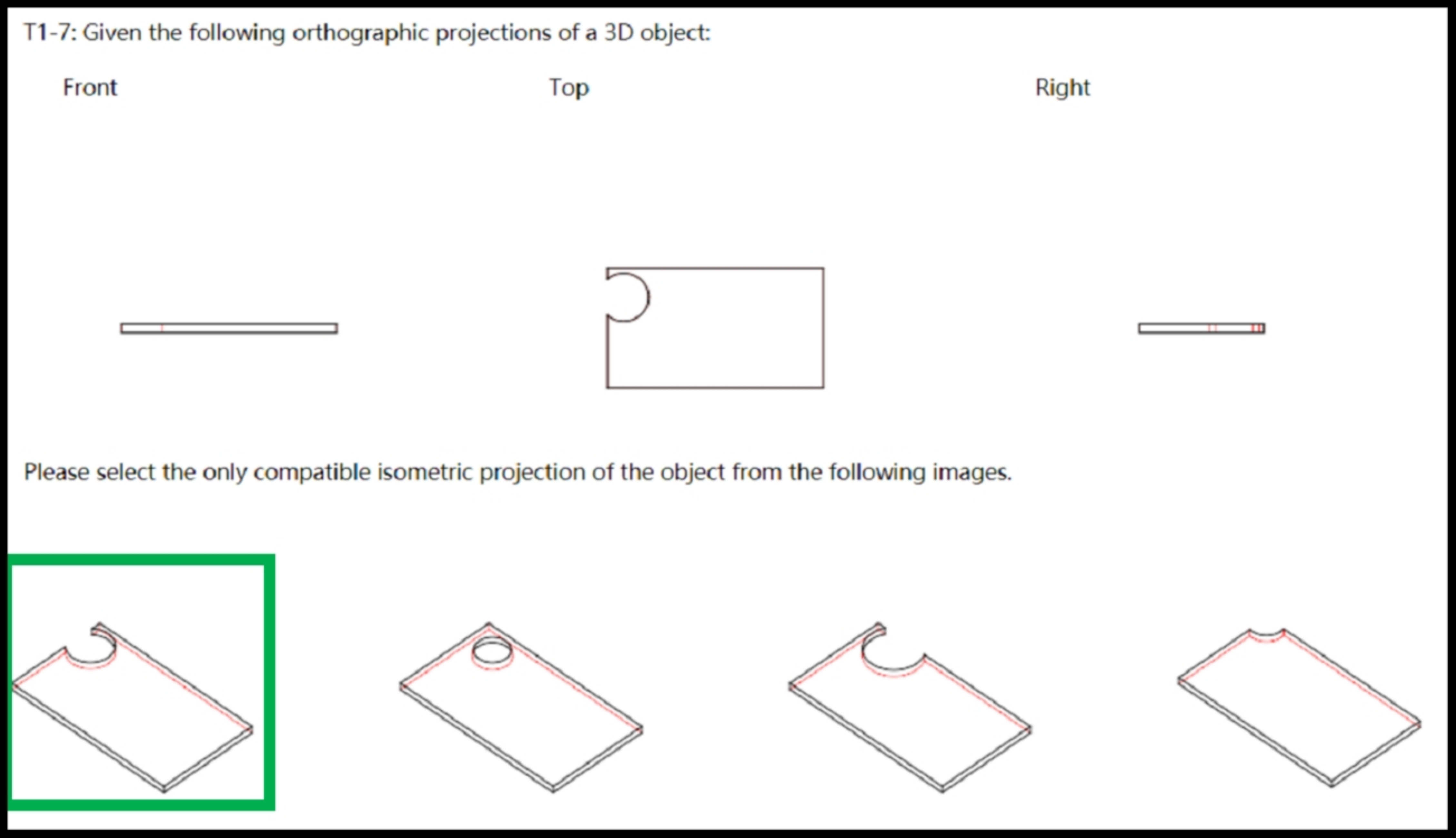}

    \caption{Examples of the ``3-View to Isometric" task shown in our crowd-sourcing website. Correct answers are highlighted by green rectangles. Best view in color.}
    \label{fig:3D-View to Isometric}
    \vspace{-3mm}
    \end{figure*}

\begin{figure*}
    \centering
    \includegraphics[width=\columnwidth]{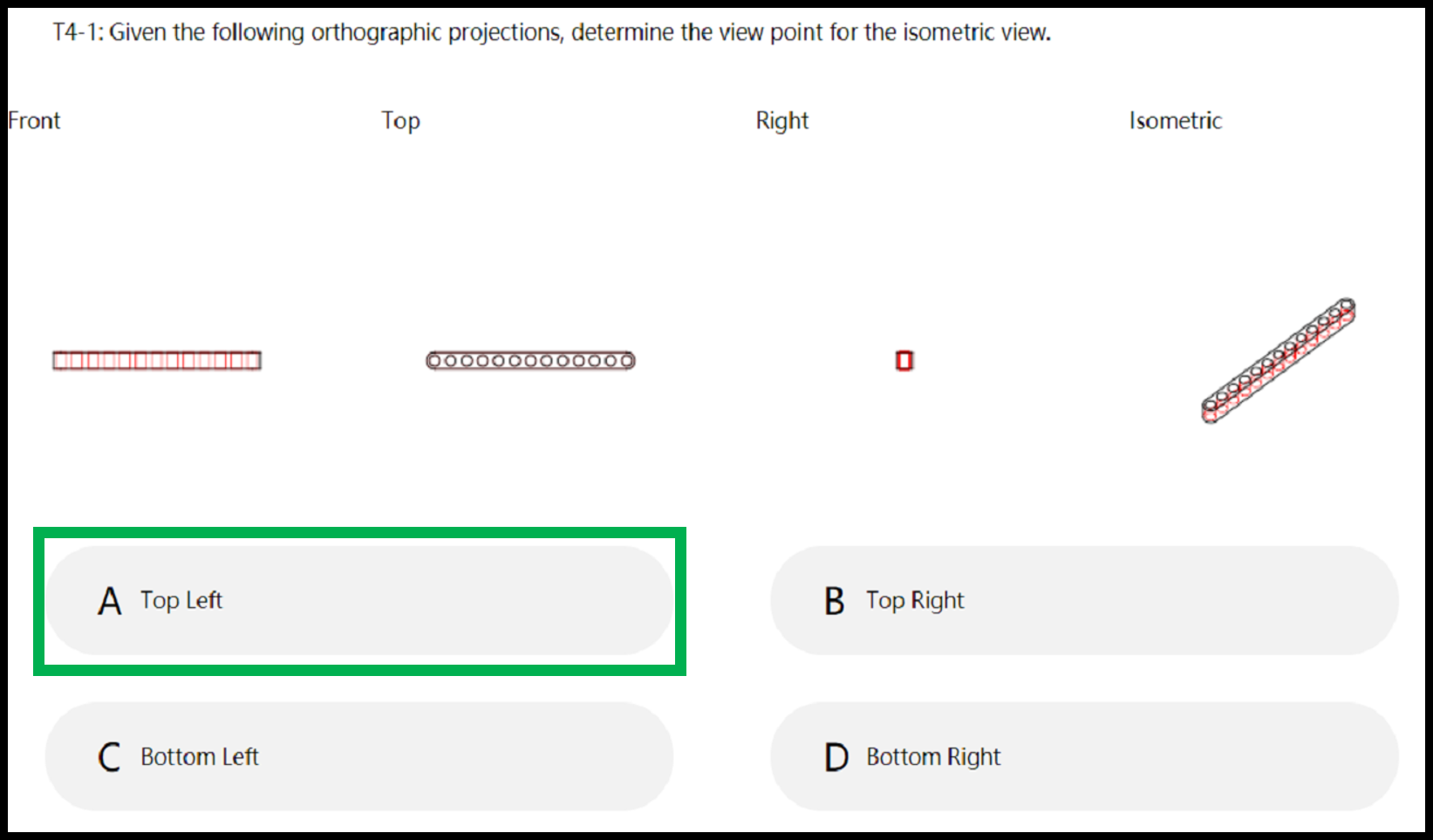}
    \includegraphics[width=\columnwidth]{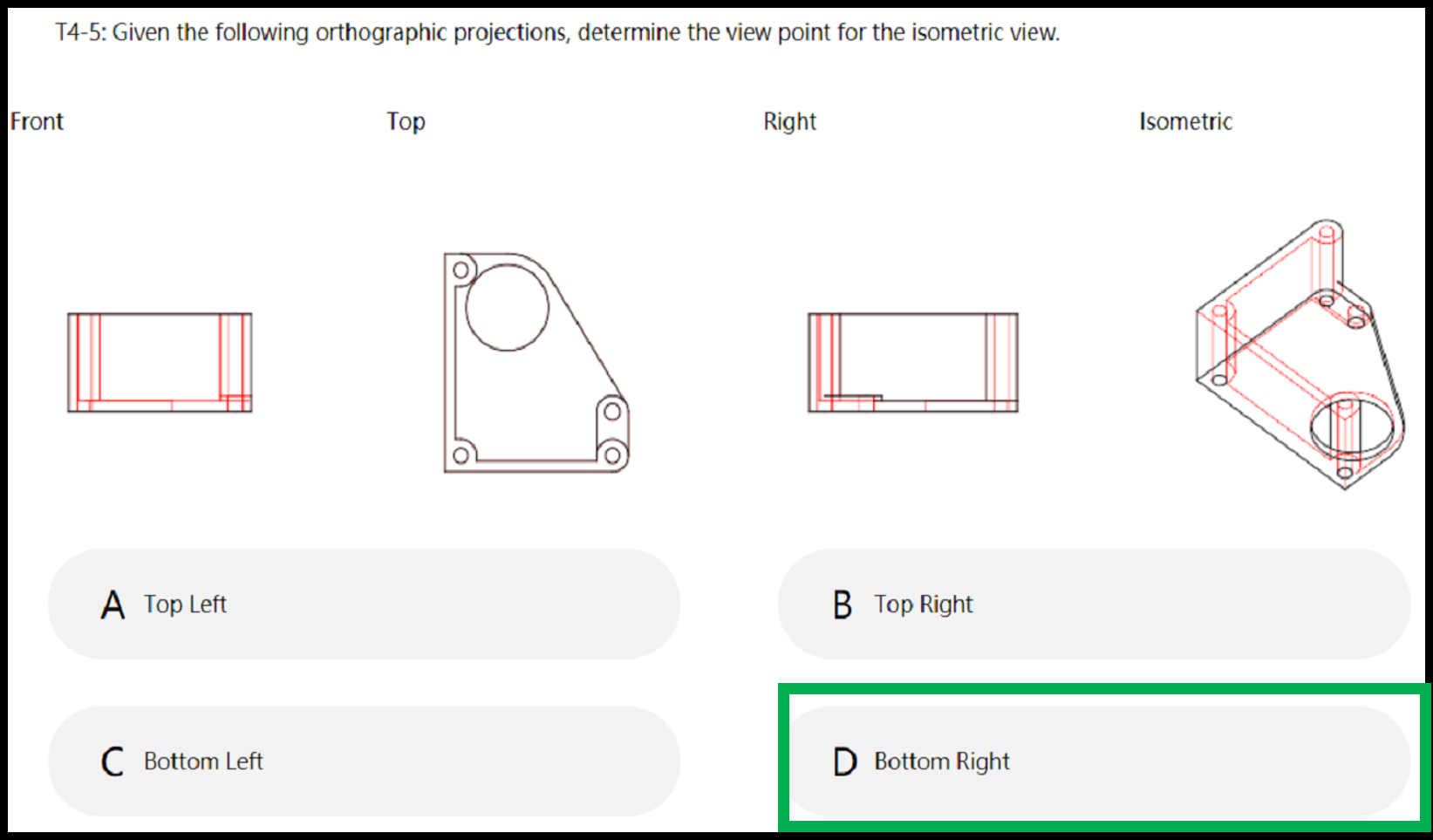} \includegraphics[width=\columnwidth]{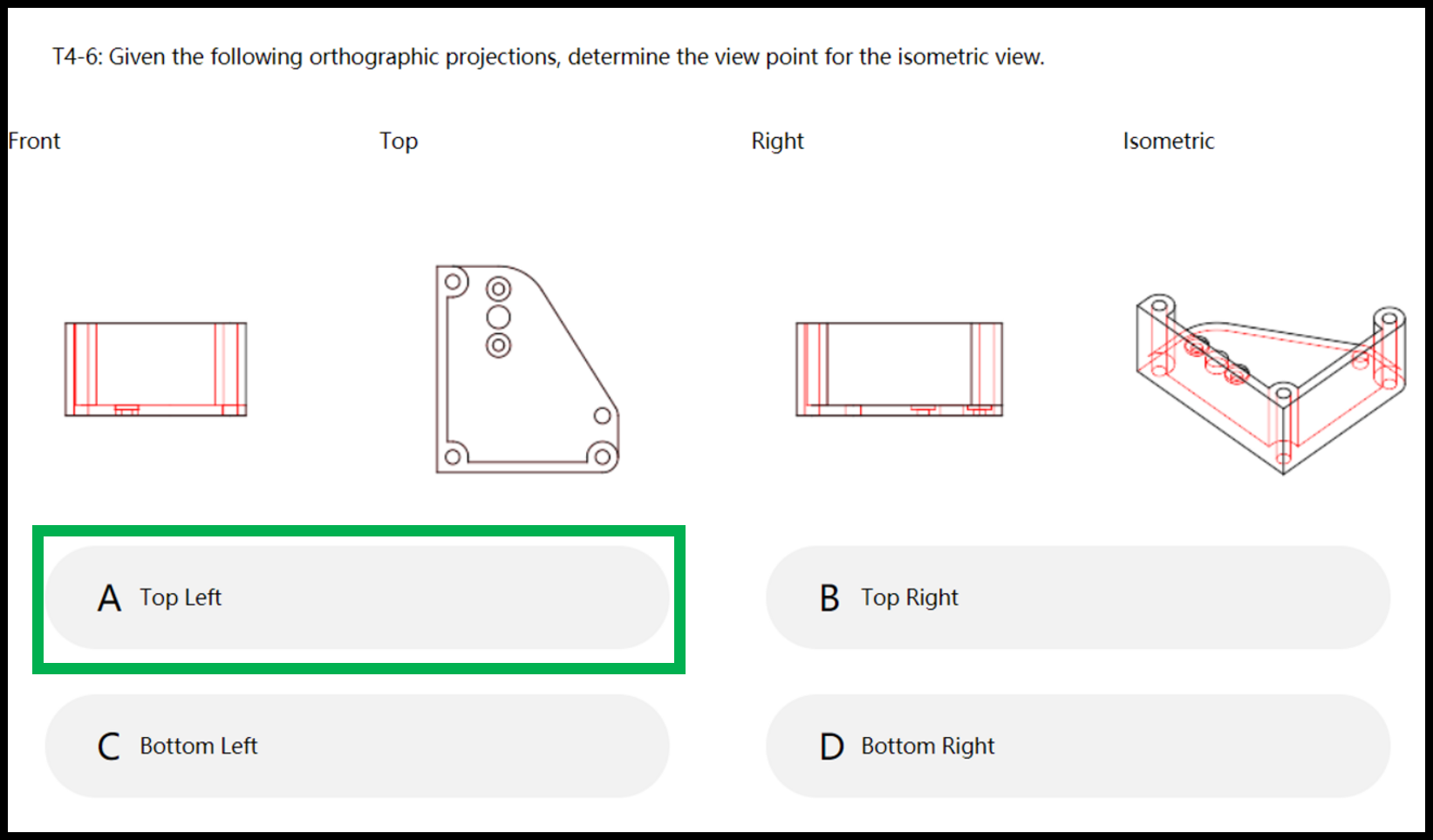}
    \includegraphics[width=\columnwidth]{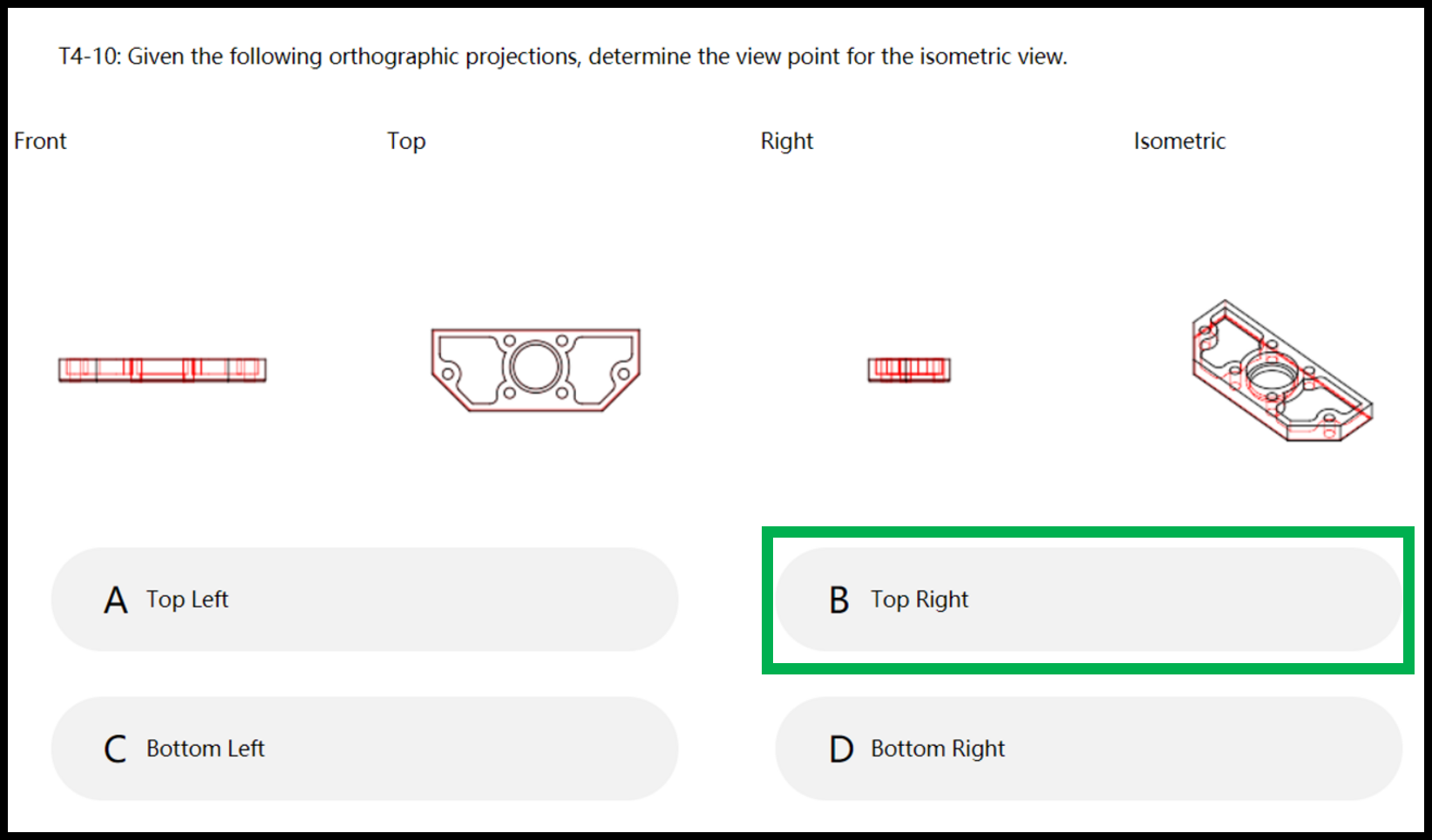}
    \includegraphics[width=\columnwidth]{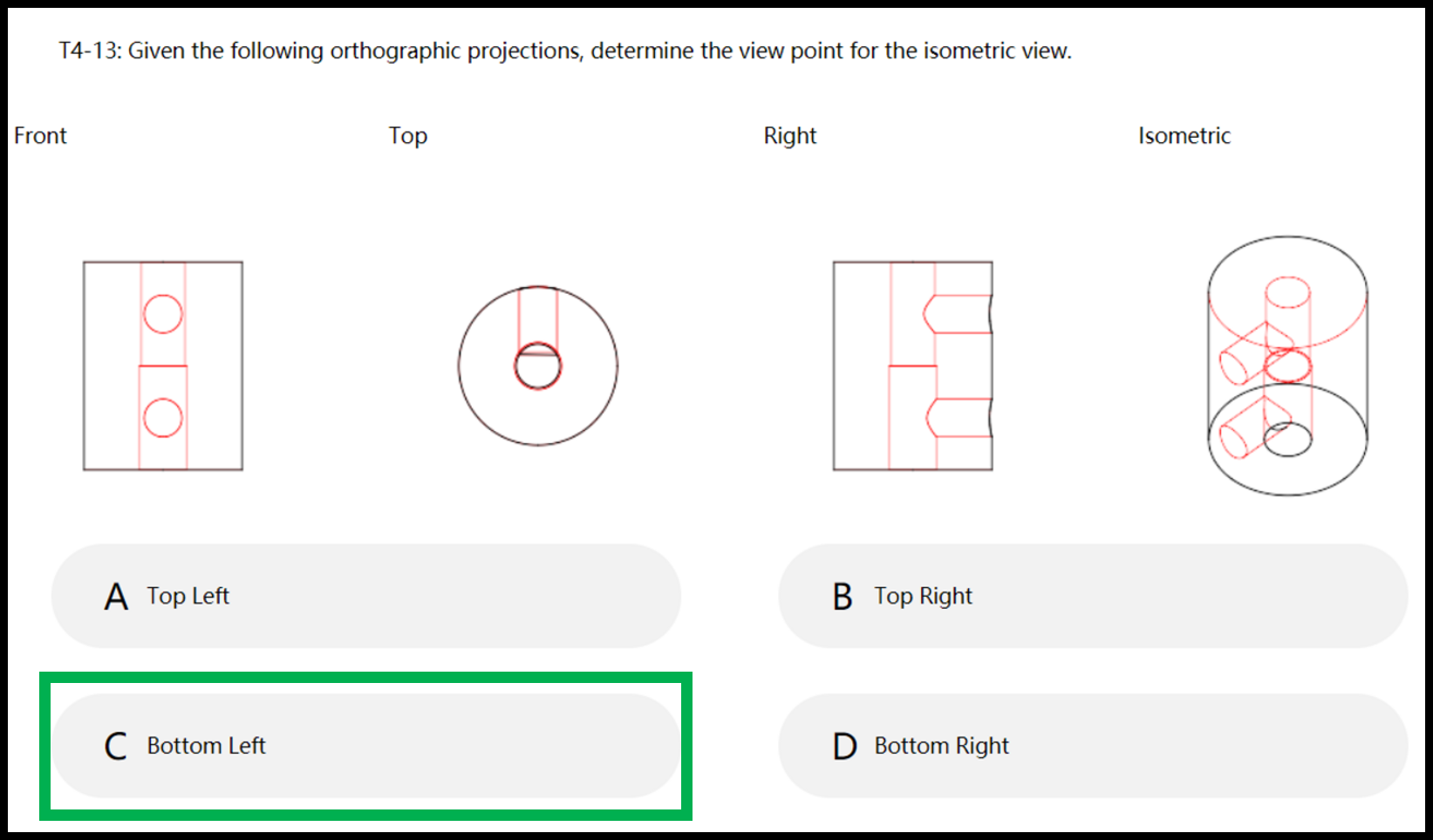} \includegraphics[width=\columnwidth]{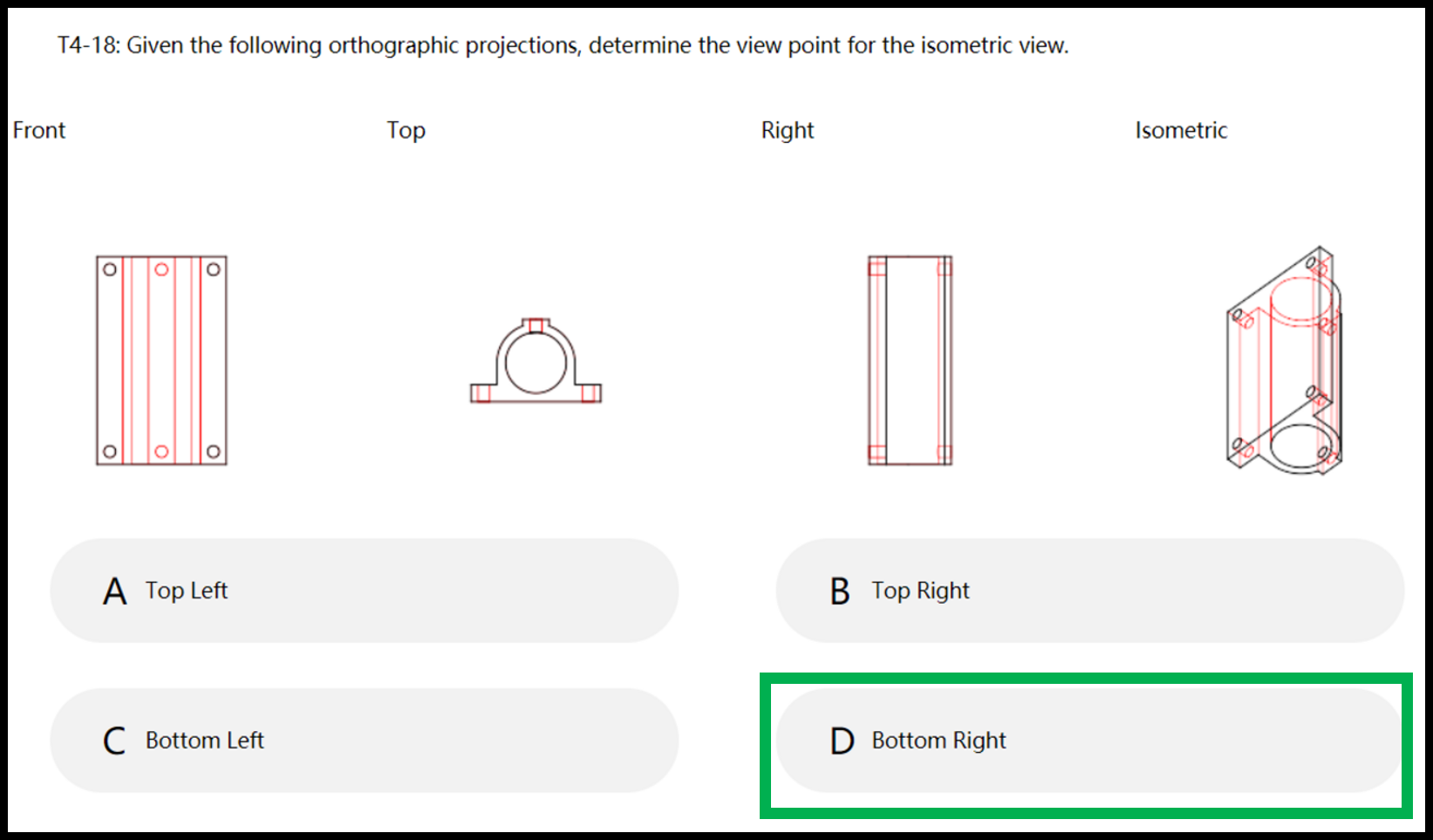}
    \includegraphics[width=\columnwidth]{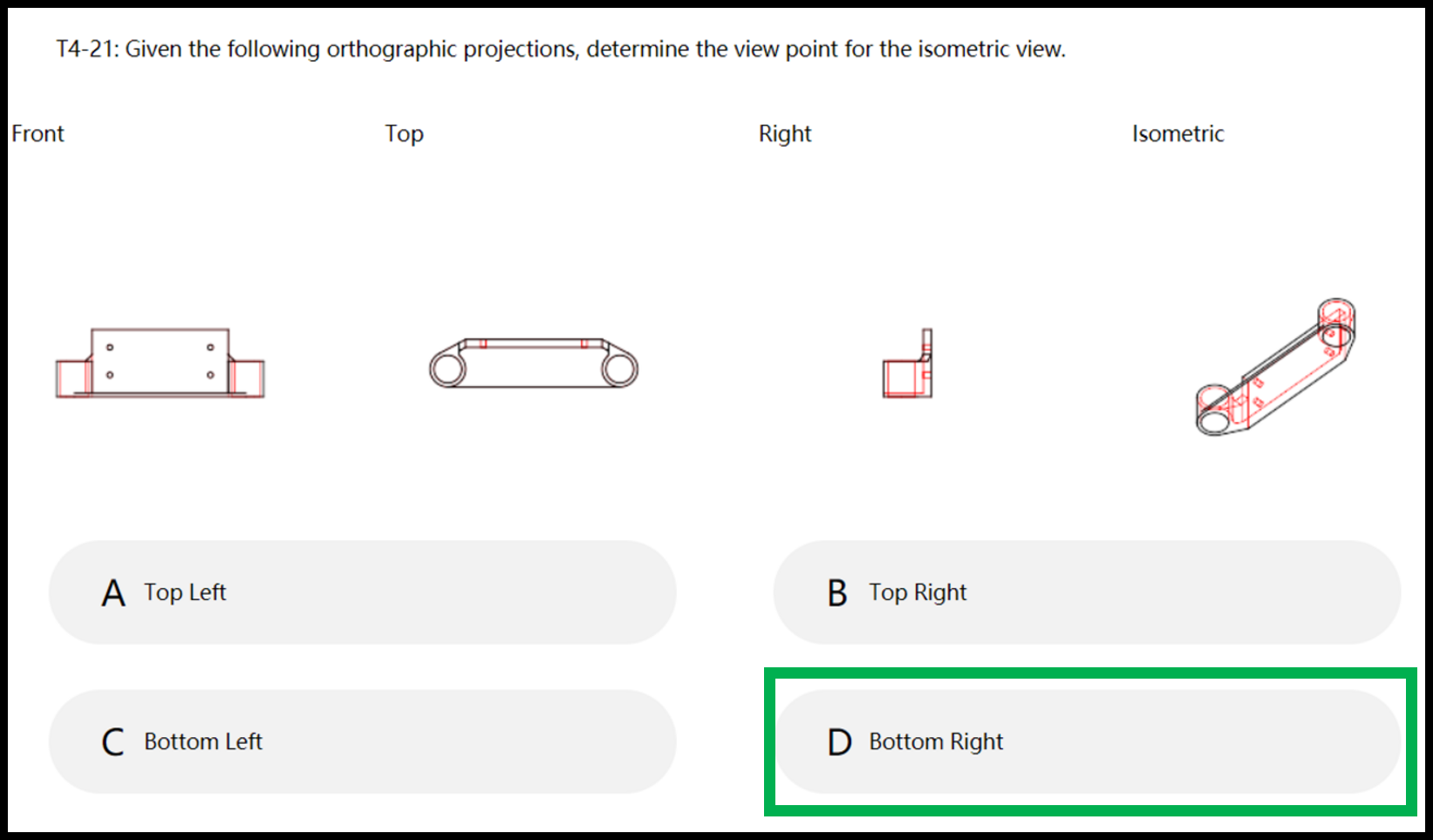}
    \includegraphics[width=\columnwidth]{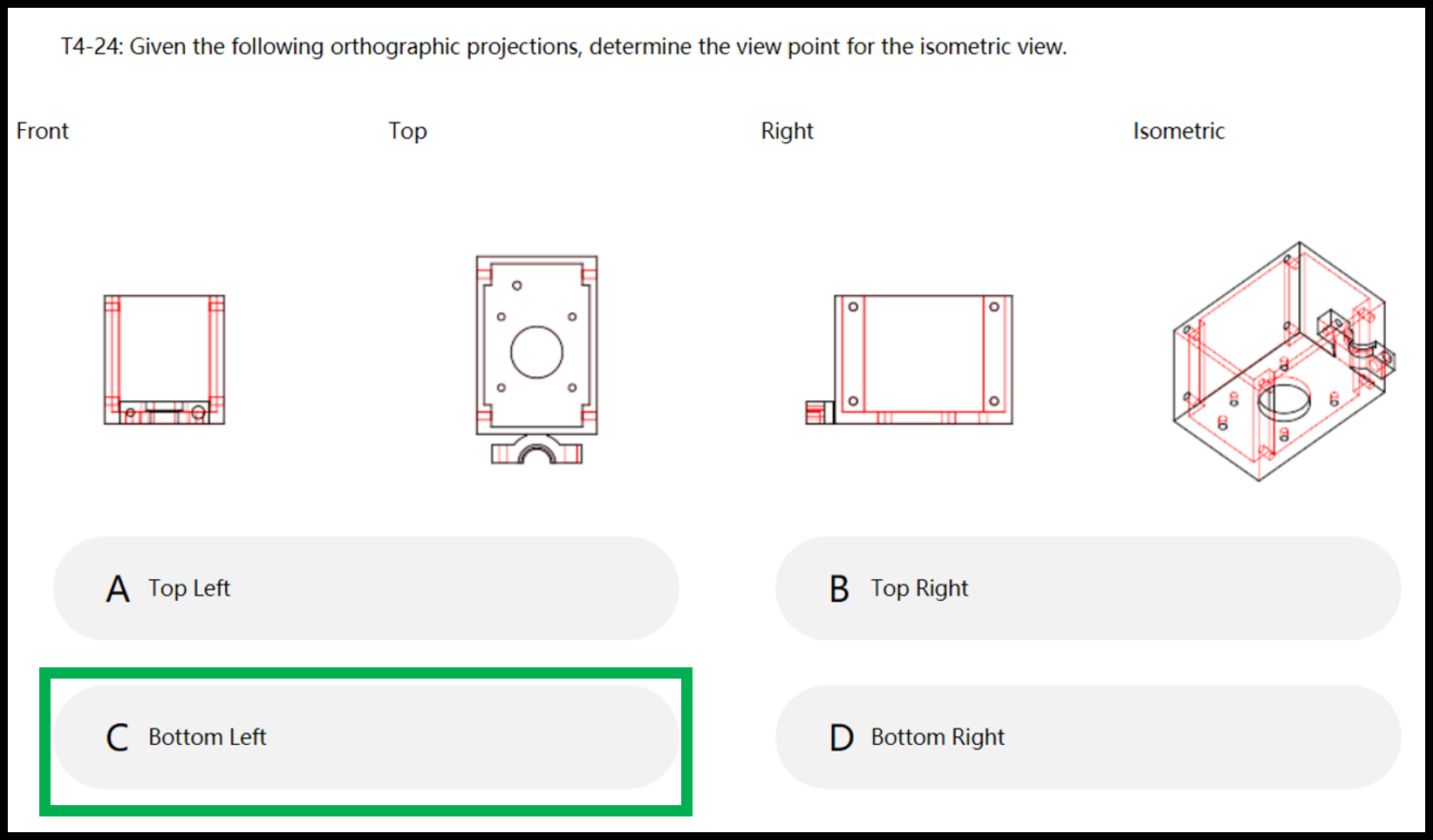}

    \caption{Examples of the ``Isometric to Pose" task shown in our crowd-sourcing website. Correct answers are highlighted by green rectangles. Best view in color.}
    \label{fig:Isometric to Pose}
    \vspace{-3mm}
    \end{figure*}

\begin{figure*}
    \centering
    \includegraphics[width=\columnwidth]{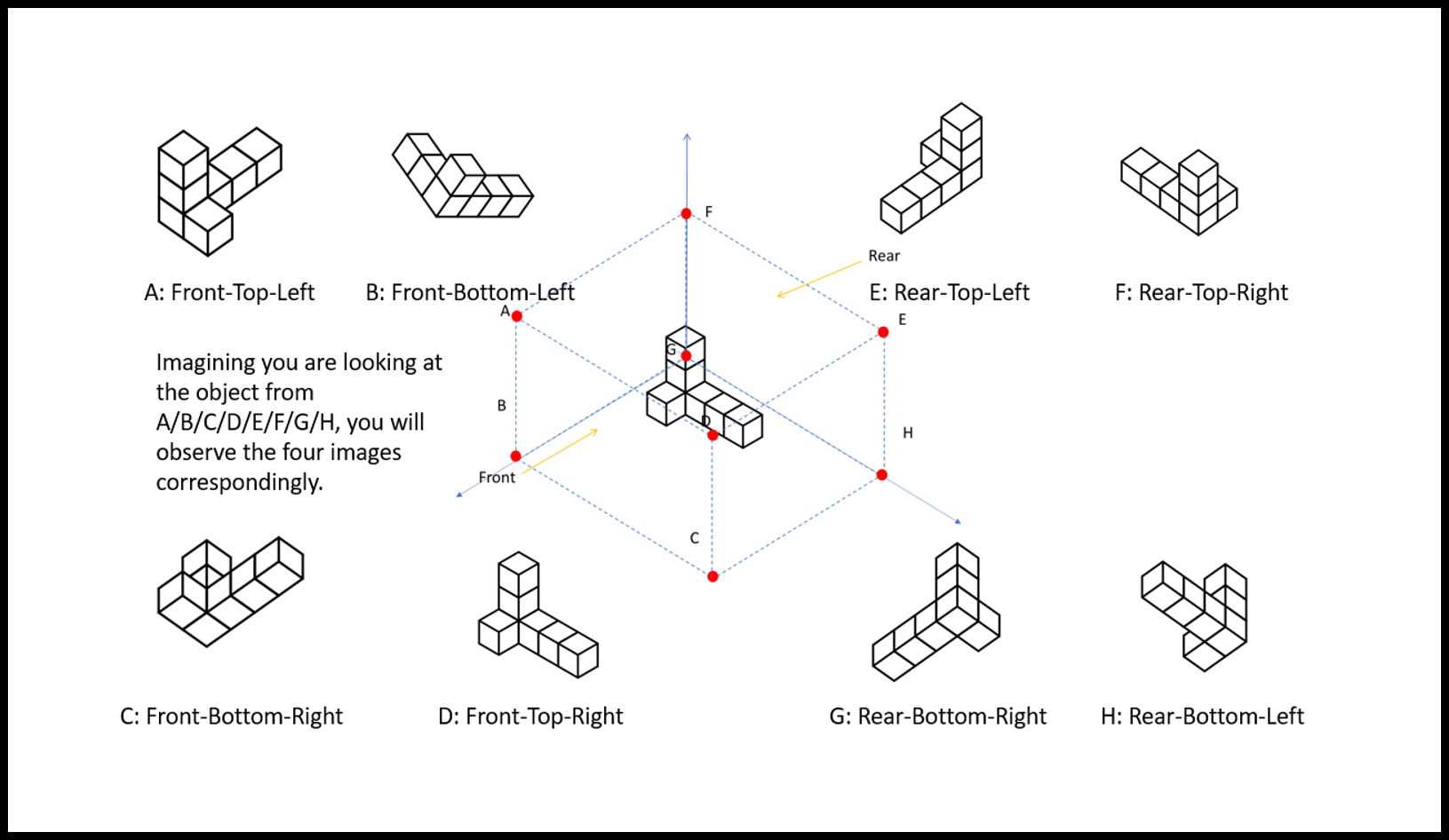}
    \includegraphics[width=\columnwidth]{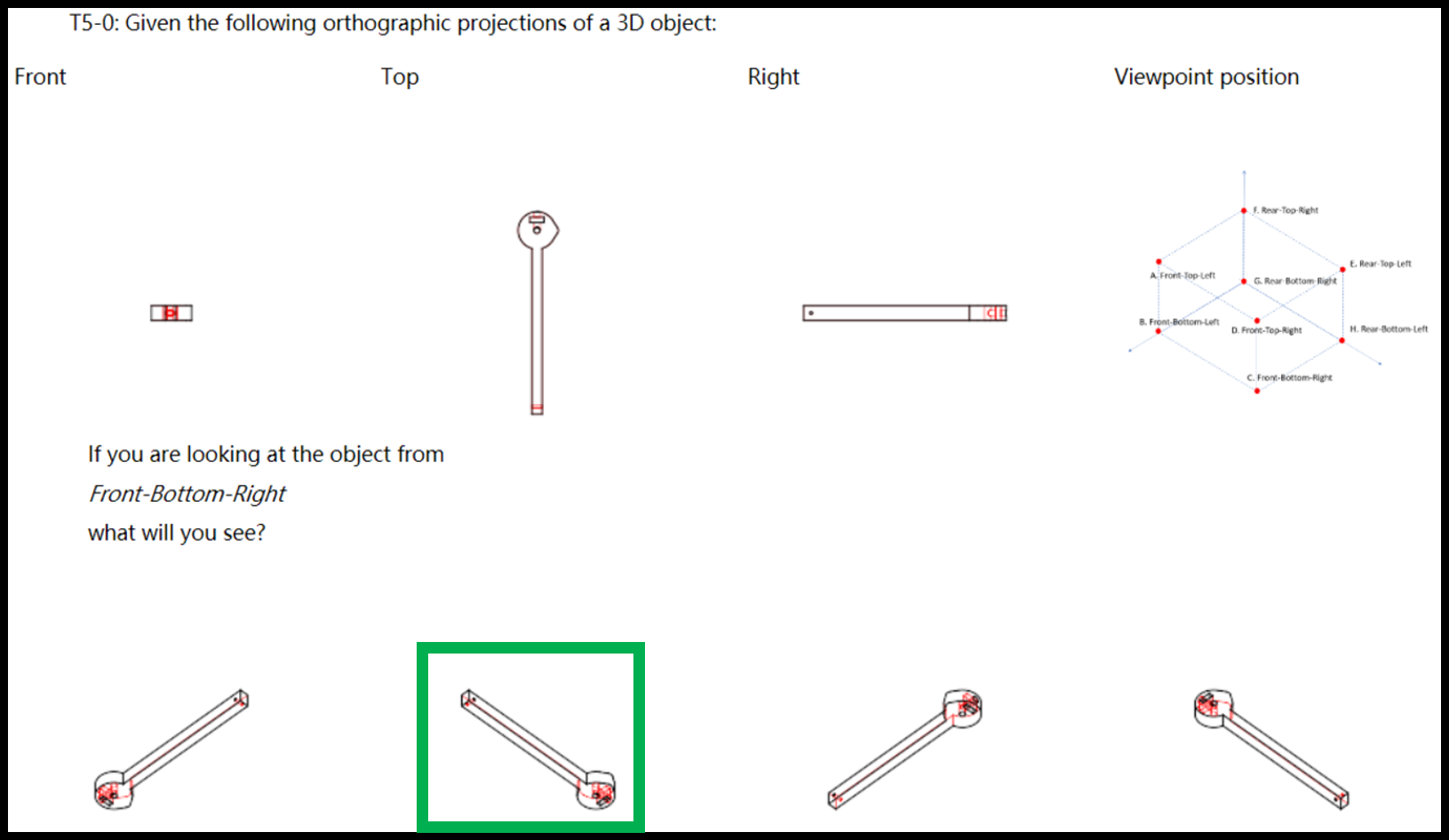}
    \includegraphics[width=\columnwidth]{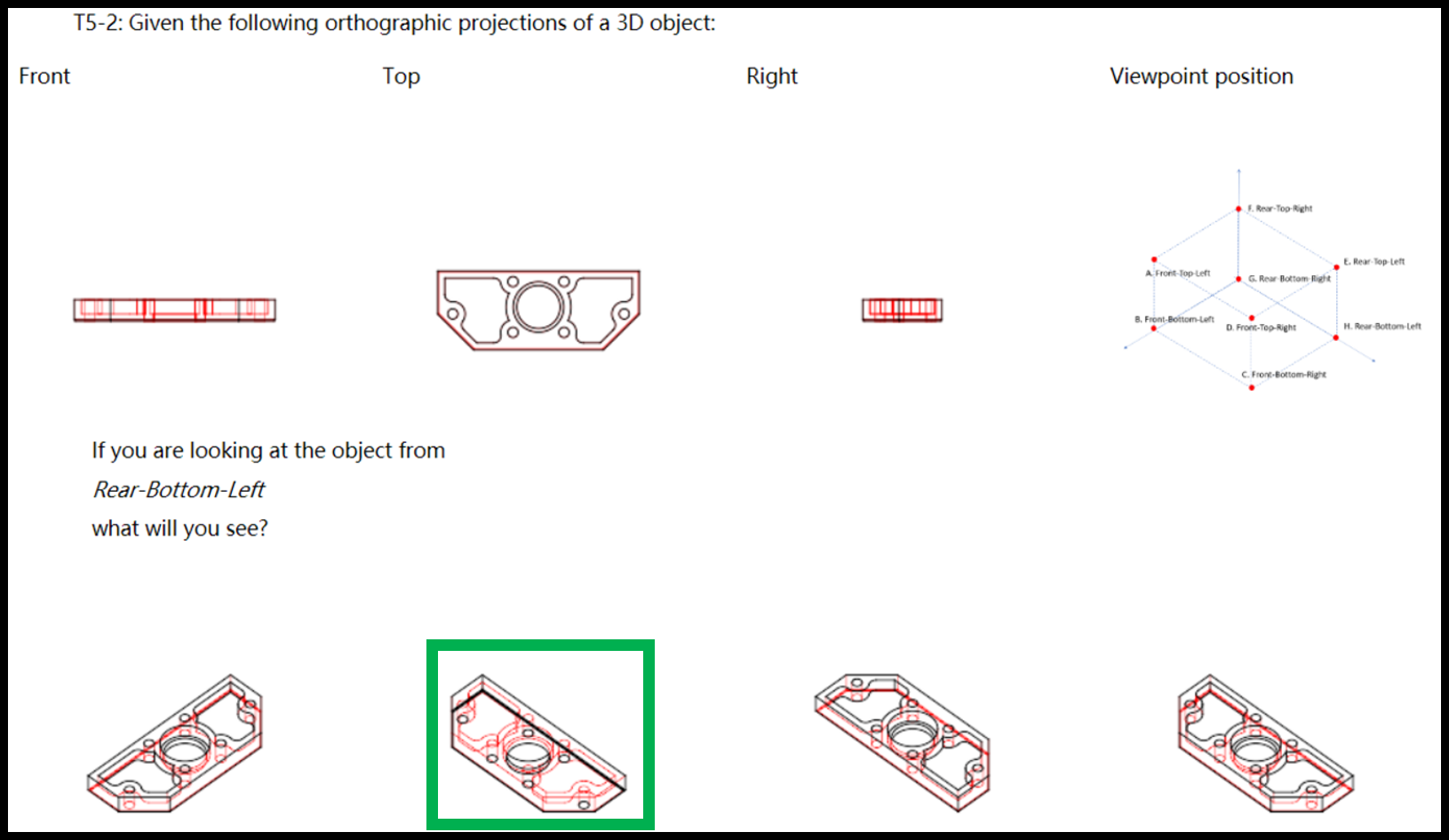}
    \includegraphics[width=\columnwidth]{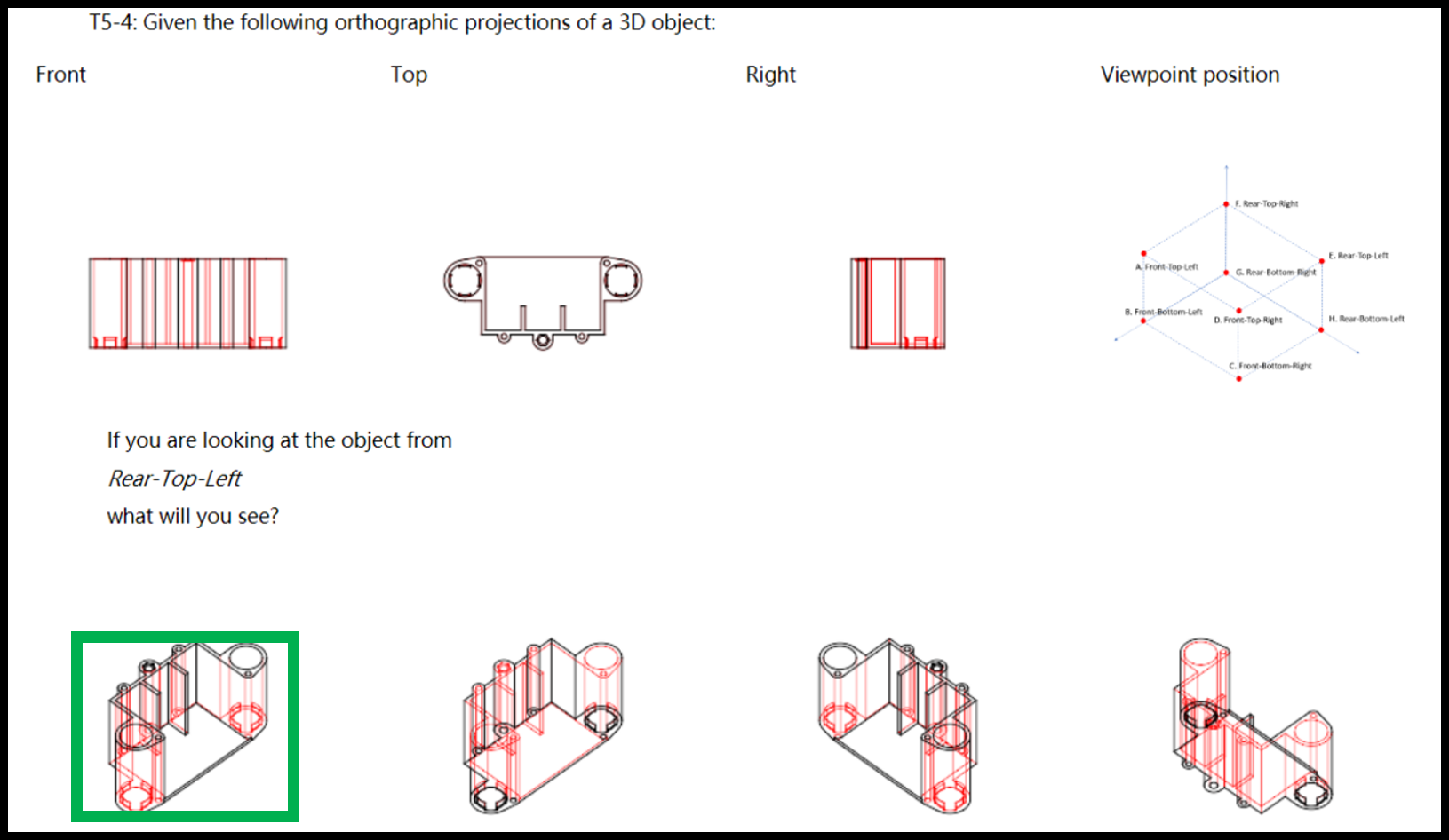}
    \includegraphics[width=\columnwidth]{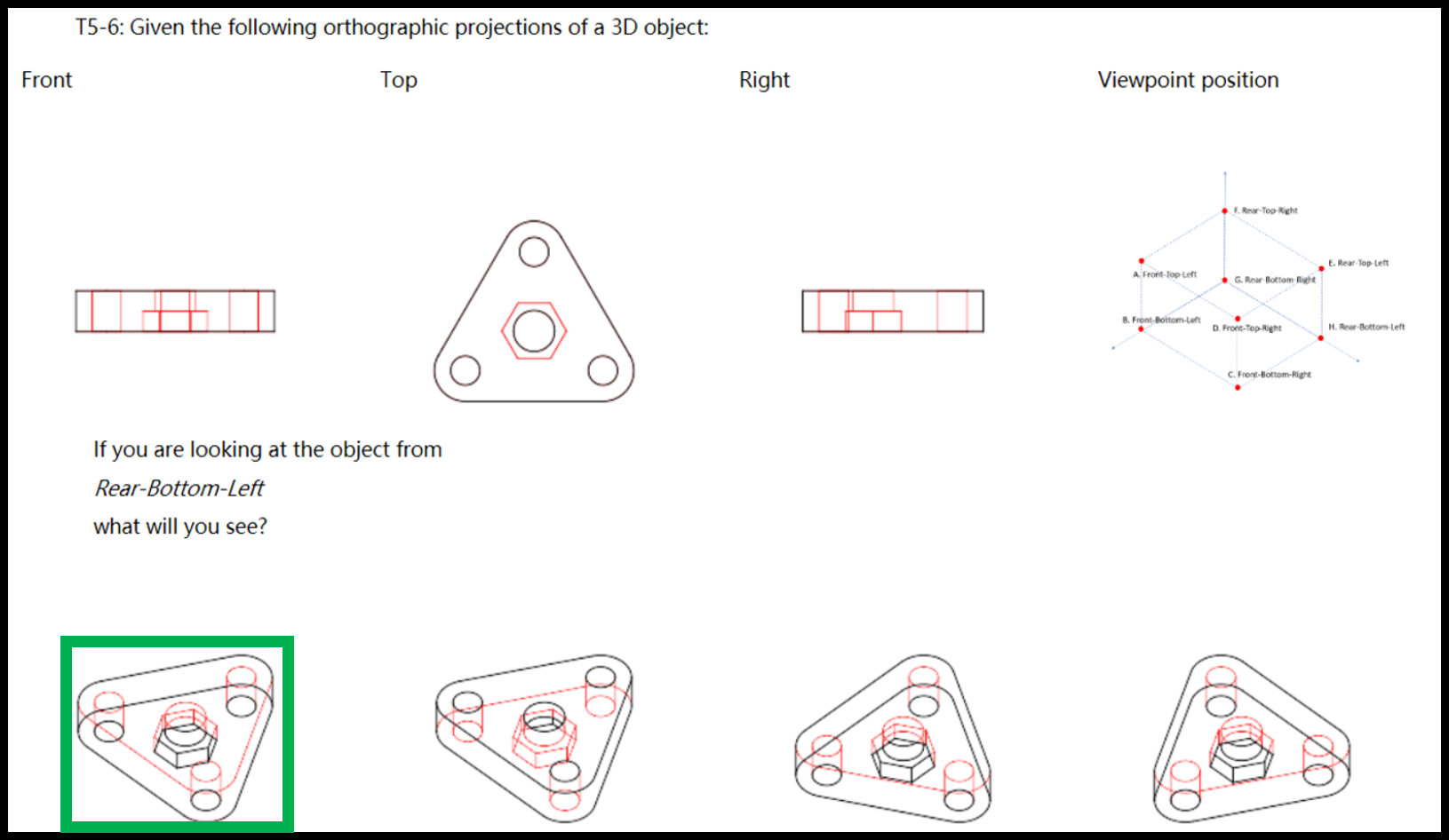}
    \includegraphics[width=\columnwidth]{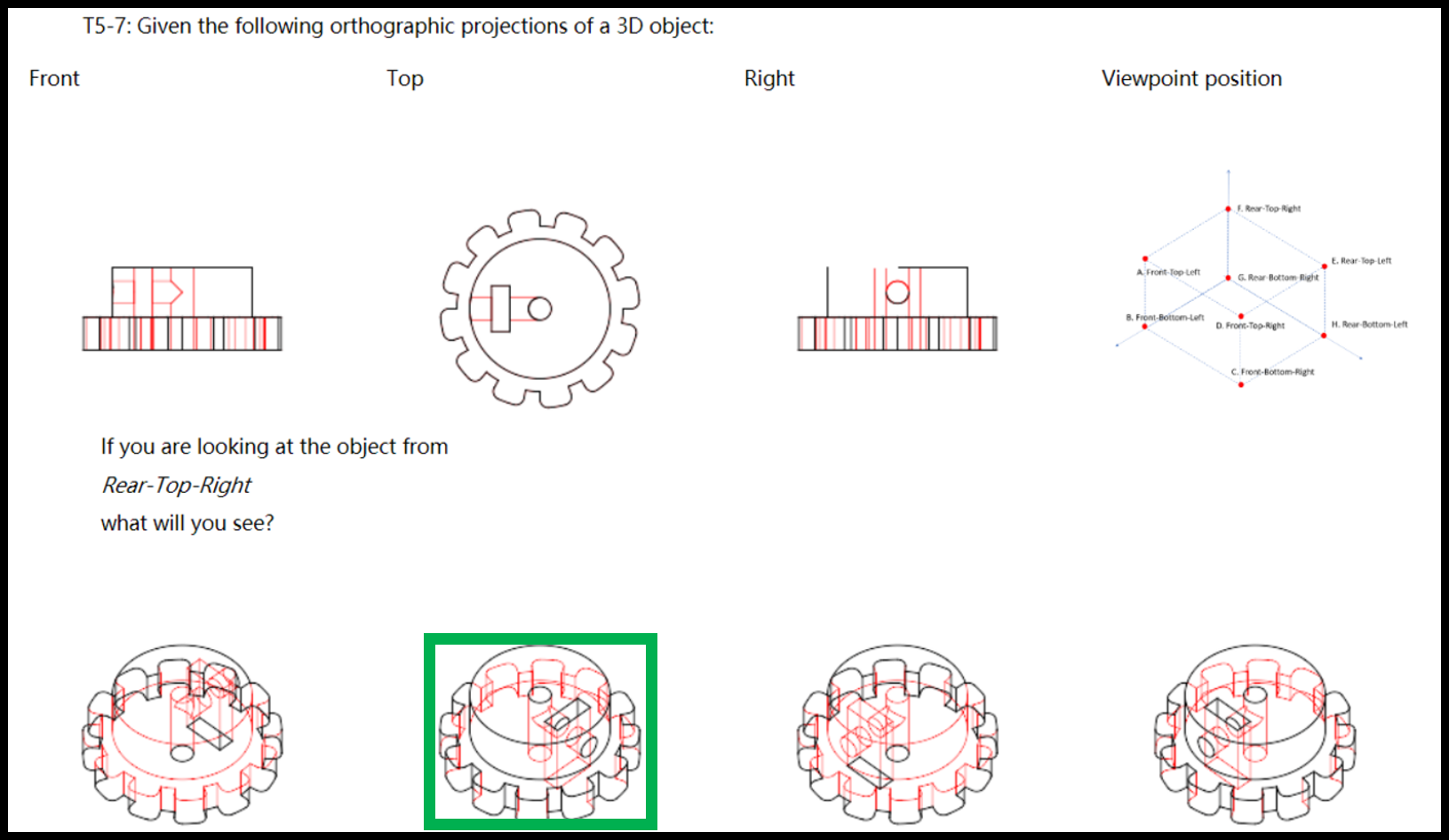}
    \includegraphics[width=\columnwidth]{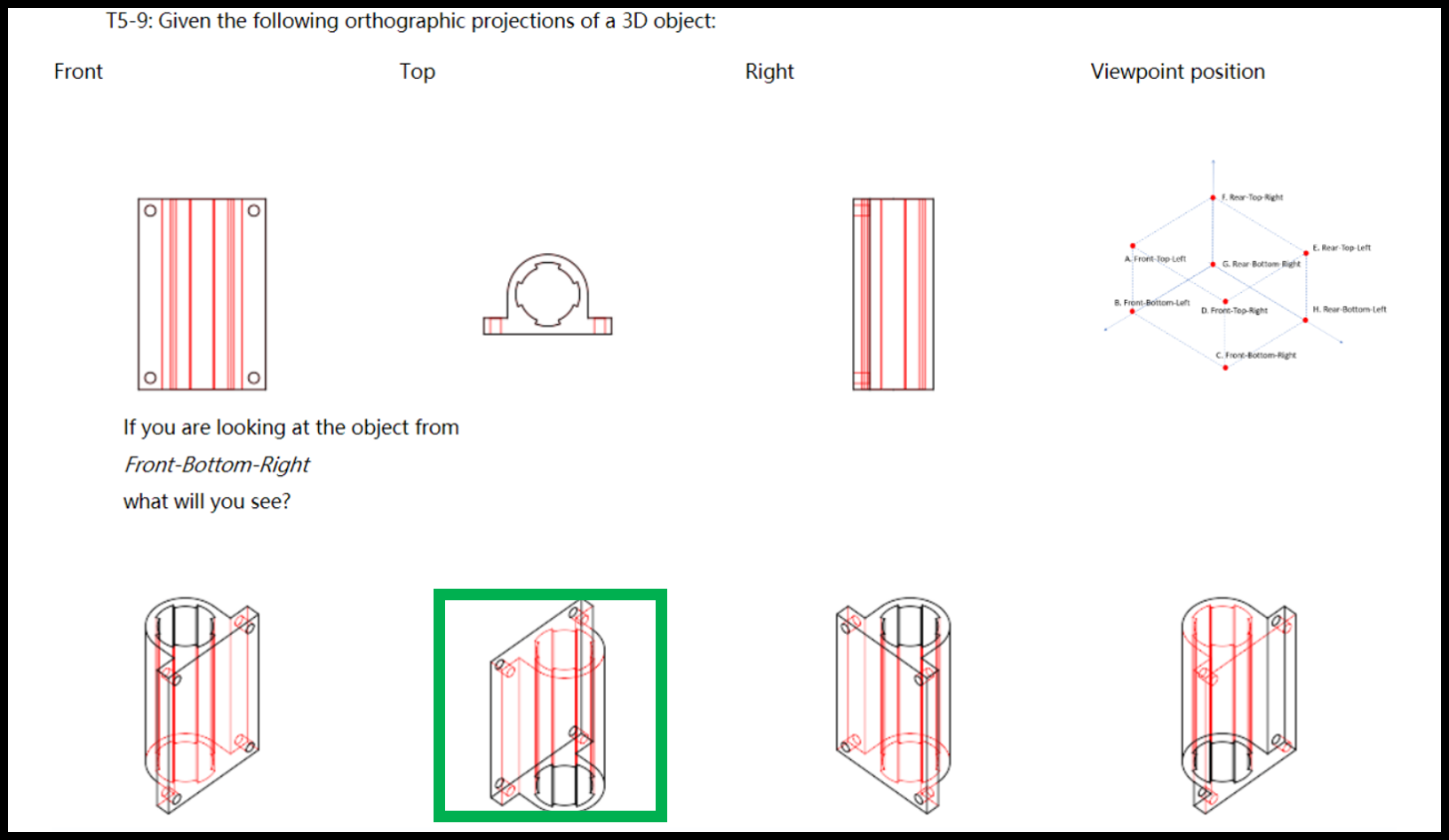}
    \includegraphics[width=\columnwidth]{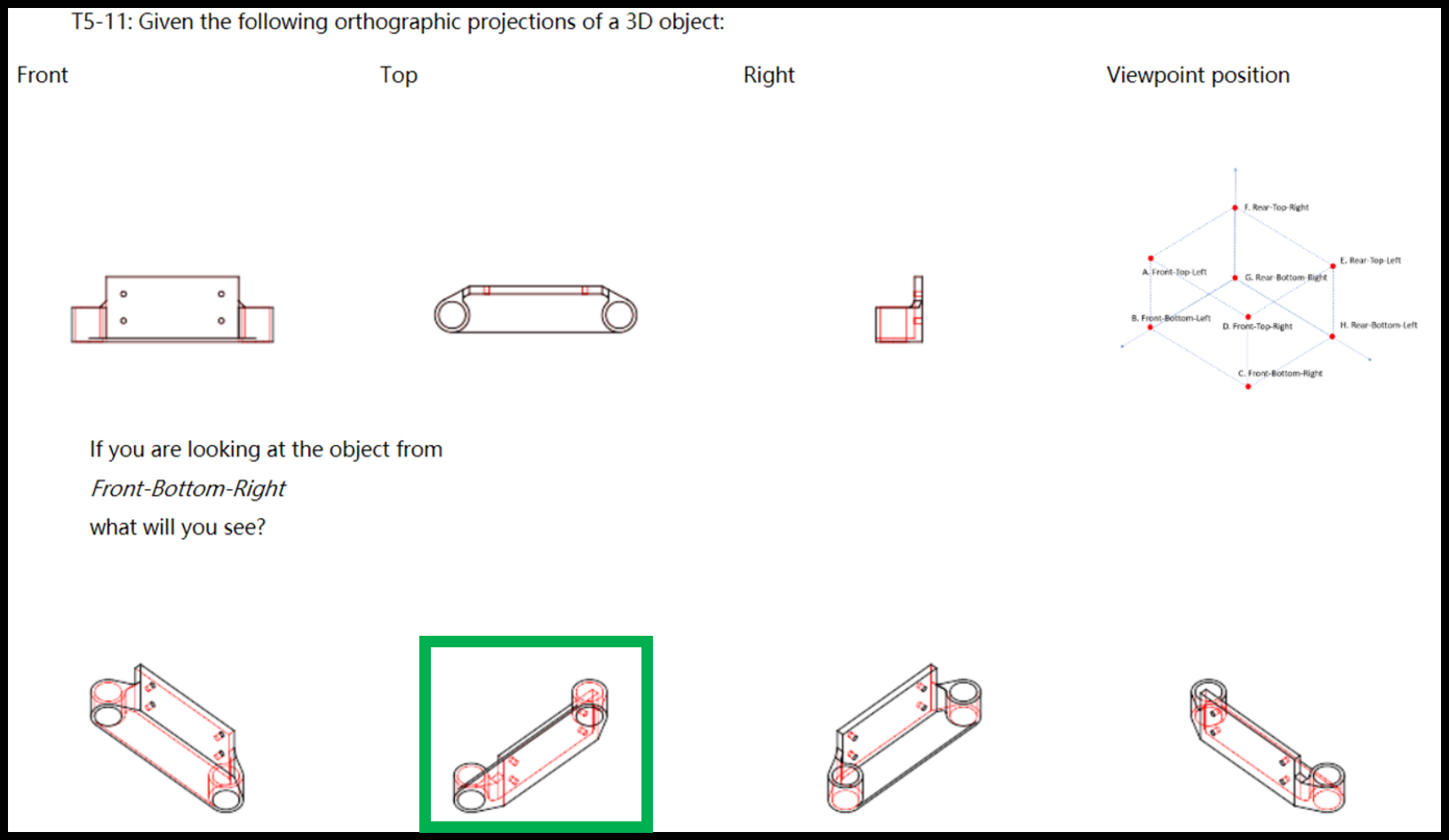}

    \caption{Examples of the ``Pose to Isometric" task shown in our crowd-sourcing website. Correct answers are highlighted by green rectangles. The eight poses are explained on the left column of the first row, and also in each question. Best view in color.}
    \label{fig:Pose to Isometric}
    \vspace{-3mm}
    \end{figure*}

Figure~\ref{fig:3D-View to Isometric},~\ref{fig:Isometric to Pose} and~\ref{fig:Pose to Isometric} show our crowd-sourcing website for collecting human performance, with example questions for each tasks respectively. 

\end{document}